\documentclass[]{LeafLabLaTexTemplate}
\input{preamble}

\usepackage{graphicx}
\usepackage{booktabs}

\usepackage{booktabs}
\usepackage[table,xcdraw]{xcolor}
\usepackage{makecell}
\usepackage{multirow} 

\usepackage{algorithm}
\usepackage{algpseudocode}
\usepackage{longtable}

\usepackage{booktabs}
\usepackage{caption}
\usepackage{subcaption}
\usepackage{amsmath} 
\usepackage{amsfonts}

\usepackage{wrapfig}
\usepackage{hyperref}
\usepackage{adjustbox}
\usepackage{algorithm}
\usepackage{algpseudocode}
\usepackage{longtable}
\usepackage{graphicx}
\usepackage{booktabs}
\usepackage{natbib}

\usepackage{etoc}
\usepackage{titletoc}

\usepackage[table,dvipsnames]{xcolor}
    
\newcolumntype{Y}{>{\centering\arraybackslash}X}
\newif\ifaddtitlelogo

\addtitlelogofalse

\title{AutoWorld: Learning Multi-Agent Traffic Simulation with Self-Supervised World Models}

\author[]{Mozhgan Pourkeshavarz}
\author[]{Tianran Liu}
\author[]{Nicholas Rhinehart}

\affiliation[]{Learning, Embodied Autonomy, and Forecasting (LEAF) Lab, University of Toronto}

\abstract{
Simulation with realistic traffic agents is essential for validating autonomous driving systems. Existing data-driven simulators learn agent behavior from higher-level abstractions such as 3D bounding boxes and polylines, inferred by upstream perception pipelines. These lossy abstractions discard sensory context that directly shapes agent behavior, limiting the distributional realism that simulation aims to reproduce. To address this limitation, we propose AutoWorld, a traffic simulation framework that grounds agent behavior in raw sensor observations through a self-supervised world model trained on LiDAR occupancy data. Given world model samples, AutoWorld constructs a coarse-to-fine predictive scene context as input to a multi-agent motion generation model. Furthermore, we designed a motion-aware latent supervision objective that enriches AutoWorld's latent representation of scene dynamics. To better exploit this latent space during inference, AutoWorld employs a cascaded Determinantal Point Process framework to guide diversity-aware sampling across both the world model and motion model. Experiments on nuPlan and Waymo demonstrate that AutoWorld improves generation realism, with larger gains in partially observed scenarios where trajectory abstractions are most limited, reducing collision rates by 36\%. AutoWorld also remains physically plausible under planner-controlled ego behavior and supports controllable, diverse safety-critical scenario generation.
}

\metadata[
\raisebox{-0.35em}{\includegraphics[width=0.05\linewidth]{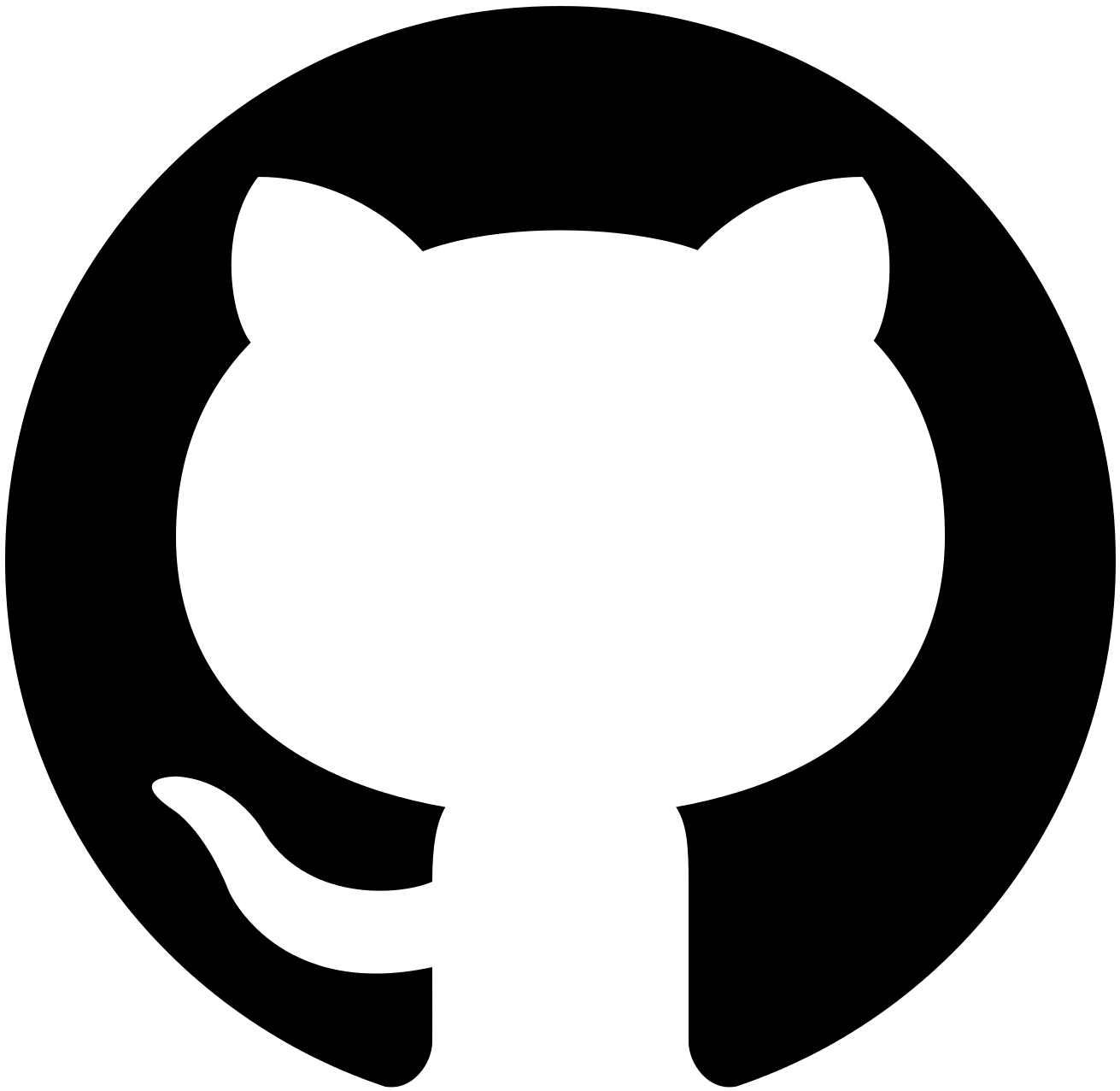}}~~ Project Website]{\href{https://github.com}{\textbf{https://github.com/mpourkeshavarz/AutoWorld}}
}

\definecolor{teagreen}{HTML}{EEF7DD}
\definecolor{peachpuff}{HTML}{FFF1E2}
\definecolor{lightcyan}{HTML}{EAF4FB}

\definecolor{teagreen}{HTML}{EEF7DD}
\definecolor{peachpuff}{HTML}{FFF1E2}
\definecolor{lightcyan}{HTML}{EAF4FB}

\usepackage[table]{xcolor}

\colorlet{rmmHigh}{blue!22}
\colorlet{rmmMid}{blue!17}   
\colorlet{rmmLow}{blue!5}

\colorlet{fvdBest}{orange!22}
\colorlet{fvdMid}{orange!15} 
\colorlet{fvdWorst}{orange!5}

\colorlet{ioudyHigh}{teal!22}
\colorlet{ioudyMid}{teal!20} 
\colorlet{ioudyLow}{teal!5}

\colorlet{ablRmmHigh}{blue!22}     
\colorlet{ablRmmMoDpp}{blue!16}    
\colorlet{ablRmmNoDpp}{blue!13}    
\colorlet{ablRmmOccDpp}{blue!11}   
\colorlet{ablRmmLow}{blue!5}       

\colorlet{rmmHigh}{blue!22}
\colorlet{rmmMid}{blue!17}   
\colorlet{rmmLow}{blue!5}

\colorlet{fvdBest}{orange!22}
\colorlet{fvdMid}{orange!15} 
\colorlet{fvdWorst}{orange!5}

\colorlet{ioudyHigh}{teal!22}
\colorlet{ioudyMid}{teal!20} 
\colorlet{ioudyLow}{teal!5}

\colorlet{ablRmmHigh}{blue!22}     
\colorlet{ablRmmMoDpp}{blue!16}    
\colorlet{ablRmmNoDpp}{blue!13}    
\colorlet{ablRmmOccDpp}{blue!11}   
\colorlet{ablRmmLow}{blue!5}       

\definecolor{gapBlueLow}{RGB}{232,244,255}
\definecolor{gapBlueMod}{RGB}{170,210,240}
\definecolor{gapBlueHigh}{RGB}{80,150,210}

\definecolor{gapBlue}{RGB}{45,120,190}
\definecolor{dzOrange}{RGB}{230,115,35}

\definecolor{dzOrangeLow}{RGB}{255,246,224}
\definecolor{dzOrangeMod}{RGB}{255,205,123}
\definecolor{dzOrangeHigh}{RGB}{245,135,48}

\begin{document}

\maketitle

\section{Introduction}

Traffic simulation is essential for developing and testing autonomous driving systems. Data-driven approaches have gained increasing attention for their ability to overcome limitations of heuristic-based simulators. Existing methods \cite{huang2024versatile,jiang2024scenediffuser,zhou2024behaviorgpt,wu2024smart,philion2024trajeglish,zhang2025trajtok,zhang2025closed,pei2025advancing,combot} learn agent behavior from motion forecasting datasets \cite{ettinger2021large,caesar2020nuscenes}, which represent driving scenes as sequences of 3D bounding boxes and polylines, abstractions inferred by upstream perception pipelines. However, these trajectory abstractions are a lossy compression of the original sensory observations, discarding rich scene information that shapes how real drivers behave, including geometric structure and occlusion context. For example, a driver slowing near a blind intersection responds to raw sensory geometry that no bounding box encodes. Models trained on these abstractions therefore learn the kinematic response without the sensory cause, which limits their ability to reproduce the distributional realism that faithful simulation requires. While \cite{chen2024womd} demonstrated that augmenting trajectory data with raw LiDAR observations improves agent behavior generation, how to effectively harness this signal for traffic simulation remains largely unexplored.

Models that forecast future sensor observations, with or without action conditioning, are broadly referred to as ``world models'' and have proven effective for modeling high-dimensional environments \cite{zhu2024sora,kong20253d}. In autonomous driving, video-generative world models have been widely explored, with a primary focus on photorealistic scene synthesis \cite{hu2023gaia,gao2024vista,russell2025gaia,yang2024generalized,wang2024driving}. To favor geometric information, a complementary line of work investigates world modeling through LiDAR data, framed as LiDAR or occupancy forecasting \cite{Weng2021,Weng2022,Zhang2023,liu2025foundationallidarworldmodels}. Yet, these methods typically rely on semantic occupancy prediction, which requires costly dense annotation, and apply uniform supervision across spatial regions, underemphasizing the motion dynamics most relevant to behavior generation for traffic simulation.

\begin{figure*}[t]
    \centering 
    \includegraphics[width=0.9\columnwidth,  trim=1.6cm 12cm 1.4cm 3.7cm, clip]{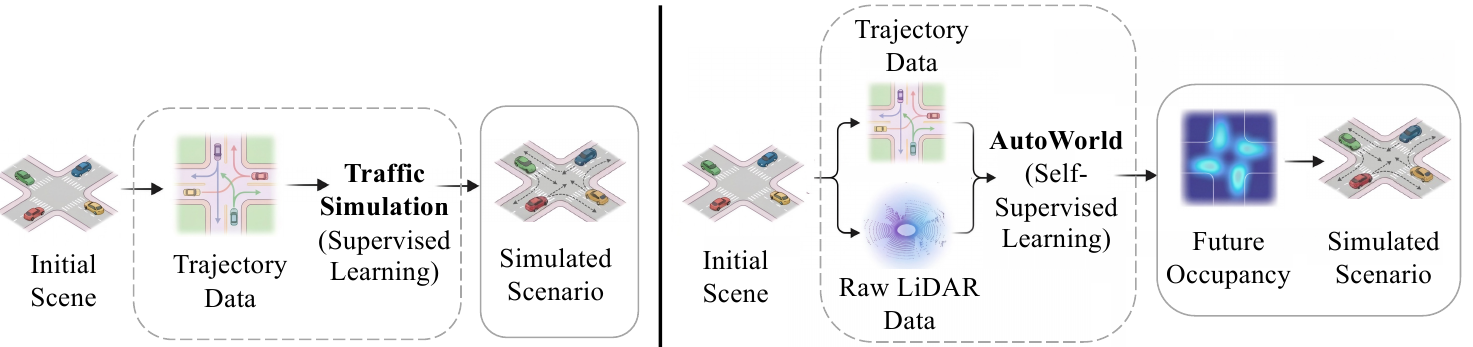}\vspace{-1mm}
    \caption{\textbf{Comparison of traffic simulation approaches}. (Left) Conventional traffic simulation methods. (Right) AutoWorld grounds simulation in future scene occupancies predicted by a self-supervised LiDAR world model, preserving sensory context discarded by trajectory representations.} \vspace{-6mm}
    \label{fig:teaser_fig}
\end{figure*}

In this work, rather than learning agent behavior solely from trajectory abstractions, we propose to ground traffic simulation in a world model learned directly from raw LiDAR observations (See Fig.~\ref{fig:teaser_fig}). Our key insight is that a self-supervised world model of the driving environment can capture the geometric and dynamic context that trajectory data discards, providing a richer foundation for realistic behavior generation. To this end, we introduce \textbf{AutoWorld}, a traffic simulation framework that conditions a multi-agent motion generation model on world model predictions, transferring the world model's sensor-grounded understanding of scene dynamics into the simulation process.

AutoWorld trains a latent LiDAR world model with a motion-aware supervision objective that prioritizes dynamic scene elements without using semantic labels, allowing fully self-supervised training. We then propose a conditional diffusion policy that leverages future latent occupancies and a predictive scene context, encapsulating the anticipated occupancy sequence, to model the joint distribution of future agent behaviors. By doing so, AutoWorld is equipped with a coarse-to-fine context derived from sensory observation. Notably, agent identity is not provided during world model learning; the world model captures only occupancy dynamics, while agent-specific behaviors are produced by the diffusion model, allowing the two components to play complementary roles. Furthermore, to better exploit the latent representations of both models, we propose a cascaded latent sampling strategy applied at inference time without additional training. This strategy promotes structured exploration across two complementary levels: the scene level, guiding sampling of future occupancies from the world model, and the agent level, driving variation in the diffusion-based motion generation. We formalize this using a flow space determinantal point process (DPP) \cite{kulesza2012determinantal} with a quality-weighted kernel that balances and sample realism, enabling AutoWorld to cover the space of plausible scene evolutions and agent behaviors without modifying training.

In summary, our \textbf{main contributions} are: (1) We propose AutoWorld, a traffic simulation framework that grounds traffic simulation in raw sensor observations through a self-supervised LiDAR world model, addressing the fundamental limitations of trajectory-based approaches. (2) We introduce a motion-aware latent supervision objective that emphasizes dynamic scene elements, enabling self-supervised world model training without semantic labels. (3) We propose a cascaded latent sampling strategy based on a quality-weighted DPP that promotes structured exploration of the joint latent space of both models at inference. (4) Extensive experiments on nuPlan and Waymo demonstrate AutoWorld's competitive generation realism, with larger gains under partial observability. AutoWorld remains reactive to planner-induced ego behaviors and enables controllable and diverse safety-critical scenario generation, while ablations validate its key components.

\section{Related Work}
\vspace{-1mm}
We describe two related areas: (1) Traffic Simulation and (2) World Models in Autonomous Driving.

\textbf{Traffic Simulation.} Traffic simulation models the distribution of multi-agent behaviors in dynamic driving environments. Recently, data-driven approaches have gained significant attention as they can overcome limitations of heuristic-based simulators such as poor adaptability to ego-vehicle behavior. Prior work spans CVAEs \cite{igl2023hierarchical,suo2021trafficsim,xu2022bits}, Transformer-based architectures \cite{girgis2021latent,ngiam2021scene,shi2024mtr++,zhou2023query}, diffusion models \cite{huang2024versatile,jiang2023motiondiffuser,lu2024scenecontrol,pronovost2023scenario,zhong2023language,zhong2023guided}, and more recently autoregressive next-token formulations for joint trajectory generation \cite{wu2024smart,zhang2025trajtok,zhang2025closed,zhou2024behaviorgpt,philion2024trajeglish,peng2025infgen}. Despite architectural differences, these methods learn behavior solely from trajectory abstractions, which discard sensory cues shaping real driving behavior \cite{chen2024womd}. In contrast, AutoWorld grounds behavior generation in raw sensor observations through self-supervised world models, mitigating the information bottleneck of trajectory-only approaches.

\textbf{World Models in Autonomous Driving.} World models represent and forecast the evolution of driving environments. One line of work uses future sensor prediction for end-to-end driving, either jointly optimized with planning \cite{pan2022iso,hu2022model,li2024think2drive,zheng2024occworld} or used for pretraining perception encoders \cite{min2023uniworld,yang2024visual,min2024driveworld}. Another line develops generative neural simulators that produce future video \cite{hu2023gaia,wang2024drivedreamer,wang2024driving}, LiDAR \cite{Weng2021,Weng2022,Zhang2023,liu2025foundationallidarworldmodels}, or multimodal observations \cite{bogdoll2025muvo} for synthesis, counterfactuals, or imagination-based planning. These methods mainly emphasize fidelity or controllability, often decoupled from behavior modeling. Rather than treating the world model as a standalone generator, AutoWorld integrates its predictions into motion generation, providing sensor-grounded scene context for traffic simulation.

\begin{figure*}[t]
    \centering  \vspace{-1mm}
    \includegraphics[width=\columnwidth,  trim=1.5cm 10cm 1.65cm 3.8cm, clip]{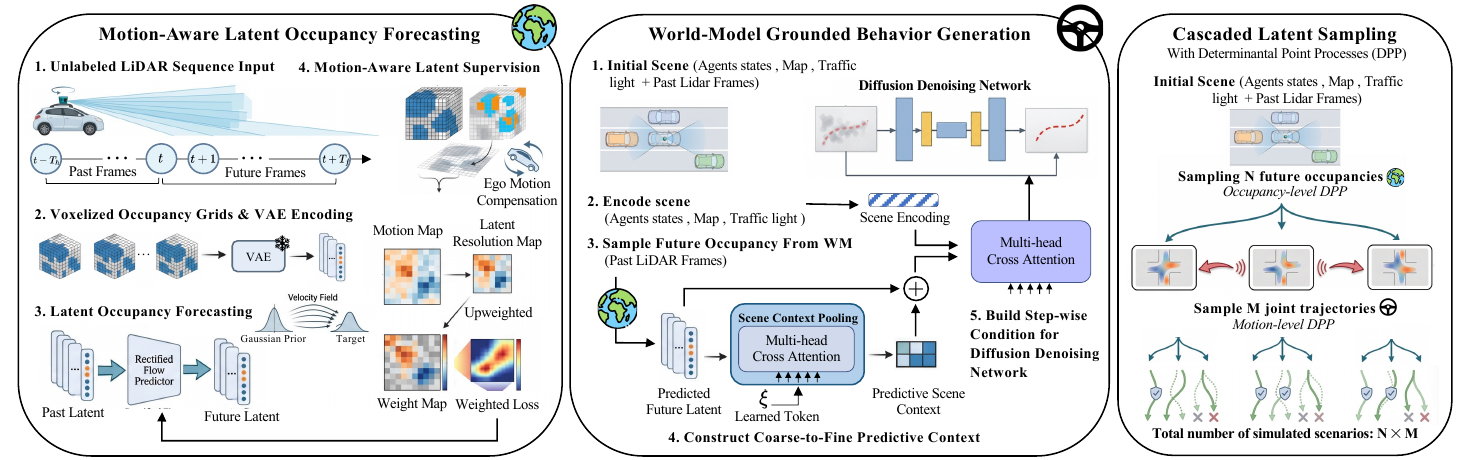}
    \caption{
\textbf{Overview of AutoWorld}. AutoWorld learns latent scene dynamics from unlabeled LiDAR occupancy sequences and uses predicted future occupancies to condition diffusion-based motion generation. At inference, a cascaded sampling strategy further improves use of the learned latent spaces across both scene evolution and agent behavior without additional training.
    } \vspace{-6mm}
    \label{fig:main_fig}
\end{figure*}

\vspace{-1mm}
\section{AutoWorld} 
\vspace{-1mm}
AutoWorld grounds traffic simulation in raw sensor observations by building simulation around a world-modeling formulation. Instead of modeling future agent behavior in isolation, AutoWorld models the joint future distribution of sensor observations and agent states, factorized into a self-supervised LiDAR world model and a diffusion-based motion generation model. This factorization makes world modeling the interface through which raw sensor data informs traffic simulation: the world model learns predictive scene dynamics from high-dimensional unlabeled LiDAR data, while the motion model retains structured trajectory-based inputs for agent-centric behavior generation.

Specifically, we first learn a motion-aware latent occupancy world model in a fully self-supervised manner from unlabeled LiDAR data to forecast future latent occupancies (Sec.~\ref{WM}). From these predictions, we build a coarse-to-fine representation of the future scene to condition the diffusion-based motion generation model. (Sec.~\ref{Behavior_Generation}). Finally, to promote structured exploration of plausible multi-agent behaviors, we introduce a cascaded latent sampling strategy applied at inference time without additional training (Sec.~\ref{Latent_Diversity}). The overall framework is illustrated in Fig.~\ref{fig:main_fig}.

\subsection{Problem Formulation.} 
Given past states of $A$ agents over $T_h$ and contextual information, traffic simulation predicts their future states over $T_f$. At timestep $t$, agent states are $s_t = [s_t^1, \ldots, s_t^A]$, where $s_t^i = (x_t^i, y_t^i, v_t^i, \theta_t^i)$ denotes the 2D position, speed, and yaw of agent $i$. We denote the history by $S_{t-T_h:t} = [s_{t-T_h}, \ldots,s_t]$, with LiDAR observations $O_{t-T_h:t}$ available over the same horizon. Context includes an HD map $V \in \mathbb{R}^{v \times d_v}$ with up to $v$ vector points of dimension $d_v$, and traffic-light states $R \in \mathbb{R}^{r \times d_r}$ for up to $r$ lights with feature dimension $d_r$. Actions are $a_t = [a_t^1, \ldots, a_t^A]$, where $a_t^i = (\dot{v}_t^i, \dot{\theta}_t^i)$ denotes acceleration and yaw rate. Our flow-matching and diffusion timesteps, denoted $k_\mu$ and $k_f$, are distinct from the state-sequence timestep $t$.
\label{Problem_Formulation}


\subsection{Motion-Aware Latent Occupancy Forecasting} \label{WM}
\textbf{Latent Occupancy Forecasting.}
We formulate our world model as a LiDAR-based occupancy predictor operating in the latent space of voxelized occupancies encoded by a VAE. We assume access to additional unlabeled LiDAR sequences $O_{t-T_h: t+T_f}$, which are used to train the latent occupancy predictor. Let $Z_{t-T_h:t}$ denote past latent occupancies and $Z_{t+1:t+T_f}$ the future sequence to be predicted. We build upon \cite{liu2025foundationallidarworldmodels}, which adopts a rectified flow formulation \cite{liu2022flow}. We denote the conditioning information as 
$c^\mu = \{Z_{t-T_h:t}, \tau_{t-T_h:t}^{\text{ego}}\}$,
where $\tau_{t-T_h:t}^{\text{ego}}$ denotes the ego vehicle's past trajectory embedding. The model learns a time-dependent velocity field $\mu_\theta(z_t, t, c^\mu)$ that transports samples from a Gaussian prior toward a target future latent occupancy $\tilde{z}$. During training, we sample flow matching timestep $k_\mu \sim \mathcal{U}(0,1)$ and $z_0 \sim \mathcal{N}(0,I)$, construct the interpolation $z_t = (1-t)z_0 + t\tilde{z}$, and optimize the rectified flow objective in Eq.~\ref{eq:reflow}:
\begin{equation}
\label{eq:reflow}
\mathcal{L}^{\text{world}}(\theta)
=
\mathbb{E}_{k_\mu,\, z_0,\, \tilde{z}}
\left\|
\mu_\theta(z_{k_\mu}, k_\mu, c^\mu) - (\tilde{z} - z_0)
\right\|_2^2 .
\end{equation}

\textbf{Motion-Aware Latent Supervision.} Occupancy grids are spatially imbalanced. Static structures dominate the volume while dynamic objects occupy sparse but behaviorally critical regions. Uniform flow supervision therefore biases learning toward static structure and underweights motion-driven evolution — an issue amplified in our fully unsupervised setting, which lacks the semantic labels that prior methods use to implicitly separate dynamic from static content.

To address this, we introduce a motion-aware supervision strategy derived directly from temporal occupancy changes. Let $Y_t$ and $Y_{t+\Delta}$ denote non-semantic occupancy grids with validity masks $M_t$ and $M_{t+\Delta}$. After compensating for ego motion by warping $Y_t$ into the frame of $t+\Delta$, producing $\tilde{Y}_{t\rightarrow t+\Delta}$, we compute voxel-wise occupancy transitions:
\begin{equation*}
C_{t,\Delta}(v) = \mathbf{1}[Y_{t+\Delta}(v) \neq \tilde{Y}_{t\rightarrow t+\Delta}(v)]
\cdot \tilde{M}_{t\rightarrow t+\Delta}(v)
\cdot M_{t+\Delta}(v),
\end{equation*}
where $\tilde{M}_{t\rightarrow t+\Delta}$ is the validity mask $M_t$ warped to the frame of $t+\Delta$. We downsample this motion map to the latent resolution to obtain $\bar{C}_{t,\Delta} \in \mathbb{R}^{H \times W}$ and construct a latent weight map $W_t(i,j) = 1 + \lambda \bar{C}_{t,\Delta}(i,j)$,
where $\lambda > 0$ controls the strength of motion emphasis. The resulting motion-aware rectified flow objective is
\begin{equation*}
\mathcal{L}^{\text{world}}(\theta)
=
\mathbb{E}_{k_\mu, z_0, \tilde{z}}
\sum_{i,j}
W_t(i,j)
\left\|
\mu_\theta(z_{k_\mu}, {k_\mu}, c^\mu)_{i,j,:}
-
(\tilde{z}-z_0)_{i,j,:}
\right\|_2^2 ,
\end{equation*}
where $W_t$ is normalized to unit spatial mean. This formulation preserves the standard rectified flow objective while biasing gradient magnitude toward temporally changing regions.

\subsection{World-Model-Grounded Traffic Simulation} \label{Behavior_Generation}
We formulate traffic simulation as conditional diffusion over joint future agent trajectories in the action space. Let $\tau_0 = a^{1:A}_{t+1:t+T_f}$ denote the clean future action trajectory over the prediction horizon $T_f$. The diffusion model generates trajectories by initializing from Gaussian noise $\tau_{K_f} \sim \mathcal{N}(0, I)$ and progressively reversing a predefined noising process over $K_f$ steps. This reverse process defines a conditional Markov chain $p_\psi(\tau_{k_f-1} \mid \tau_{k_f}, c^f)$, where $c^f$ denotes the conditioning information. At each denoising step $k_f = K_f, \ldots, 1$, the transition is:
\begin{equation*}
p_\psi\left(\tau_{{k_f}-1} \mid \tau_{k_f}, c^{f}\right)
:=
\mathcal{N}\left(
\tau_{k_f-1};
f_\psi\left(\tau_{k_f}, k_f, c^f\right),
\Sigma_{k_f}
\right),
\end{equation*}
where $f_\psi$ and $\Sigma_{k_f}$ are the reverse-transition mean predictor and variance schedule, respectively.

\textbf{World-model Conditioning.} Given the world-model rollout $\hat{Z}_{t+1:t+T_f}=[\hat{z}_{t+1}, \ldots,\hat{z}_{t+T_f}]$, we construct a compact predictive scene context by temporally subsampling the predicted latent occupancies and aggregating them through multi-head cross-attention pooling (``MHCA'') with a learned token, defined as $g=\operatorname{MHCA}(\mathrm{Q}=\xi, \mathrm{K}=\mathrm{V}=\phi_1(\left\{\hat{z}_{t+e \delta}\right\}_{e=1}^{\left\lfloor T_f / \delta\right\rfloor}))$, where $\phi_1(\cdot)$ indicates a learnable linear projection. The resulting context $g$ summarizes the forecasted occupancies, where each latent occupancy encodes a predicted scene configuration without explicit agent identities.

Next, we condition the diffusion denoiser on both local and global predictive cues, forming a coarse-to-fine guidance mechanism for multi-agent behavior generation. Specifically, for each future timestep $t+r$, where $r=1,\ldots,T_f$, we combine the corresponding per-step latent occupancy $\hat{z}_{t+r}$ with the global predictive scene context $g$. As such, the timestep-specific diffusion condition is defined as $c_{t+r}^f=\mathrm{MHCA}\left(\mathrm{Q}=h, \mathrm{~K}=\mathrm{V}=\phi_2\left(\left[\hat{z}_{t+r} \oplus g\right]\right)\right)$, where $h$ denotes the scene encoding obtained from a query-centric Transformer~\cite{shi2024mtr++}. The $\oplus$ indicates concatenation along the token dimension, and $\phi_2(\cdot)$ maps the concatenated latent occupancies into a shared space. 
The full diffusion condition is therefore $c^f=\{c^f_{t+r}\}_{r=1}^{T_f}$, which combines historical scene context with predicted future occupancies. This allows the diffusion model to generate agent-specific behaviors grounded in past observations while being guided by the world-model.

\subsection{Cascaded Latent Sampling} \label{Latent_Diversity}
To better exploit the learned latent spaces, we introduce a cascaded latent sampling strategy that applies Determinantal Point Processes (DPPs) \cite{kulesza2012determinantal,morshed2025diverseflow} at both the world-modeling and behavior-generation stages. At inference, this mechanism promotes structured exploration by discouraging similar latent samples, yielding diverse yet plausible future scene evolutions and agent behaviors.

\textbf{Determinantal Point Processes.} Let $\mathcal X = \{\hat{x}^{(i)}\}_{i=1}^{K_s}$ be the ground set of generated candidates, and let $\hat{x}^i\in\mathbb R^d$. Let $\kappa: \mathbb{R}^d \times \mathbb{R}^d \rightarrow \mathbb{R}$ be a symmetric positive semidefinite kernel (e.g. cosine similarity), and define the kernel matrix $\Lambda  \in \mathbb{R}^{{K_s} \times {K_s}}$ by $\Lambda_{i j}=\kappa(\hat{x}^{(i)}, \hat{x}^{(j)})$. Following the standard L-ensemble DPP \cite{kulesza2012determinantal} formulation on a finite ground set, we define a DPP-based probability mass for the $K_s$ jointly generated samples, i.e., treating $\mathcal X$ as the ground set, we evaluate the DPP probability of selecting the full subset as:
$\mathcal P_{\kappa}(\mathcal X)=\frac{\operatorname{det}(\Lambda)}{\operatorname{det}(\Lambda+I)}$.
$\mathcal{P}_\kappa$ increases when samples are diverse and approximately linearly independent, and decreases as redundancy increases, becoming 0 when $\mathcal X$ contains any exact duplicates under $\kappa$ (equivalently, when $\Lambda$ is rank-deficient). This provides a principled continuous measure of diversity. However, maximizing diversity alone may push samples away from the data manifold. To balance diversity and plausibility, we incorporate quality-aware modulation \cite{morshed2025diverseflow} by assigning each sample a quality weight $q^{(i)}\in(0,1]$ and defining $\kappa_q(\hat x^{(i)},\hat x^{(j)})=q^{(i)}\kappa(\hat x^{(i)},\hat x^{(j)})q^{(j)}$. This yields the quality-balanced kernel $\Lambda^q=\operatorname{diag}(q)\Lambda\operatorname{diag}(q)$ and the quality-aware DPP probability $\mathcal P_\kappa^q(\mathcal X)=\frac{\operatorname{det}(\Lambda^q)}{\operatorname{det}(\Lambda^q+I)}$, whose maximization balances diversity and sample quality.

\textbf{DPP-Guided Sampling.} To jointly generate ${K_s}$ diverse candidates $\left\{\hat{x}^{(i)}\right\}_{i=1}^{K_s}$, we evaluate $\mathcal{P}_{\kappa}^q$ over the predicted target outputs and incorporate a repulsive guidance term proportional to $\nabla \log \mathcal{P}_{\kappa}^q$ into the flow ODE during sampling. This term encourages candidate trajectories to spread across distinct modes of the predictive distribution while maintaining high-quality solutions. The guidance is applied only at inference time and remains independent of the underlying training objective.

We apply diversity-guided sampling at both world modeling and behavior generation through a two-stage cascaded inference procedure. First, we sample $N$ future latent occupancies from the world model $\mathcal X^\text{world}\doteq \{\hat{Z}_{t+1: t+T_f}^{(i)}\}_{i=1}^N$, where each latent rollout represents a distinct plausible future occupancy development. We quantify rollout quality via the Mahalanobis distance of the one-step latent estimate to a Gaussian fit of the training data. Let $\mathcal P^{q,\text{world}} \doteq \mathcal P_{\kappa}^q(\mathcal X^\text{world})$. The diversity-guided world-model sampling dynamics follow the continuous-time flow formulation given by
\begin{equation*}
\hat{z}^{(i)}_{k_\mu -1 }=\mu_\theta(\hat{z}_{k_{\mu}}^{(i)}, {k_{\mu}}, c^\mu)-\gamma_{\mu}({k_{\mu}}) \nabla_{z_{k_{\mu}}^{(i)}} \log \mathcal{P}_{k_{\mu}}^{q, \text{world}}, \quad i=1, \ldots, N,
\end{equation*}
where $\gamma(.)$ controls the strength of diversification. Next, conditioned on the step-aware representation derived from each sampled latent representation, we generate $M$ agent behavior trajectories using our conditional diffusion model, $\{\tau^{(i, j)}\}_{j=1}^M$. Since behavior generation proceeds via discrete reverse diffusion steps, we apply diversity guidance as a gradient shift to the reverse-step mean at each timestep $k_f$. Let $\mathcal P^{q,\text{motion}} \doteq \mathcal P_{\kappa_q}(\mathcal X^\text{motion})$. The diversity-guided sampling dynamics for the behavior trajectory simulation, similar to the dynamics for the world model, becomes
\begin{equation*}
\tau_{k_{f}-1}^{(i,j)} =
f_\psi(\tau_{k_f}^{(i,j)}, k_f, c^{f(i)})
- \gamma_f(k_{f})\,
\nabla_{\tau_{k_f}^{(i,j)}} \log \mathcal{P}_{k_f}^{q,\text{motion}}
\quad j=1,\ldots,M,
\end{equation*}
where $c^{f(i)}$ denotes the conditioning signal derived from the $i$-th world-model rollout. For notational simplicity, we omit the stochastic noise term from the variance schedule. Trajectory embeddings are obtained from the scene encoder over agent states. At the behavior level, quality weights encode trajectory-space realism constraints, including collisions, kinematic infeasibility, and off-road violations \cite{gulino2023waymax}, preventing diversity from favoring physically invalid rollouts. Overall, cascaded diversity-guided inference produces $N \times M$ diverse and plausible rollouts.

\begin{table}[t]
\centering
\caption{\textbf{Simulation realism and physical plausibility on nuPlan.} Realism measures the distributional deviation from logged driving statistics (longitudinal/lateral acceleration magnitude and jerk); other metrics assess physical plausibility. D-dir. Viol. denotes driving-direction violation.} \vspace{-2mm}
\label{tab:nuplan}
\resizebox{0.9\columnwidth}{!}{%
\begin{tabular}{lcccccc}
\Xhline{0.75pt}
\textbf{Model}  
& \textbf{Realism}$(\downarrow)$ 
&  \textbf{Collision}$(\downarrow)$
&  \textbf{Off-road Time}$(\downarrow)$ 
&  \textbf{D-dir. Viol.}$(\downarrow)$
& \hspace{0.2mm}  \textbf{Kinematic}$(\uparrow)$ \\
\Xhline{0.5pt}

CTG \cite{zhong2023guided}  & 0.584 & 27.79 & 25.22 & 16.59 & 79.37 \\

GUMP \cite{hu2024solving}  & 0.472 & 23.81 &  30.57 & 20.52 & 73.48 \\

SceneD. \cite{qian2024nuscenes}  & 0.355 & 36.93 & \underline{14.02} & \underline{12.59} & 89.35 \\

D. Policy \cite{chi2025diffusion}  & 0.604 & 33.68 & 19.84 & 21.00 & 88.29 \\

Nexus \cite{zhou2025decoupled}  & \underline{0.318} & 27.19 & 16.03 & 15.64 & 84.41 \\

OMEGA \cite{li2026optimization}  & 0.348 & \underline{23.91} & 14.68 & 19.48 & \underline{90.63} \\


\rowcolor{gray!15}
\textbf{AutoWorld (Ours)} & \textbf{0.266} & \textbf{8.14} & \textbf{5.93} & \textbf{7.31} & \textbf{94.89} \\

\Xhline{0.75pt}
\end{tabular}%
} \vspace{-4mm}
\end{table}

\vspace{-1mm}
\section{Experiments}
\vspace{-1mm}

Our experiments focus on five key questions: \textbf{Q1:} Does AutoWorld generate \textit{realistic} traffic behavior that matches real-world human driving? (Sec~\ref{Q1}) \textbf{Q2:} What benefits does sensor-grounded world modeling provide over trajectory-based simulation? (Sec~\ref{Q2}) \textbf{Q3:} Can AutoWorld exhibit realistic \textit{reactiveness} to planner-controlled ego behavior? (Sec~\ref{Q3}) \textbf{Q4:} Beyond realism and reactiveness, can AutoWorld generate \textit{diverse} and \textit{controllable} behaviors? (Sec~\ref{Q4}) \textbf{Q5:} How do AutoWorld’s key design choices contribute to its overall performance? (Sec~\ref{Q5})

\textbf{Experimental Design.} AutoWorld is trained on \textbf{nuPlan}~\cite{karnchanachari2024towards} and \textbf{Waymo}~\cite{sun2020scalability}. We follow Occ3D-Waymo~\cite{tian2023occ3d} and NuPlan-Occ~\cite{li2026scaling} to construct voxel-wise occupancy grids for world-model training. AutoWorld predicts 8 seconds ahead from 2 seconds of history on nuPlan and 1 second on Waymo in closed loop, following the Open Sim Agents benchmark. We use receding-horizon closed-loop simulation at 1 Hz for both occupancy and motion. The world model is rolled out autoregressively without access to future ground-truth occupancy. Notably, Waymo provides LiDAR data \textit{only for the observation horizon} in the test split. For DPP sampling, we use a cosine-similarity kernel at both levels with a power-cosine schedule~\cite{gao2023masked}, yielding weaker guidance early and stronger guidance toward the end of sampling. We follow each benchmark's realism criteria. On nuPlan, we measure the Wasserstein distance between simulated and logged driving-profile distributions, together with log-independent physical-plausibility metrics, including collision rate, off-road time (fraction of simulation time spent off-road), driving-direction violation rate, and kinematic feasibility rate. On Waymo, we follow the Open Sim Agents protocol and report likelihood-based realism metrics covering kinematic behavior, agent interactions, and map adherence, their aggregate Realism Meta Metric (RMM), and minADE. Thus, realism metrics are benchmark-specific and not directly comparable across datasets.

\subsection{Comparison to State of the Art} \label{Q1}
\begin{figure}[t]
    \centering \vspace{-2mm}
    \resizebox{0.85\linewidth}{!}{%
        \includegraphics[
            trim={2cm 8.3cm 1.5cm 4.8cm},
            clip
        ]{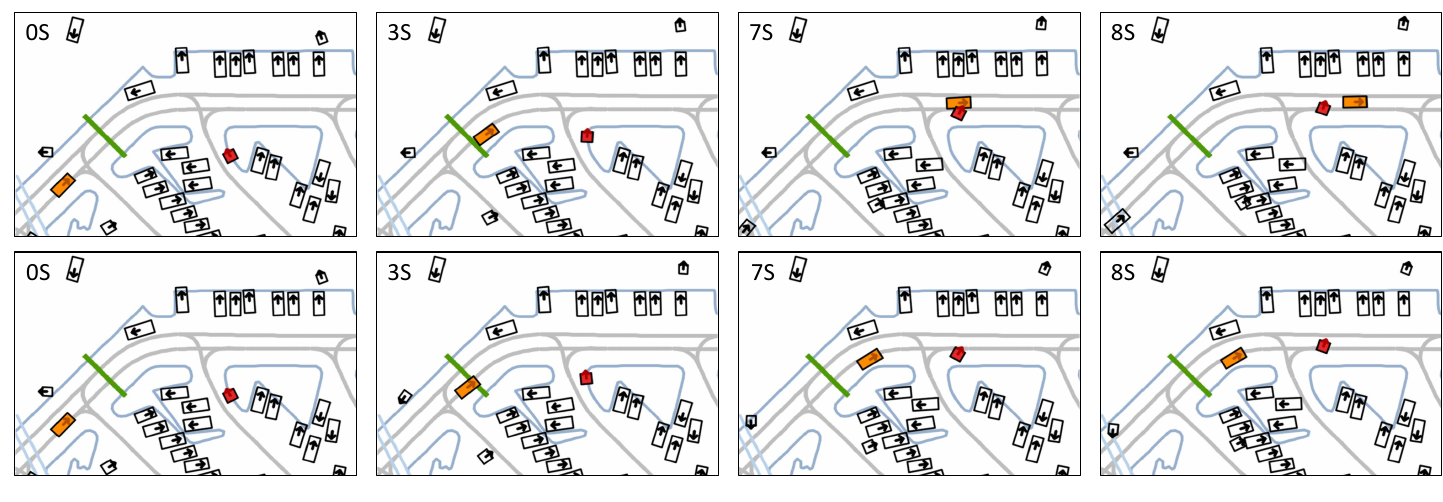}
    } \vspace{-3.5mm}
\caption{\textbf{Qualitative rollout comparison.} 
Trajectory-baseline (top) leads to a collision between the vehicle ({\color{orange}\rule{1.2ex}{1.2ex}}) and pedestrian ({\color{red}\rule{1.2ex}{1.2ex}}), while AutoWorld (bottom) maintains a realistic yielding interaction.}
    \label{fig:rollout_vis_m} \vspace{-2mm}
\end{figure}

We begin by evaluating AutoWorld on nuPlan for distributional realism against logged human driving and physical plausibility, comparing against recent and reproduced traffic-generation methods in Table~\ref{tab:nuplan}. AutoWorld achieves state-of-the-art performance across all reported metrics, improving both realism and physical plausibility over prior diffusion-based methods. Notably, it reduces collision by nearly 16 percentage points over the next-best method (8.14 vs.~23.91), while reducing off-road time (5.93 vs.~14.02 for the next best), showing that the collision reduction does not come at the expense of road adherence. We further evaluate AutoWorld on the Waymo Open Sim Agents benchmark, a complementary evaluation focused on matching the distribution of human driving behavior. Table~\ref{tab:wosac_leaderboard} groups methods by training paradigm, following \cite{DecompGAIL2026}. AutoWorld is trained solely with imitation objectives and uses no reward-based post-training. Among imitation-trained simulators, AutoWorld achieves the highest realism (RMM 0.7865) and matches RL-fine-tuned simulators without reward-based fine-tuning. See Figure~\ref{fig:rollout_vis_m} for qualitative examples.

\begin{table}[t]
\centering 
\caption{\textbf{Simulation realism under the Waymo Open Sim Agents evaluation protocol.} RMM (Realism Meta Metric) is a weighted aggregation of kinematic, interactive, and map-based metrics and serves as the primary ranking metric.} 
\vspace{-2mm}
\label{tab:wosac_leaderboard}
\resizebox{\columnwidth}{!}{%
\begin{tabular}{llccccc}
\Xhline{0.75pt}
\textbf{Training}
& \textbf{Method}
& \textbf{RMM}$(\uparrow)$ 
& \hspace{0.2mm}\textbf{Kinematic}$(\uparrow)$
& \hspace{0.2mm}\textbf{Interactive}$(\uparrow)$ 
& \hspace{0.2mm}\textbf{Map-based}$(\uparrow)$
& \hspace{0.2mm}\textbf{minADE}$(\downarrow)$ \\

\Xhline{0.5pt}

\multirow{4}{*}{%
\begin{tabular}[c]{@{}c@{}}
\textbf{RL Post-}\\
\textbf{training}
\end{tabular}}
& RLFTSim \cite{RlftsimEhsan} 
& 0.7857 & 0.4927 & 0.8129 & 0.9183 & 1.3252 \\

& SMART-R1 \cite{pei2025advancing}  
& 0.7858 & 0.4944 & 0.8110 & 0.9201 & 1.2885 \\

& DecompGAIL \cite{DecompGAIL2026} 
& 0.7864 & 0.4919 & 0.8152 & 0.9176 & 1.4209 \\ 

& LLM2AD \cite{wang2025llm} 
& 0.7779 & 0.4846 & 0.8048 & 0.9109 & 1.2827 \\

\Xhline{0.5pt}

\multirow{5}{*}{%
\begin{tabular}[c]{@{}c@{}}
\textbf{Imitation}\\
\textbf{Learning}
\end{tabular}}
& SMART \cite{wu2024smart}  
& 0.7814 & 0.4854 & 0.8089 & 0.9153 & 1.3931 \\

& CAT-K \cite{zhang2025closed}  
& 0.7846 & \textbf{0.4931} & 0.8106 & 0.9177 & \underline{1.3065} \\

& TrajTok \cite{zhang2025trajtok}  
& \underline{0.7852} & \underline{0.4887} & \underline{0.8116} & \textbf{0.9207} & 1.3179 \\ 

& SceneStreamer \cite{peng2025infgen} 
& 0.7731 & 0.4493 & 0.8084 & 0.9127 & 1.4252 \\

& \cellcolor{gray!15}\textbf{AutoWorld (Ours)}
& \cellcolor{gray!15}\textbf{0.7865} 
& \cellcolor{gray!15}\textbf{0.4931} 
& \cellcolor{gray!15}\textbf{0.8143} 
& \cellcolor{gray!15}\underline{0.9185} 
& \cellcolor{gray!15}\textbf{1.3051} \\

\Xhline{0.75pt}
\end{tabular}%
}
\vspace{-1mm}
\end{table}

\subsection{Benefit of Sensor-Grounded Simulation} \label{Q2}
To study when raw sensor data provides information beyond lossy trajectory abstractions, we consider occlusion as a representative test case \cite{chen2024womd}, following the protocol of \cite{lange2024scene}. Both Waymo and nuPlan provide
\begin{wrapfigure}{r}{0.57\linewidth}
    \centering \vspace{-2mm}
    \resizebox{0.57\columnwidth}{!}{%
        \includegraphics{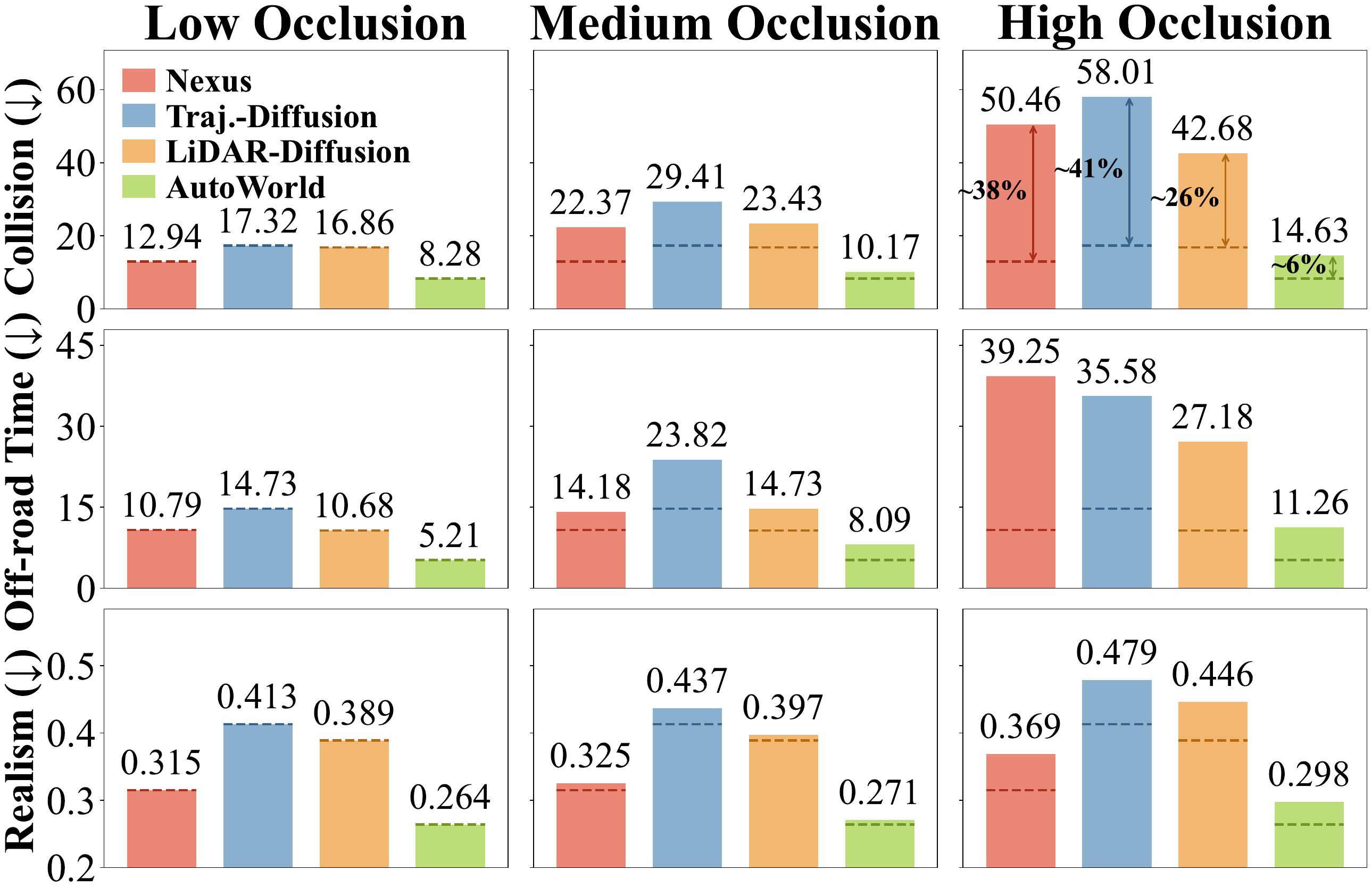}
    } \vspace{-7mm}
\caption{\textbf{Simulation performance across occlusion levels} Dashed lines denote low-occlusion performance as reference for degradation at higher occlusion levels.} \vspace{-3mm}
    \label{fig:your_label}
\end{wrapfigure}
high-quality agent tracks from offline auto-labeling pipelines \cite{ettinger2021large,karnchanachari2024towards}, allowing trajectories to persist even when agents are unobservable by ego sensors. We estimate occlusion using BEV line-of-sight ray casting from the ego vehicle and count, for each valid agent, timesteps where it is geometrically occluded but present in the trajectory labels. We compute a scene-level occlusion score and form equal-sized \textbf{Low-}, \textbf{Moderate-}, and \textbf{High-Occlusion} subsets with comparable agent densities. We compare AutoWorld against Nexus \cite{zhou2025decoupled}, a recent diffusion-based SOTA with public code and checkpoints, and two controlled diffusion baselines.Traj.-Diffusion predicts joint multi-agent actions from trajectory context only, while LiDAR-Diffusion additionally incorporates LiDAR observations, fused with HD-map features following DeTra \cite{casas2024detra}. Figure~\ref{fig:your_label} reveals a clear sensitivity of trajectory-based simulation to partial observability on nuPlan. As occlusion increases, Nexus and Traj.-Diffusion degrade substantially in realism and physical plausibility. Directly incorporating LiDAR alleviates this degradation, reducing collision rate from 58.01$\%$ to 42.68$\%$ over Traj.-Diffusion under high occlusion. AutoWorld further reduces collisions to 14.63$\%$---\textit{about 36 percentage points below Nexus}---while achieving a realism score of 0.298 and substantially less degradation across occlusion levels. These results indicate that sensory information is valuable beyond trajectory abstractions, and that \textit{explicitly modeling sensory-world dynamics} is more effective than direct sensor conditioning.

\subsection{Closed-Loop Reactiveness} \label{Q3}
\begin{wraptable}{r}{0.45\textwidth}
\centering \vspace{-3mm}
\caption{\textbf{Closed-loop reactiveness under planner-controlled ego behavior on nuPlan.} The ego vehicle is controlled by an external planner, while all other agents are generated by the simulator. E2A/A2A denote ego-to-agent and agent-to-agent collision rates, respectively.}
\vspace{-2mm}
\label{tab:reactiveness}
\resizebox{\linewidth}{!}{%
\begin{tabular}{lcccc}
\Xhline{0.75pt}

\multirow{2}{*}{\textbf{Model}}
& \multicolumn{2}{c}{\textbf{Collision}$(\downarrow)$}
& \multirow{2}{*}{\begin{tabular}[c]{@{}c@{}}
    \textbf{Off-road}\\
    \textbf{Time}$(\downarrow)$
  \end{tabular}}
& \multirow{2}{*}{\begin{tabular}[c]{@{}c@{}}
    \textbf{Kinem-}\\
    \textbf{atic}$(\uparrow)$
  \end{tabular}} \\
\cmidrule(lr){2-3}
& \textbf{E2A} & \textbf{A2A} & & \\

\Xhline{0.5pt}

\rowcolor{gray!20}
\multicolumn{5}{l}{\textbf{\textit{Rule-based Planner}}} \\

Nexus     & 34.51 & 22.02 & 17.93 & 81.59 \\
AutoWorld & 10.48 &  8.93 & 12.69 & 89.38 \\
\Xhline{0.5pt}

\rowcolor{gray!20}
\multicolumn{5}{l}{\textbf{\textit{Learning-based Planner \cite{chi2025diffusion}}}} \\

Nexus     & 38.29 & 27.24 & 21.83 & 74.46 \\
AutoWorld & 13.05 & 11.32 & 13.25 & 88.05 \\

\Xhline{0.75pt}
\end{tabular}%
}
\vspace{-3mm}
\end{wraptable}
The evaluations in Sec.~\ref{Q1} let the behavior model control both the ego and surrounding agents. To test reactiveness beyond this self-controlled setting, we instead use an independent ego planner while the simulator controls surrounding agents. We consider both rule-based and learning-based planners. As shown in Tab.~\ref{tab:reactiveness}, AutoWorld remains robust under externally controlled ego behavior, consistently outperforming Nexus across both planners. Compared with its self-controlled setting, AutoWorld retains high kinematic feasibility (94.89$\rightarrow$89.38/88.05) while maintaining substantially lower collision and off-road rates than Nexus. This demonstrates that AutoWorld can respond safely and feasibly when ego behavior is generated independently of the simulator.

\subsection{Diverse and Controllable Simulation} \label{Q4}
A useful simulator should support both \textit{diverse} and \textit{controllable} generation. We probe these properties through safety-critical scenario generation, explicitly steering the simulator toward challenging interactions with a downstream planner. Diffusion models naturally support inference-time guidance, unlike the less explicit control of next-token prediction methods. We therefore investigate whether AutoWorld can generate targeted safety-critical interactions while preserving its simulation realism. Following Safe-Sim \cite{chang2024safe}, we evaluate on nuPlan with a planner in the loop and apply the same safety-critical guidance objective to AutoWorld and Nexus. The objective steers a designated adversarial agent toward collision with the planner-controlled ego while discouraging other-agent collisions and off-road behavior. As shown in Table~\ref{tab:adv_gen}, AutoWorld achieves a higher adversarial collision rate than Nexus (41.84 vs.\ 36.12), while substantially reducing unintended collisions (7.03 vs.\ 23.68) and off-road time (3.60 vs.\ 13.92). We further assess diversity through the variance of collision properties across generated outcomes, following Safe-Sim. DPP increases diversity across all three attributes, particularly relative speed (1.27$\rightarrow$4.58) and collision point (1.53$\rightarrow$6.19), while improving both controllability and realism.

\colorlet{targetHigh}{orange!22}
\colorlet{targetMid}{orange!9}
\colorlet{targetLow}{orange!5}

\colorlet{nonTargetBest}{teal!22}
\colorlet{nonTargetMid}{teal!19}
\colorlet{nonTargetWorst}{teal!5}

\colorlet{speedHigh}{blue!22}
\colorlet{speedMid}{blue!6}
\colorlet{speedLow}{blue!5}

\colorlet{pointHigh}{Lavender!22}
\colorlet{pointMid}{Lavender!6}
\colorlet{pointLow}{Lavender!5}
\begin{figure*}[h]
\centering
\caption{\textbf{Controllable and diverse safety-critical scenario generation on nuPlan.} Target Collision denotes ego--adversarial-agent collision, while Non-Target Collision denotes all other collisions. Off-road Time is measured for the adversarial agent, and Relative Speed is the ego--adversarial speed at collision. AutoWorld achieves a higher target collision rate while substantially reducing non-target collisions; DPP sampling further increases collision diversity.}
\vspace{-2mm}
\label{tab:adv_gen}

\resizebox{0.85\textwidth}{!}{%
\begin{tabular}{lccccccc}
\Xhline{0.75pt}

& \multicolumn{4}{c}{\textbf{Collision Metrics}}
& \multicolumn{3}{c}{\textbf{Collision Diversity (Var)$(\uparrow)$}} \\
\cmidrule[0.5pt](r){2-5}
\cmidrule[0.5pt](l){6-8}

\textbf{Model}
& \textbf{\begin{tabular}[c]{@{}c@{}}Target\\ Collision$(\uparrow)$\end{tabular}}
& \textbf{\begin{tabular}[c]{@{}c@{}}Non-Target\\ Collision$(\downarrow)$\end{tabular}}
& \textbf{\begin{tabular}[c]{@{}c@{}}Off-road\\ Time$(\downarrow)$\end{tabular}}
& \textbf{Realism}$(\downarrow)$
& \textbf{\begin{tabular}[c]{@{}c@{}}Angle\\ (rad)\end{tabular}}
& \textbf{\begin{tabular}[c]{@{}c@{}}Relative\\ Speed (m/s)\end{tabular}}
& \textbf{\begin{tabular}[c]{@{}c@{}}Point\\ (m)\end{tabular}} \\
\Xhline{0.5pt}

Nexus
& \cellcolor{targetMid}36.12
& \cellcolor{nonTargetWorst}23.68
& 13.92
& 0.437
& 2.41
& \cellcolor{speedLow}1.04
& \cellcolor{pointMid}1.86 \\

AutoWorld-DPP
& \cellcolor{targetLow}34.56
& \cellcolor{nonTargetMid}9.42
& 5.28
& 0.289
& 1.93
& \cellcolor{speedMid}1.27
& \cellcolor{pointLow}1.53 \\

AutoWorld+DPP
& \cellcolor{targetHigh}\textbf{41.84}
& \cellcolor{nonTargetBest}\textbf{7.03}
& 3.60
& \textbf{0.274}
& 3.77
& \cellcolor{speedHigh}\textbf{4.58}
& \cellcolor{pointHigh}\textbf{6.19} \\

\Xhline{0.75pt}
\end{tabular} \vspace{-5mm}
}
\end{figure*}

\vspace{-1mm}
\subsection{Ablation Studies} \label{Q5}

\textbf{Motion-Aware Latent Supervision.} We study whether motion-aware latent supervision improves dynamic scene forecasting without semantic labels and whether these gains transfer to traffic simulation. We compare semantic supervision (Sem-RF), uniform non-semantic supervision (NSem-RF), and our motion-aware non-semantic supervision (NSem-MA-RF) on Waymo. Table~\ref{tab:ablation-ma} shows that uniform supervision performs reasonably on static regions but underperforms on dynamic ones, indicating a bias toward dominant static occupancy. Motion-aware supervision mitigates this bias, raising dynamic IoU to 0.71 while preserving static IoU$_\text{st}$ (0.87), and improves rollout fidelity (FVD: 51 vs. 35). These world-model gains translate to simulation, improving both realism and trajectory accuracy over uniform supervision (RMM: 0.7547 to 0.7618), nearly matching the semantic baseline in RMM. Overall, motion-aware supervision provides a label-free alternative that strengthens dynamic forecasting and closes much of the gap to semantic supervision.

\begin{table*}[t]
\centering 
\begin{minipage}{0.47\textwidth}
\centering 
\caption{\textbf{Effect of motion-aware latent supervision on Waymo.} The MA-RF objective improves world modeling and translates into gains in traffic simulation without requiring semantic labels. Except for FVD$(10^{-3})$ and minADE, higher values are better.} \vspace{-2mm}
\label{tab:ablation-ma}
\resizebox{\linewidth}{!}{%
\begin{tabular}{lccccc}
\hline
& \multicolumn{3}{c}{\textbf{World model}} 
& \multicolumn{2}{c}{\textbf{Traffic simulation}} \\  

\cmidrule[0.5pt](r){2-4}
\cmidrule[0.5pt](l){5-6}

\textbf{Model} 
& \textbf{FVD}
& \textbf{IoU$_{\text{st}}$}
& \textbf{IoU$_{\text{dy}}$}
& \textbf{RMM} 
& \textbf{minADE}  \\ 
\hline
Sem-RF       
& \cellcolor{fvdBest}23 
& 0.89 
& \cellcolor{ioudyHigh}0.73 
& \cellcolor{rmmHigh}0.7651 
& 1.4081 \\

NSem-RF     
& \cellcolor{fvdWorst}51 
& 0.85 
& \cellcolor{ioudyLow}0.58 
& \cellcolor{rmmLow}0.7547 
& 1.4826 \\

NSem-MA-RF  
& \cellcolor{fvdMid}35 
& 0.87 
& \cellcolor{ioudyMid}0.71 
& \cellcolor{rmmMid}0.7618 
& 1.4334 \\ 
\hline
\end{tabular}
}
\end{minipage}
\hspace{1mm}
\begin{minipage}{0.50\textwidth}
\centering 
\caption{\textbf{Ablation study on world-model grounding and DPP sampling strategies on Waymo}. The design choices provide complementary gains, with full AutoWorld achieving the best performance. Higher values are better.} \vspace{-2mm}
\label{tab:dpp_ablation}
\resizebox{\linewidth}{!}{%
\begin{tabular}{lccccc}
\hline
\textbf{Model}& \textbf{RMM}  
& \textbf{Kinematic} 
& \textbf{Interactive}  
& \textbf{Map-based} \\ 
\hline
AutoWorld 
& \cellcolor{ablRmmHigh}0.7746 
& 0.4878 & 0.7968 & 0.9099 \\

w/o mo-DPP   
& \cellcolor{ablRmmMoDpp}0.7648 
& 0.4846 & 0.7926 & 0.8892  \\

w/o occ-DPP   
& \cellcolor{ablRmmOccDpp}0.7563 
& 0.4833 & 0.7758 & 0.8871 \\

w/o DPP   
& \cellcolor{ablRmmNoDpp}0.7597 
& 0.4840 & 0.7839 & 0.8861  \\

w/o WM    
& \cellcolor{ablRmmLow}0.7472 
& 0.4749 & 0.7741 & 0.8682  \\ 
\hline
\end{tabular}
}
\end{minipage} \vspace{-2mm}

\end{table*}


\vspace{2mm}
\textbf{World-model Grounding and Cascaded Sampling.}
Table~\ref{tab:dpp_ablation} shows that world-model grounding is the primary contributor to AutoWorld’s performance. Removing the world model (WM) leads to the largest degradation across all metric groups, reducing RMM from $0.7746$ to $0.7472$ while also lowering kinematic, interactive, and map-based scores. Diversity sampling further provides consistent gains on top of world-model grounding. Removing motion-level DPP (mo-DPP) reduces interactive realism ($0.7968 \rightarrow 0.7926$), whereas removing occupancy-level DPP (occ-DPP) causes a larger drop in overall realism ($0.7746 \rightarrow 0.7563$). Notably, removing both DPP modules still outperforms the variant without the world model, indicating that predictive scene grounding contributes more substantially than sampling diversity alone. Combining world-model grounding with cascaded DPP sampling yields the best performance. Ablations use data from the Waymo interactive validation split.

\begin{wrapfigure}{r}{0.35\linewidth}
  \vspace{-2mm}
  \centering
  \includegraphics[width=1\linewidth]{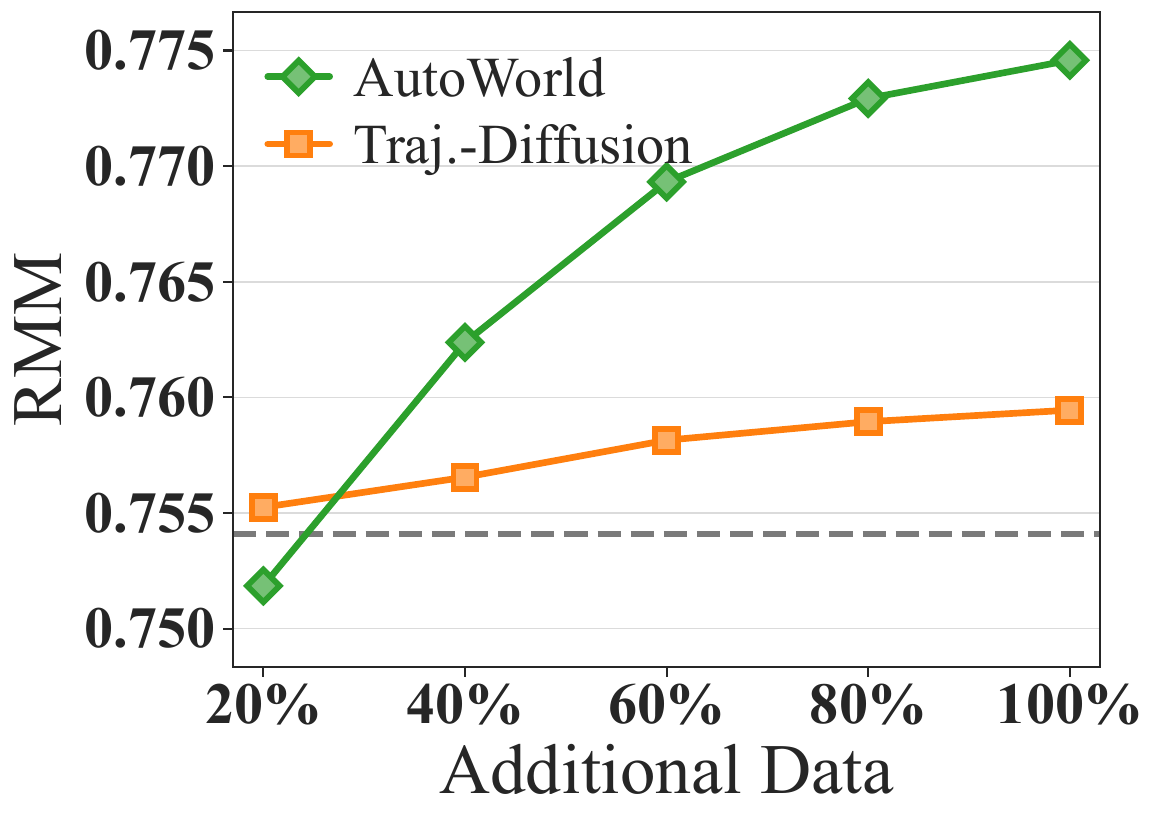}\vspace{-3mm}
\caption{\textbf{Scaling with different data modalities on Waymo.} Dashed line denotes Traj.-Diffusion without additional data.}
\vspace{-3mm}
  \label{fig:scale}
\end{wrapfigure}

\vspace{2mm}
\textbf{Scaling with Additional Data.} Finally, we investigate whether sensor-grounded world modeling provides a more scalable path for leveraging additional data. We compare against the trajectory-diffusion baseline from Sec.~\ref{Q2}, augmented to match AutoWorld's parameter count. The baseline converts additional data into trajectory supervision, while AutoWorld uses the corresponding LiDAR observations to incrementally train its world model. As shown in Fig.~\ref{fig:scale}, AutoWorld improves more consistently with data, while the trajectory baseline quickly saturates. This suggests that world modeling better exploits sensor data by transforming raw observations into predictive scene context. While AutoWorld's gains also diminish at larger scales, stronger world models and tighter integration of scene dynamics with behavior generation may enable further scaling, pointing to future directions.

\vspace{-2mm}
\section{Conclusion}
\vspace{-1mm}
We introduced \textbf{AutoWorld}, a world-model-grounded framework for traffic simulation that uses raw LiDAR observations to complement structured trajectory representations. Across nuPlan and Waymo, AutoWorld improves simulation realism and physical plausibility, with larger gains under partial observability. Additional experiments showed that AutoWorld reacts consistently to planner-induced ego behaviors and supports controllable safety-critical scenario generation, while diverse sampling produces varied yet realistic agent behaviors. We further demonstrated more favorable scaling with additional raw sensor data than a trajectory-only baseline. Our results highlight sensor-grounded world modeling as a promising direction for realistic and scalable traffic simulation beyond trajectory abstractions. Future work can explore stronger multimodal world models, where cameras provide complementary semantic and appearance cues for more realistic and human-like behavior, as well as tighter coupling between sensory-world prediction and behavior generation.


\section{Acknowledgments}
This research was enabled in part by access to NVIDIA A100 GPUs and Google TPUs.

\bibliographystyle{plainnat}
\bibliography{main}

\newpage

\appendix

\renewcommand{\thesection}{\Alph{section}}
\renewcommand{\theHsection}{appendix.\Alph{section}}

\startcontents[appendix]

{\noindent\LARGE\bfseries Appendix\par}
\vspace{1em}

\printcontents[appendix]{}{1}{}

\vspace{1em}
\newpage

\setcounter{section}{0}
\section{The AutoWorld Workflow} \label{Workflow}
Algorithms~\ref{alg:train} and~\ref{alg:infer} summarize AutoWorld's training and inference procedures.

\begin{figure}[h]
\centering
\begin{minipage}[t]{0.45\textwidth}
\begin{algorithm}[H]
\scriptsize
\caption{\small AutoWorld Training}
\label{alg:train}
\begin{algorithmic}[1]

 \State \textbf{Input:}
 \State {\tiny unlabeled data $\mathcal{D}_\mu\!=\!\{(O_{t-T_h:t+T_f})_{i}\}_{i=1}^{N_\mu}$}
\Statex {\tiny labeled data ${\mathcal{D}_f\!=\!\{(O_{t-T_h:t}, S_{t-T_h:t+T_f}, V)_{i}\}}_{i=1}^{N_f}$}
\Statex {\tiny VAE encoder $\mathcal{E}$, world model $\mu_\theta$} 
\Statex {\tiny  motion model $f_\psi$ with projections $\phi_1,\phi_2$}
\State \textbf{Output:} trained parameters $\theta,\psi$
\State \textbf{Required:} hyperparameters $\lambda,\Delta,\delta$

\Statex
\State $\mathcal{D}_\mu'=\{Y_{t-T_h:t+T_f}=\textsc{Vox}(O_{t-T_h:t+T_f})\}$ \label{algline:train_voxelization_start}
\State $\mathcal{D}_\mu''=\{Z_{t-T_h:t+T_f}=\mathcal{E}(Y_{t-T_h:t+T_f})\}$ \label{algline:train_voxelization_end}

\Statex
\Statex \textcolor{blue}{// \textbf{(a)} Initialize motion-aware weight maps} 
\For{$Y_{t-T_h:t+T_f}\in\mathcal{D}_\mu'$} \label{algline:motion_aware_weight_maps_start}
    \For{$t\in[t-T_h:t+T_f]$}
    \State $\tilde{Y}_{t \leftarrow t+\Delta}\gets Y_t$
    \State $\bar C_{t,\Delta}\gets\textsc{Transition}(Y_{t+\Delta},\tilde{Y}_{t \leftarrow t+\Delta})$
    \State $W_t\gets1+\lambda \bar C_{t,\Delta}$  \label{algline:motion_aware_weight_maps_end}
    \EndFor
\EndFor

\Statex
\Statex \textcolor{blue}{// \textbf{(b)} World modeling (WM)}

\For{$Z_{t-T_h:t+T_f}\in\mathcal{D}_\mu''$} \label{algline:wm_start}
    \State Condition $c^\mu\gets\{Z_{t-T_h:t},\tau^{\text{ego}}_{t-T_h:t}\}$
    \State Target $\tilde z\gets Z_{t+1:t+T_f}$
    \State Sampling $z_0\sim\mathcal N(0,I),\;k_\mu\sim\mathcal U(0,1)$
    \State Interpolation $z_{k_\mu}\gets(1-k_\mu)z_0+k_\mu\tilde z$
    \Statex \hspace{4mm} \textcolor{blue}{// motion-aware latent supervision}
    \State $\mathcal L^{world}(\theta)\gets$
    \Statex $\hspace{10mm}\sum_{i,j}W_t(i,j)\|\mu_\theta(z_{k_\mu},k_\mu,c^\mu)_{i,j,:}$
    \Statex $\hspace{15mm} -(\tilde z-z_0)_{i,j,:}\|_2^2$
    \State $\theta\gets\textsc{Update}(\theta,\nabla_\theta\mathcal L^{world})$ \label{algline:wm_end}
\EndFor

\Statex
\Statex \textcolor{blue}{// \textbf{(c)} WM-conditioned motion generation}
\For{$(O_{t-T_h:t},S_{t-T_h:t+T_f},V)\in\mathcal D_f$} \label{algline:motion_start}
    \State $Z_{t-T_h:t}\gets\mathcal E(\textsc{Vox}(O_{t-T_h:t}))$
    \State Condition $c^\mu\gets\{Z_{t-T_h:t},\tau^{\text{ego}}_{t-T_h:t}\}$
    \Statex \hspace{4mm} \textcolor{blue}{// sample one future occupancy rollout}
    \State $\hat Z_{t+1:t+T_f}\gets\mu_\theta(\cdot,c^\mu)$ \label{algline:sample_from_wm}
    \State $g\gets\mathrm{MHCA}(\phi_1\!(\{\hat z_{t+e\delta}\}_{e=1}^{[ T_f/\delta]}))$\label{algline:predictive_scene_context}
    \Statex \hspace{4mm} \textcolor{blue}{// per-time-step condition}
    \State $c_t^f\gets\{\mathrm{MHCA}(\phi_2([\hat z_t;g]))\}_{t=t+1}^{t+T_f}$ \label{algline:diff_condition}
    \State $\mathcal L^{motion}(\psi)\gets\textsc{DiffLoss}(f_\psi(\cdot,c_t^f))$
    \State $\psi\gets\textsc{Update}(\psi,\nabla_\psi\mathcal L^{motion})$
    \label{algline:motion_end}
\EndFor

\end{algorithmic}
\end{algorithm}
\end{minipage}
\begin{minipage}[t]{0.4\textwidth}
\begin{algorithm}[H]
\scriptsize
\caption{\small AutoWorld Inference}
\label{alg:infer}
\begin{algorithmic}[1]

\State \textbf{Input:} initial scene $(O_{t-T_h:t},S_{t-T_h:t},V)$,
\Statex $\quad$VAE encoder $\mathcal{E}$, world model $\mu_\theta$,
\Statex $\quad$motion model $f_\psi$
\State \textbf{Output:} $(N\times M)$ simulated rollouts $\text{Sen}$
\State \textbf{Required:} sample counts $(N\times M)$

\Statex
\State $Z_{t-T_h:t}\gets \mathcal E(\textsc{Voxelize}(O_{t-T_h:t}))$ \label{algline:vox_then_encode}
\State condition $c^\mu\gets\{Z_{t-T_h:t},\tau^{\text{ego}}_{t-T_h:t}\}$

\Statex
\Statex \textcolor{blue}{// Determinantal Point Process (DPP)-}
\Statex \textcolor{blue}{ guided world-model sampling}
\For{$i=1,\dots,N$} \label{algline:dpp_wm_start}
    \State $z^{(i)}_{K_\mu}\sim\mathcal N(0,I)$
\EndFor
\For{$k_\mu=K_\mu,\dots,1$}
    \State $\mathcal P^{q,\text{world}}_{k_\mu}\gets \textsc{DPP}(\{z^{(i)}_{k_\mu}\}_{i=1}^{N})$
    \For{$i=1,\dots,N$}
        \State \label{algline:dpp_wm_end} $z^{(i)}_{k_\mu-1}\gets \mu_\theta(z^{(i)}_{k_\mu},k_\mu,c^\mu)-$ 
        \Statex $\hspace{15mm}\gamma_\mu(k_\mu)\nabla_{z^{(i)}_{k_\mu}}\log \mathcal P^{q,\text{world}}_{k_\mu}$ 
    \EndFor
\EndFor

\State \label{algline:dpp_res_wm} $\hat Z_{t+1:t+T_f}^{(i)}\gets z_0^{(i)},\quad i=1,\dots,N$ 

\Statex
\Statex \textcolor{blue}{// cascaded latent sampling}
\For{$i=1,\dots,N$}\label{algline:cascade_start}
    \State $g^{(i)}\gets \mathrm{MHCA}(\phi_1(\{\hat z_{t+e\delta}^{(i)}\}_{e=1}^{\lfloor T_f/\delta\rfloor}))$ \label{algline:cascade_g}
    \Statex \hspace{4mm} \textcolor{blue}{// per time-step condition}
    \State $c_t^{f(i)}\gets\{\mathrm{MHCA}(\phi_2([\hat z_t^{(i)};g^{(i)}]))\}_{t=t+1}^{t+T_f}$\label{algline:cascade_c}
    \For{$j=1,\dots,M$}
        \State $\tau_{K_f}^{(i,j)}\sim\mathcal N(0,I)$
    \EndFor
    \For{$k_f=K_f,\dots,1$}
        \State $\mathcal P_{k_f}^{q,\text{motion}}\gets \textsc{DPP}(\{\tau_{k_f}^{(i,j)}\}_{j=1}^{M})$
        \For{$j=1,\dots,M$} 
            \State $\tau_{k_f-1}^{(i,j)}\gets f_\psi(\tau_{k_f}^{(i,j)},k_f,c_t^{f(i)})-$ \label{algline:cascade_end}
            \Statex $\hspace{20mm}\gamma_f(k_f)\nabla_{\tau_{k_f}^{(i,j)}}\log \mathcal P_{k_f}^{q,\text{motion}}$ 
        \EndFor
    \EndFor
    \State \label{algline:final} $\hat{\tau}_{t+1:t+T_f}^{(i,j)} \gets  \tau_0^{(i,j)},\quad i=1,\dots,N,\;$  
    \Statex $\hspace{15mm} j=1,\dots,M$

\EndFor

\end{algorithmic}
\end{algorithm}
\end{minipage}
\end{figure}

\subsection{Training} 
Algorithm~\ref{alg:train} begins by voxelizing LiDAR sequences and encoding them into latent occupancies using the VAE encoder  (L\ref{algline:train_voxelization_start}--\ref{algline:train_voxelization_end}). \textcolor{blue}{(a)} Motion-aware weight maps are then computed (L\ref{algline:motion_aware_weight_maps_start}--\ref{algline:motion_aware_weight_maps_end})
by compensating for ego motion and measuring occupancy transitions. These weights emphasize dynamic regions and are used to reweight the world-model supervision. \textcolor{blue}{(b)} The world model is trained to predict future latent occupancies (L\ref{algline:wm_start}--\ref{algline:wm_end}). Given past latents and the ego past trajectory, a noisy latent state is constructed by interpolating between Gaussian noise and the target latent sequence. The model learns a velocity field that matches the displacement between the noise and the target under the motion-aware weighted loss. \textcolor{blue}{(c)} The WM-conditioned motion generator is trained using labeled trajectory data (L\ref{algline:motion_start}--\ref{algline:motion_end}). For each scene, the world model produces a single future occupancy rollout (L\ref{algline:sample_from_wm}). These predicted occupancies are aggregated to form the predictive scene context (L\ref{algline:predictive_scene_context}) and combined with per-timestep latents to form conditioning signals for the diffusion model (L\ref{algline:diff_condition}). The motion model is optimized using the standard diffusion loss. Training therefore uses one world-model rollout and one trajectory sample per scene, \textit{without any diversity mechanism}.

\vspace{1mm}
\subsection{Inference} 
Algorithm~\ref{alg:infer} generates diverse simulations by introducing DPP-guided sampling at inference time. After encoding the observed LiDAR history (L\ref{algline:vox_then_encode}), the world model produces multiple latent scene rollouts (L\ref{algline:dpp_wm_start}--\ref{algline:dpp_wm_end}). Sampling begins from Gaussian noise and is iteratively refined. At each step, a quality-weighted DPP objective is computed across the current samples. The gradient of the log determinant introduces a repulsive force between samples, encouraging diversity while preserving high quality predictions. This process yields $N$ diverse future scene rollouts $\hat{Z}_{t+1: t+T_f}^{(i)}$ (L\ref{algline:dpp_res_wm}). Given these scene forecasts, the motion model generates trajectories using cascaded latent sampling (L\ref{algline:cascade_start}--\ref{algline:cascade_end}). For each predicted latent occupancy sequence, a global context embedding $g^{(i)}$ is computed (L\ref{algline:cascade_g}) and combined with timestep latents to form the diffusion conditioning signals (L\ref{algline:cascade_c}). The diffusion process then generates $M$ trajectories per scene rollout. As in the world model, DPP guidance is applied during sampling to encourage trajectory diversity while maintaining consistency with the predicted scene dynamics. The final set of $N \times M$ trajectories corresponds to $N \times M$ diverse and plausible driving scenarios (L\ref{algline:final}).

\section{Additional Experiments} \label{additional_exp}

\subsection{Effect of Coarse-to-Fine Predictive Scene Context}
We study the effect of coarse-to-fine guidance from the world model in a controlled setting. To isolate this component, we disable DPP and use IID sampling at both stages. We compare a variant that conditions only on future latent occupancies with one that additionally incorporates the predictive scene context. As shown in Table~\ref{tab:coarse_to_fine}, adding the predictive scene context improves both realism and trajectory accuracy, increasing RMM from $0.7562$ to $0.7618$ and reducing minADE from $1.4519$ to $1.4334$. This suggests that aggregating future latents into a global context provides useful long-horizon structure for multi-agent coordination, whereas per-step latents alone offer only local guidance. Overall, the predictive scene context serves as a coarse-to-fine signal that improves the quality of generated traffic scenarios.

\begin{table}[h]
\centering 
\caption{\textbf{Effect of coarse-to-fine predictive scene context on traffic simulation.} We compare conditioning on future latent occupancies alone versus combining them with a predictive scene context, under a fixed sampling budget with IID sampling. Incorporating the predictive scene context improves overall realism.}
\resizebox{0.5\columnwidth}{!}{%
\begin{tabular}{lcc}
    \toprule
    \textbf{Conditioning} & \textbf{RMM} $(\uparrow)$ & \textbf{minADE} $(\downarrow)$ \\
    \midrule
    Future latent only & 0.7562 & 1.4519 \\
    + Pred. scene context & \textbf{0.7618} & \textbf{1.4334} \\
    \bottomrule
\end{tabular}%
}
\vspace{-2mm}
\label{tab:coarse_to_fine}
\end{table}

\subsection{Effect of Budget Allocation between World Model and Motion Generation.}

Table~\ref{tab:wosac_ablation} studies how a fixed budget of 32 rollouts is allocated between world-model hypotheses ($N$) and motion samples per hypothesis ($M$). The baseline refers to AutoWorld's diffusion motion generator without world-model grounding. The same baseline is also used as the trajectory-only variant in Sec. 4.3 of the main paper. The results show that adding world-model grounding improves over this baseline across all budget allocations, but the allocation matters: using only one scene hypothesis ($1\times32$) limits future-scene diversity, while using too few motion samples per hypothesis ($32\times1$) weakens behavioral realism. Performance is highest with the balanced $8\times4$ allocation, suggesting that allocating budget to both world-model hypotheses and motion samples is more effective than concentrating it on either stage alone. With cascaded latent sampling, $4\times8$ achieves the best performance, improving RMM to 0.7746 and substantially reducing minADE. This indicates that once scene hypotheses are selected diversely, allocating more samples to motion generation better exploits the rollout budget. For fair comparison, all ablations in Sec.~4.4 of the main paper use a fixed $4\times8$ allocation, corresponding to the best-performing configuration under a budget of 32 rollouts.

\begin{table}[h]
\centering \vspace{-3mm}
\caption{\textbf{Effect of budget allocation between world model and motion generation.} $N$ and $M$ denote the numbers of world-model hypotheses and motion samples per hypothesis, respectively. Allocating the full budget to either world-model hypotheses or motion samples hurts realism, while a balanced allocation achieves the best overall performance. The ``${}^\dagger$'' denotes the model used for leaderboard submission.}
\label{tab:wosac_ablation}
\resizebox{\columnwidth}{!}{%
\begin{tabular}{lcccccc}
\Xhline{0.75pt}
\textbf{Model} 
& \textbf{\textit{N}$\times$\textit{M}}
& \textbf{RMM}
& \textbf{Kinematic}
& \textbf{Interactive}
& \textbf{Map-based}
& \textbf{minADE} \\ 
\hline

Baseline & -$\times$32 & 0.7472 & 0.4749 & 0.7741 & 0.8682 & 1.4892 \\  
\Xhline{0.15pt}

\multirow{4}{*}{\begin{tabular}[c]{@{}l@{}}+ World model grounding\end{tabular}}
& 1$\times$32  & 0.7610 & 0.4849 & 0.7855 & 0.8873 & 1.4369 \\
& 4$\times$8   & 0.7597 & 0.4840 & 0.7839 & 0.8861 & 1.4256 \\
& 8$\times$4   & 0.7618 & 0.4846 & 0.7876 & 0.8870 & 1.4334 \\
& 32$\times$1  & 0.7523 & 0.4796 & 0.7728 & 0.8819 & 1.4178 \\  
\Xhline{0.25pt}

\multicolumn{7}{l}{\textit{Full AutoWorld (Cascaded Diversity; only meaningful when $N>1$ and $M>1$)}} \\
\multirow{2}{*}{+ Cascaded Latent Sampling}
& 4$\times 8^{\dagger}$ & 0.7746 & 0.4878 & 0.7968 & 0.9099 & 1.3102 \\
& 8$\times$4             & 0.7728 & 0.4889 & 0.7948 & 0.9067 & 1.3097 \\ 
\Xhline{0.75pt}
 
\end{tabular}%
}  \vspace{-2mm}
\end{table}

\subsection{Multimodal Behavior Coverage}

We also qualitatively illustrate AutoWorld's multimodal behaviors by visualizing the generated trajectories alongside the \emph{SDC paths} provided in the Waymo dataset~\cite{ettinger2021large}\footnote{\href{https://waymo.com/intl/es/open/download/\#:~:text=\%5Bnew\%5D\%20v1.3.1\%20Oct\%202025\%3A\%20Added\%20sdc_paths\%2C\%20which\%20are\%20a\%20new\%20WOMD\%20feature\%20indicating\%20example\%20valid\%20future\%20routes\%20that\%20the\%20self\%2Ddriving\%20car\%20(SDC)\%20could\%20take.\%20Check\%20out\%20the\%20sdc_paths\%20demo\%20on\%20GitHub\%20to\%20get\%20started.\%20\%2D\%20files}{Waymo Open Motion Dataset: v1.3.1}}. The self-driving car (SDC) refers to the ego vehicle in each Waymo scenario. The SDC paths annotation provides a set of candidate route polylines representing plausible future routes the ego vehicle could follow according to the road topology, such as lane following, turning, or merging behaviors. Since these paths are geometric route hypotheses rather than time-aligned trajectories, they often extend beyond the 8-second forecasting horizon used in motion prediction. For visualization, we therefore first retain only single continuous path segments and truncate each path to a comparable horizon based on the future distance traveled by the ground-truth SDC. To avoid displaying many nearly identical routes that arise from map discretization, the truncated paths are further deduplicated by resampling them to a fixed number of points and removing paths whose mean pointwise distance falls below a similarity threshold. This process yields a compact set of distinct route hypotheses that represent the primary topological futures available to the ego in the scene.

Figure~\ref{fig:SDC_coverage} visualizes the ground-truth SDC trajectory, the resulting candidate SDC paths, and 32 predicted ego rollouts generated by AutoWorld. For visualization, SDC paths are loaded using the procedure described in the official Waymax tutorial\footnote{\href{https://github.com/waymo-research/waymax/blob/main/docs/notebooks/sdc_paths_demo.ipynb}{Official Waymax tutorial}}. Across scenarios, AutoWorld demonstrates meaningful multimodal coverage, producing diverse behaviors such as alternative turning directions and lane changes within a single scene, capturing both topological and kinematic diversity. While most candidate routes are covered, a few SDC paths corresponding to more complex topological traversals remain uncovered. We view this as an opportunity for future work, where incorporating explicit road-topology reasoning into the sampling process may enable better coverage of such complex routes over longer horizons.

\newpage

\begin{center}

\captionof{figure}{\textbf{Qualitative analysis of multimodal behavior coverage}. The ground-truth ego trajectory, candidate SDC paths from Waymo, and 32 rollouts generated by AutoWorld are visualized (left to right). The model produces diverse behaviors within a single scenario, including alternative turning directions and lane changes, with diversity that also reflects different kinematic profiles.}  
\label{fig:SDC_coverage}
\begin{longtable}{ccc}

\includegraphics[width=0.30\textwidth]{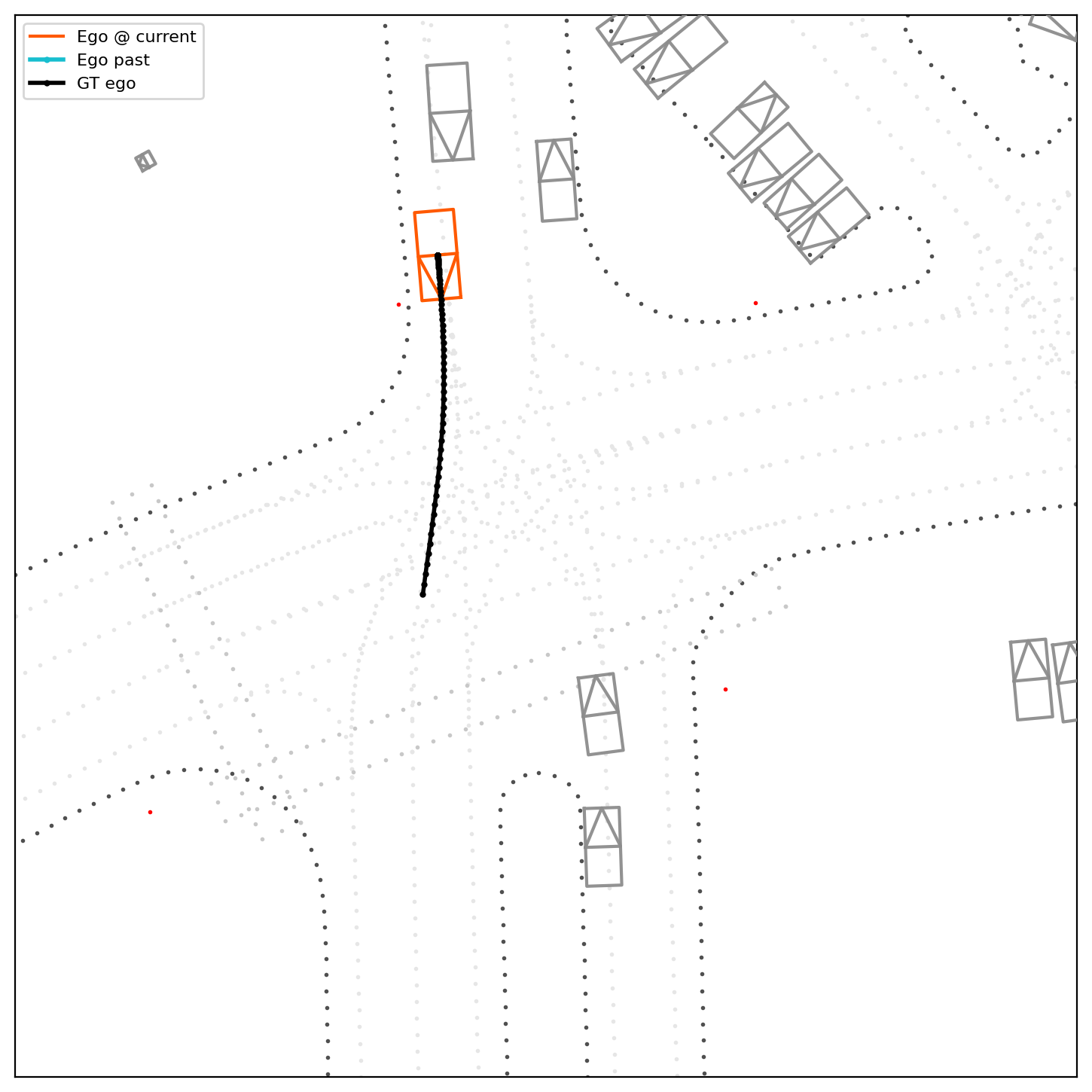} &
\includegraphics[width=0.30\textwidth]{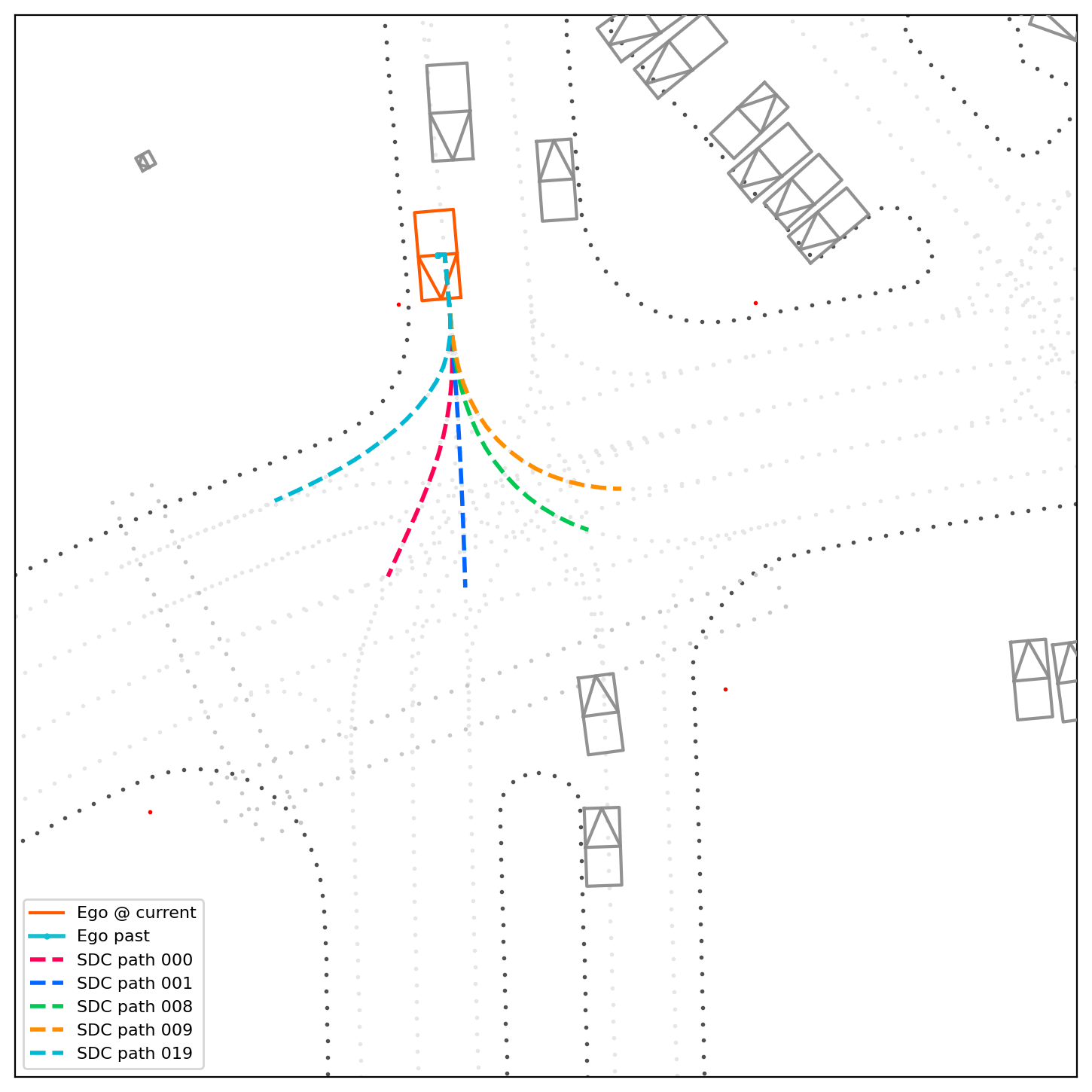} &
\includegraphics[width=0.30\textwidth]{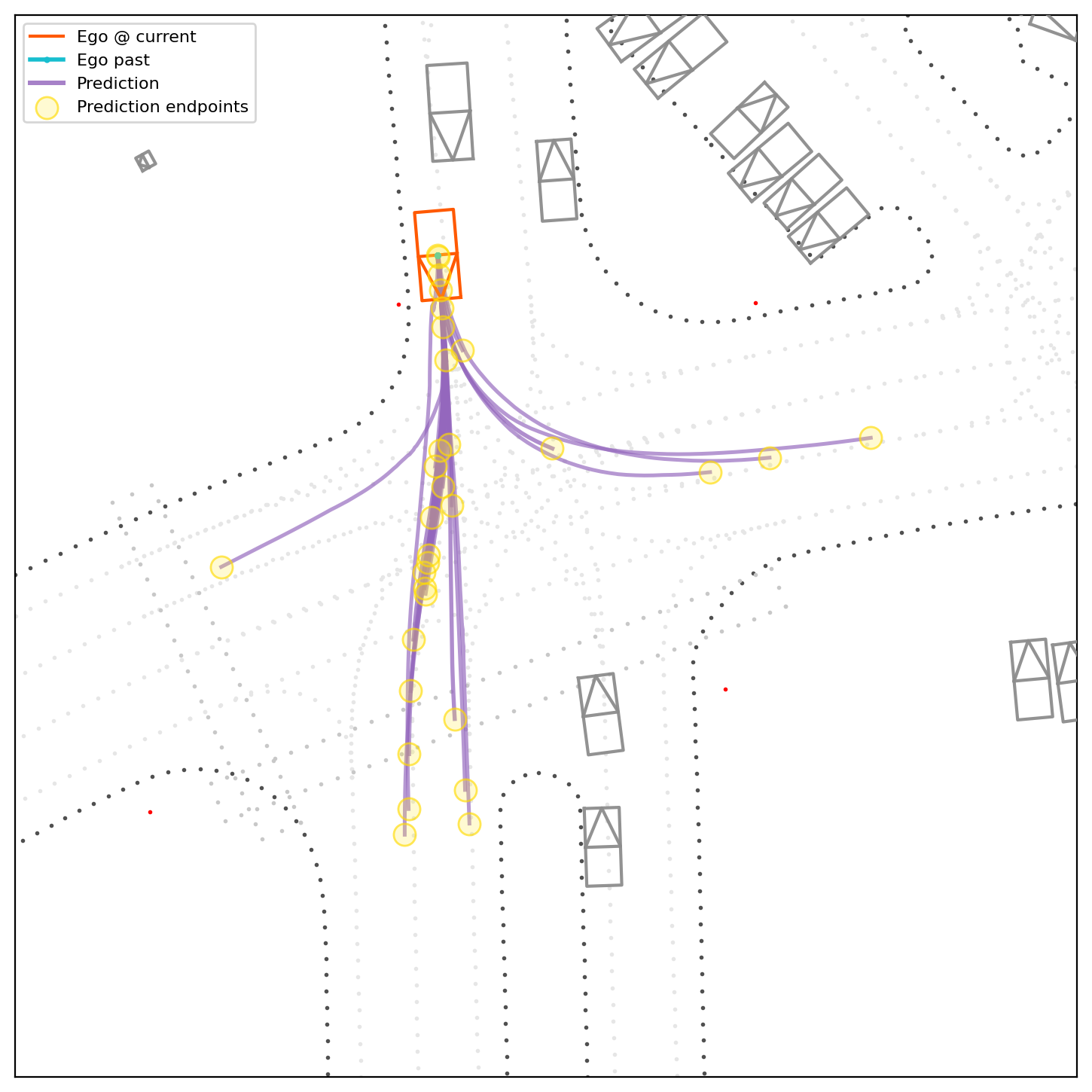} \\
\includegraphics[width=0.30\textwidth]{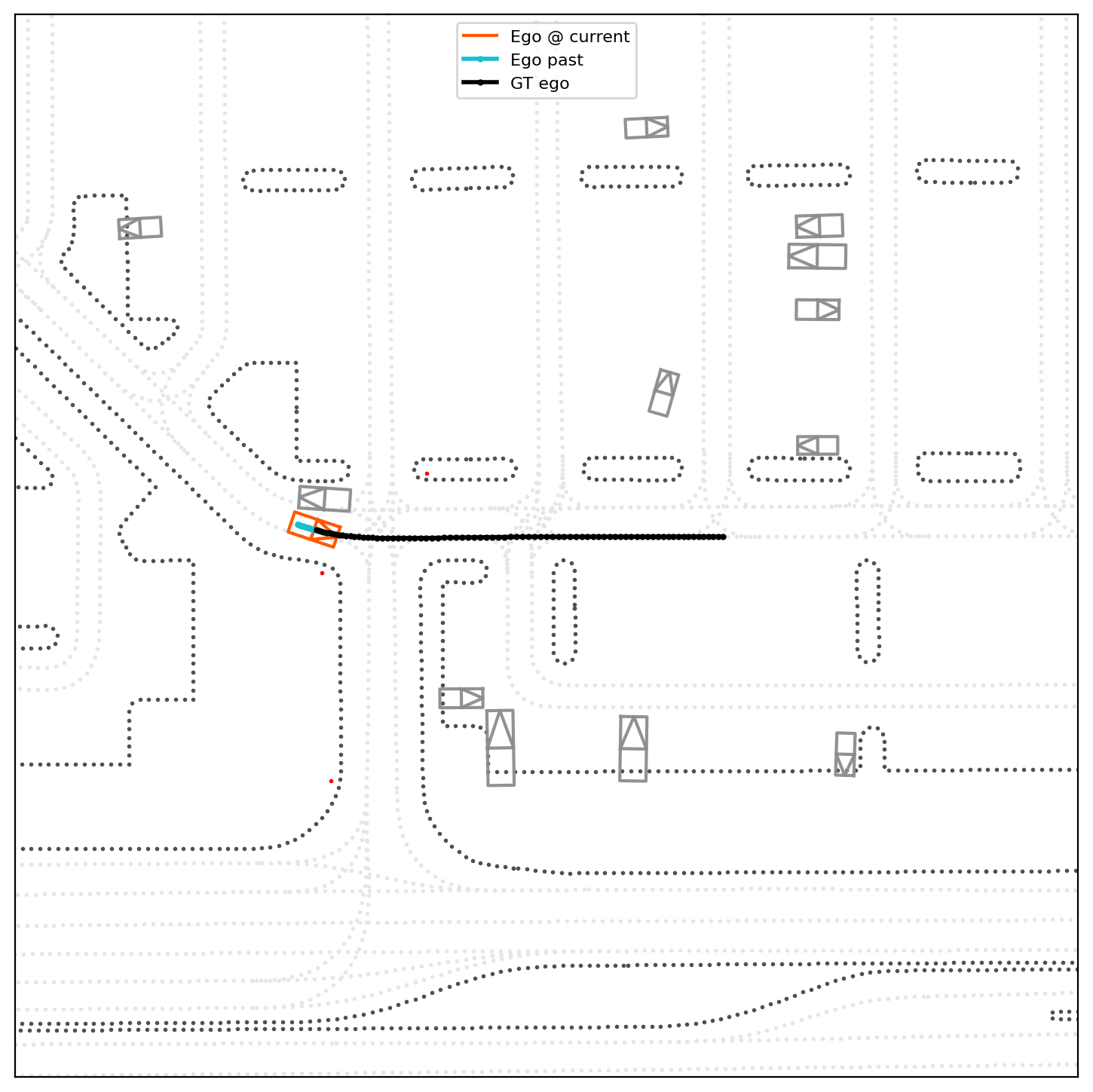} &
\includegraphics[width=0.30\textwidth]{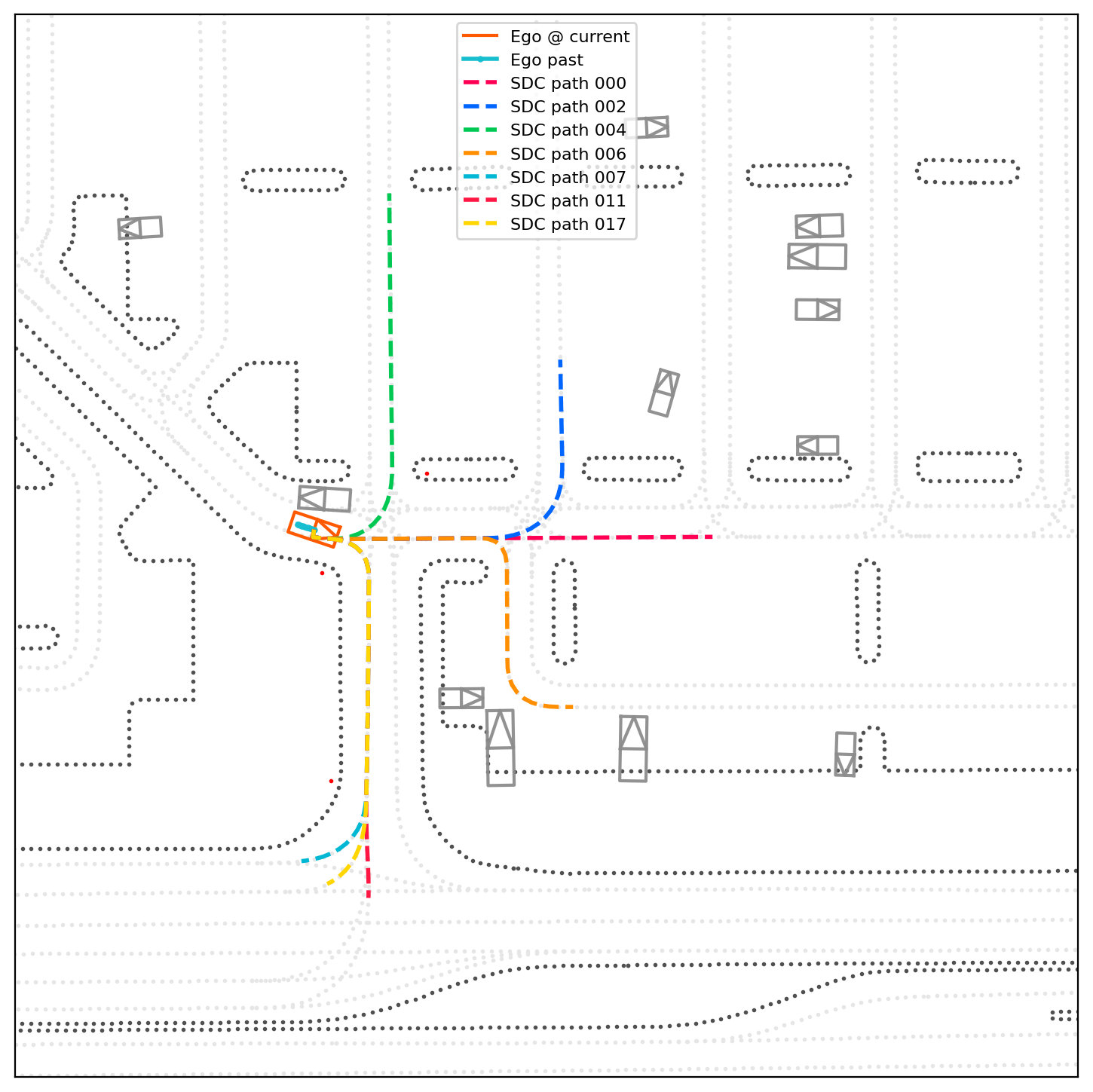} &
\includegraphics[width=0.30\textwidth]{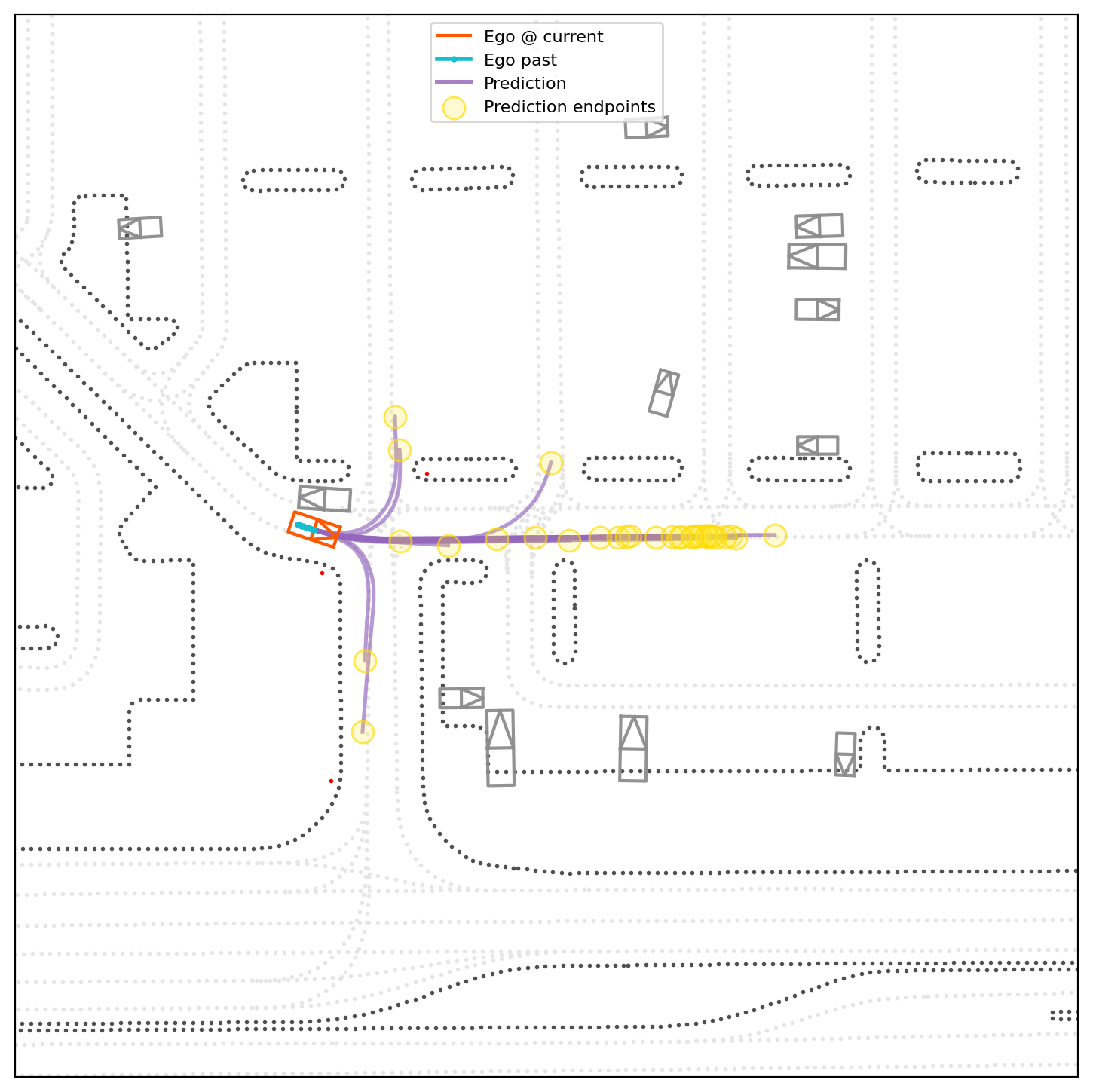} \\
\includegraphics[width=0.30\textwidth]{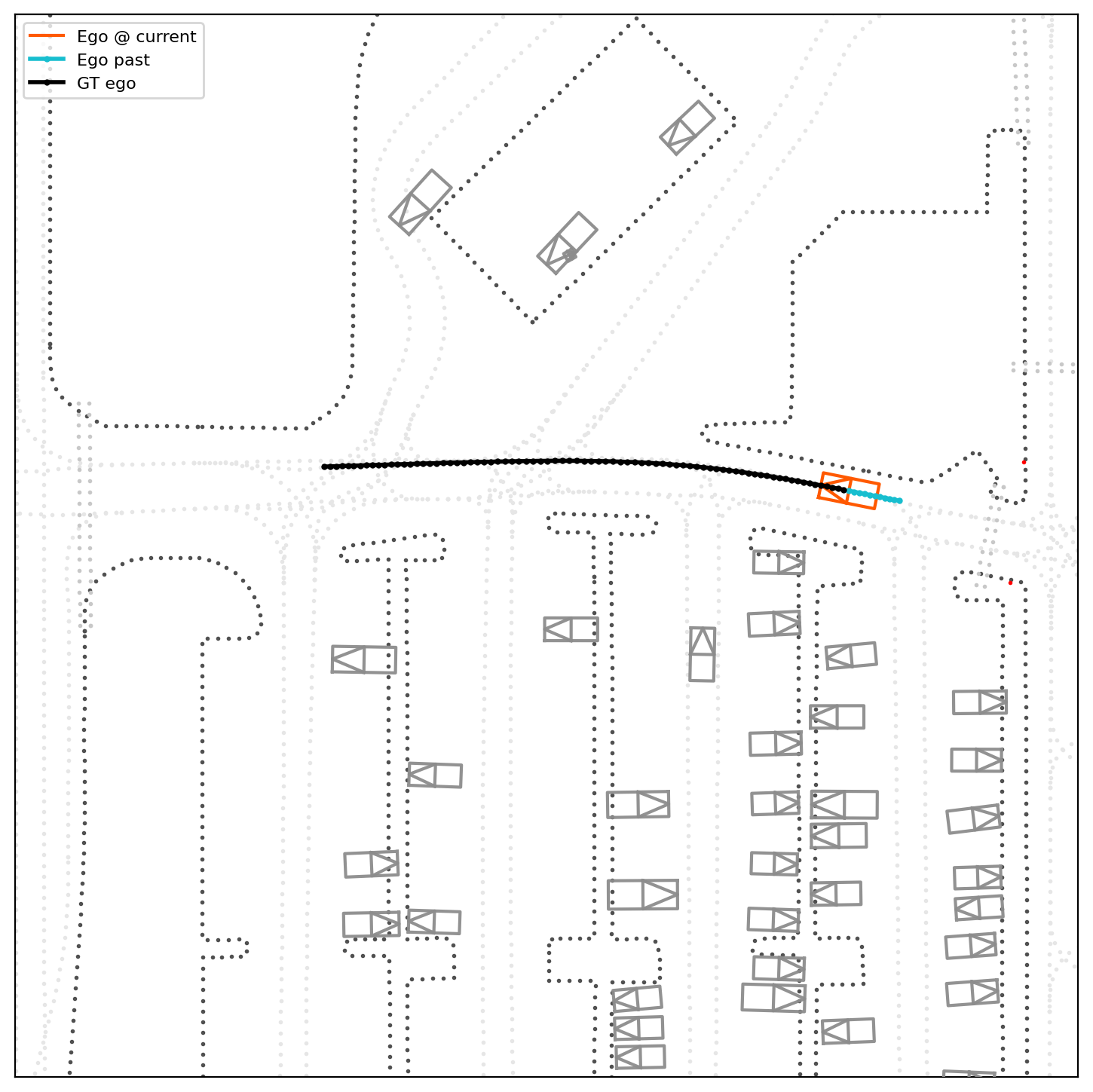} &
\includegraphics[width=0.30\textwidth]{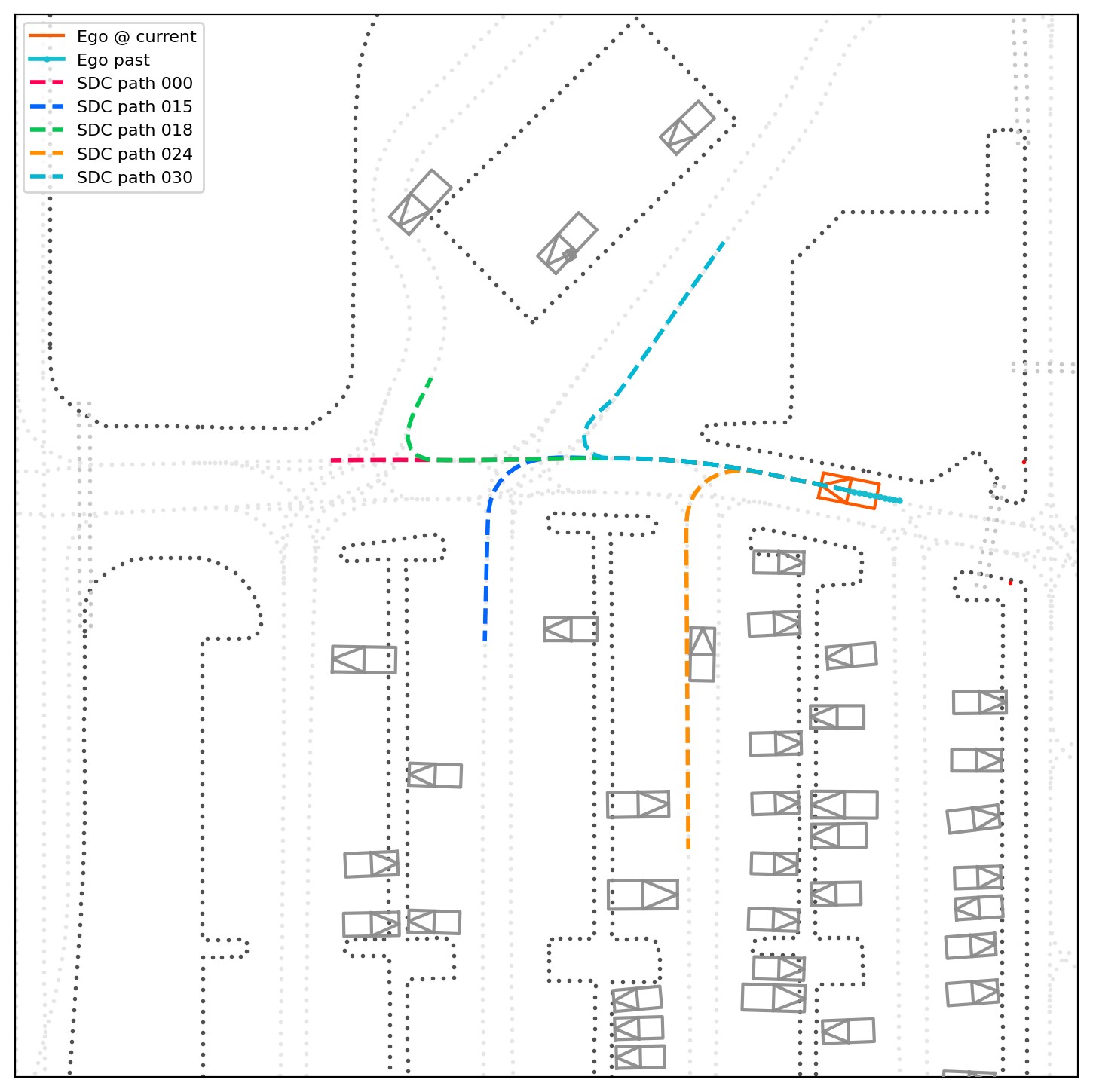} &
\includegraphics[width=0.30\textwidth]{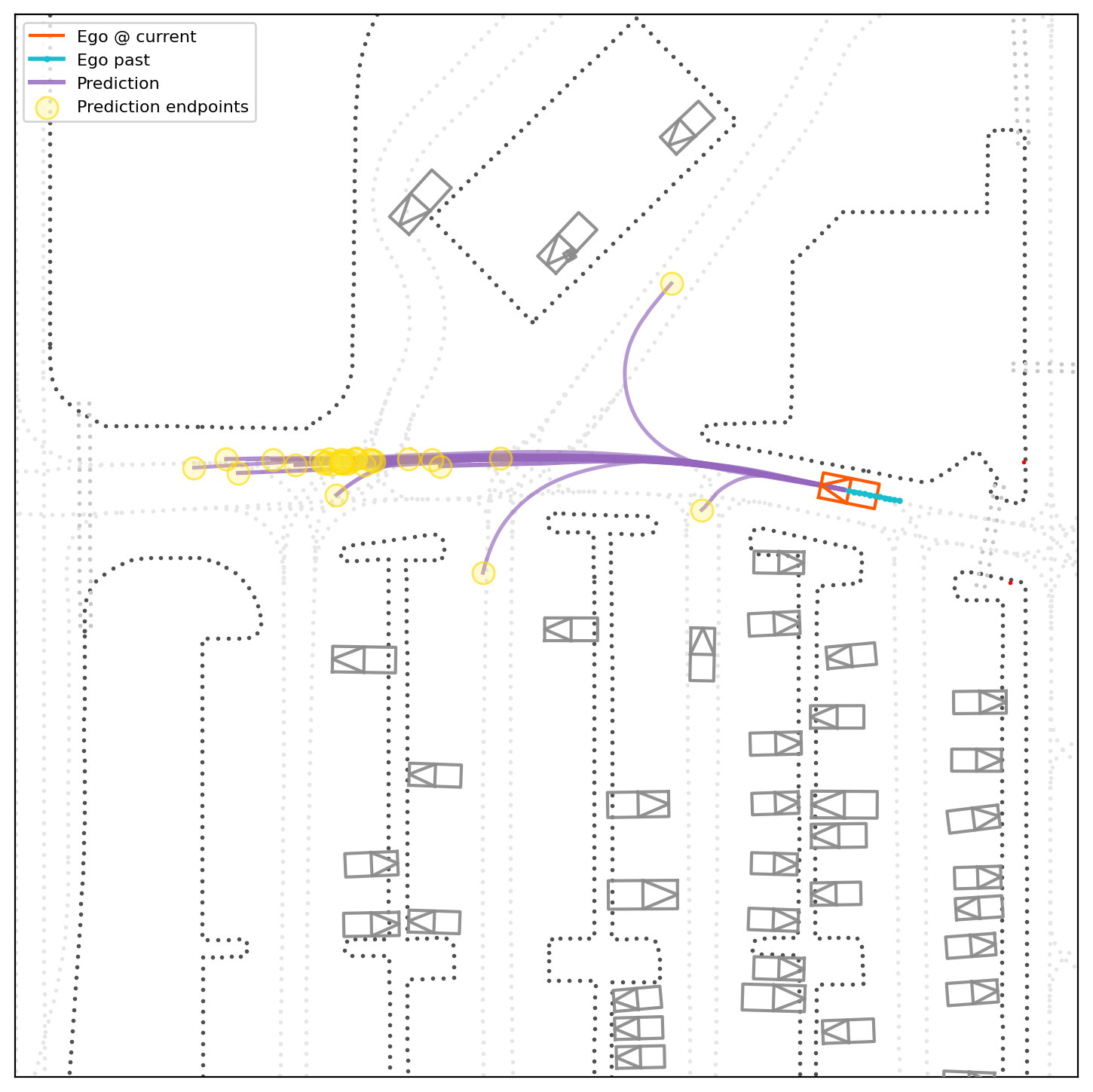} \\
\includegraphics[width=0.30\textwidth]{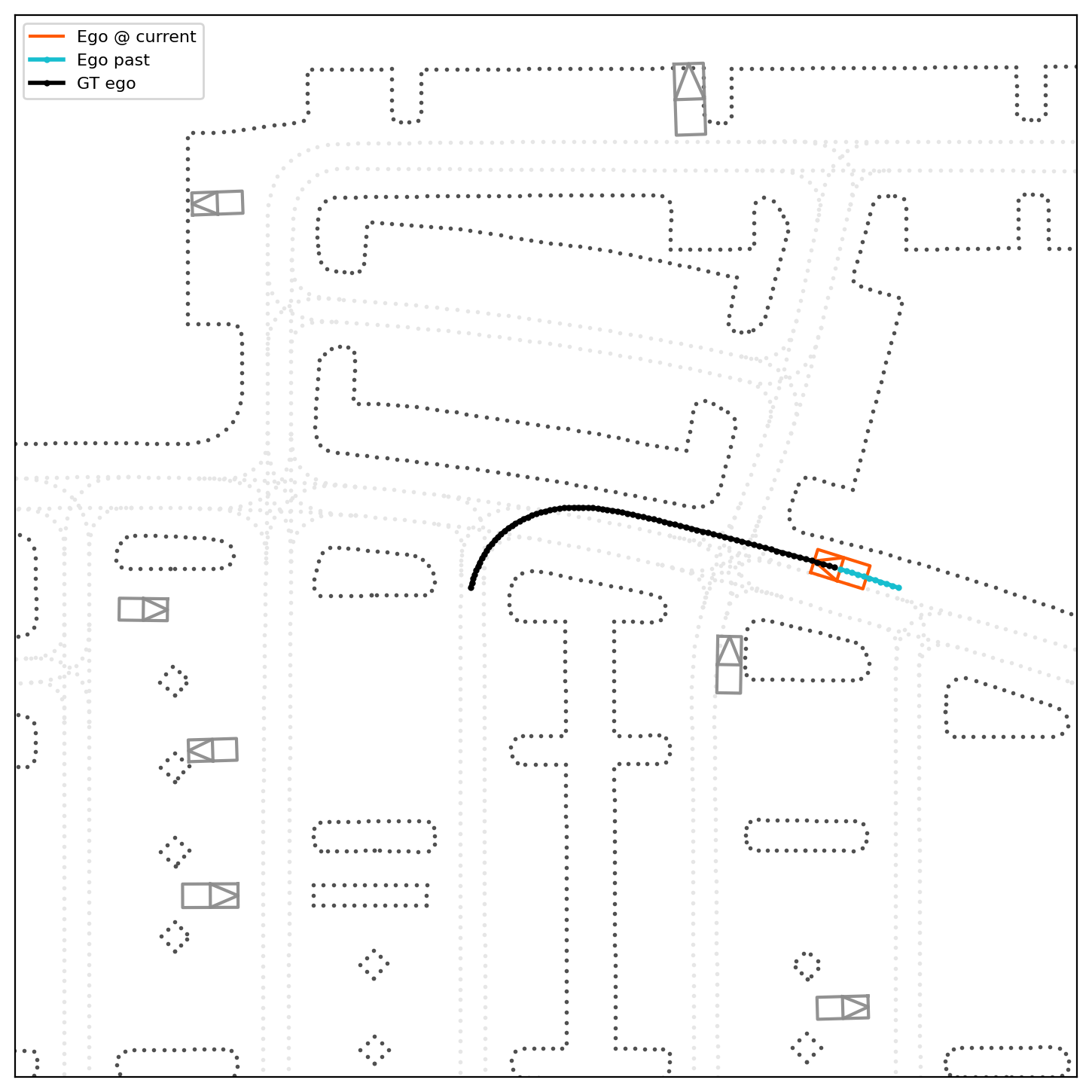} &
\includegraphics[width=0.30\textwidth]{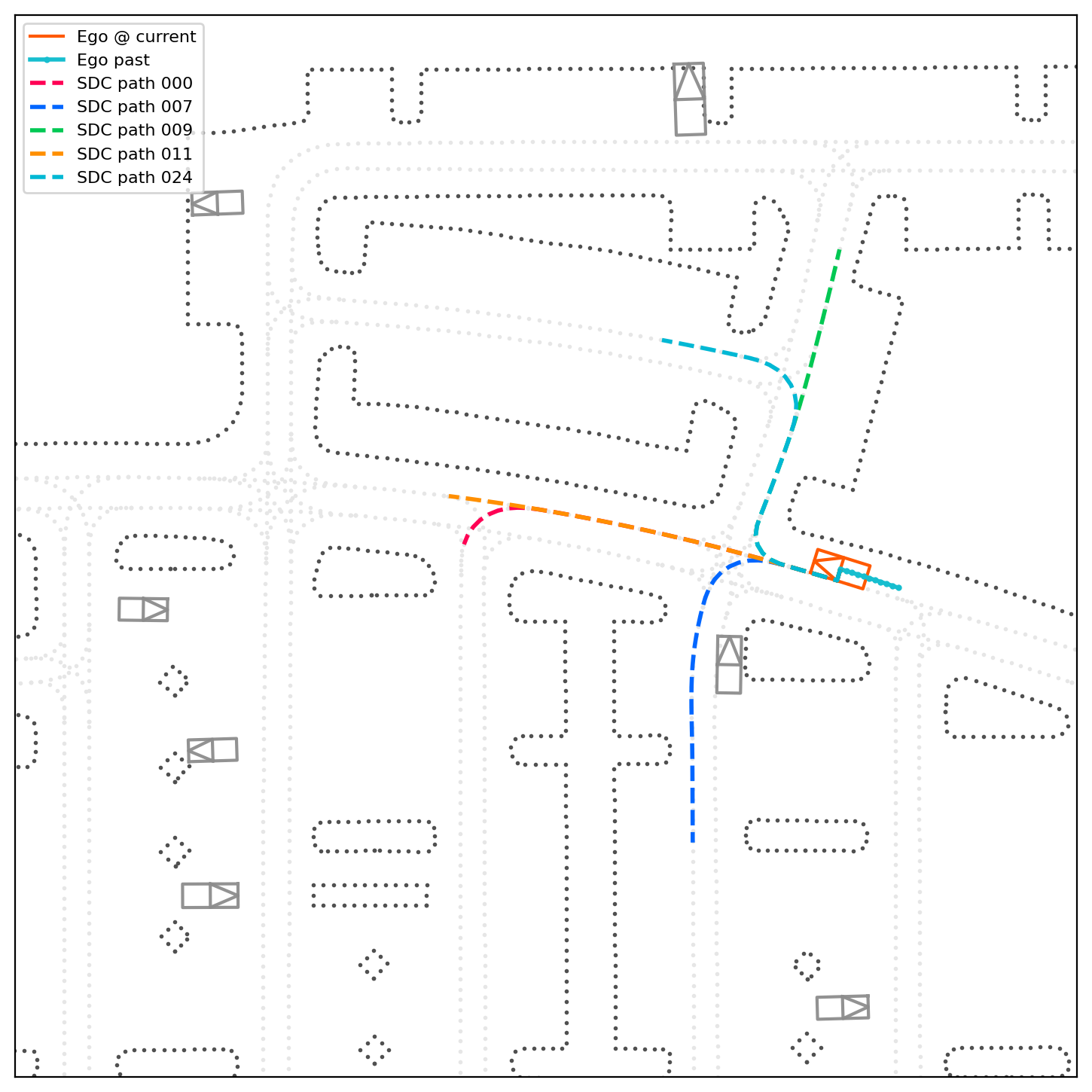} &
\includegraphics[width=0.30\textwidth]{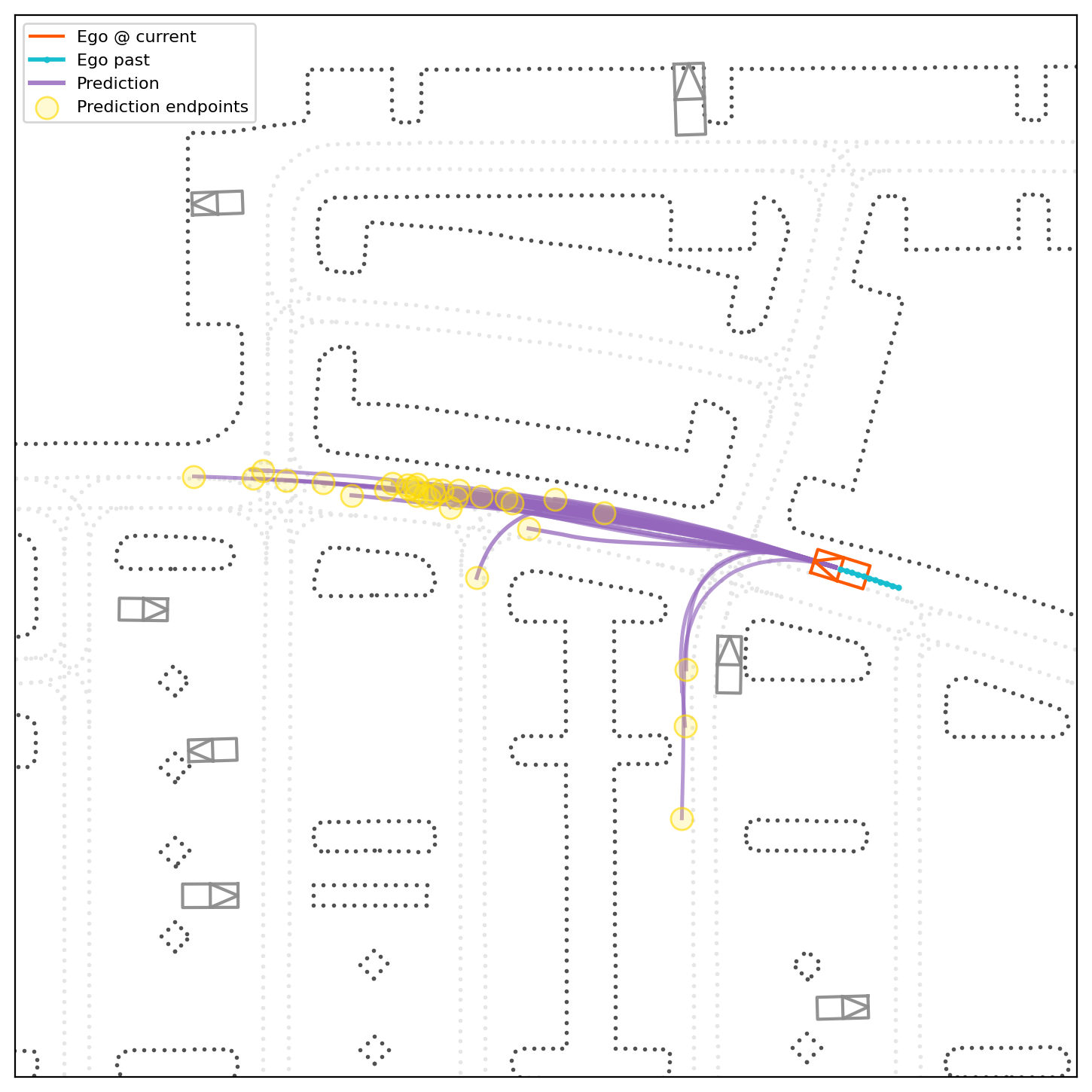} \\
\includegraphics[width=0.30\textwidth]{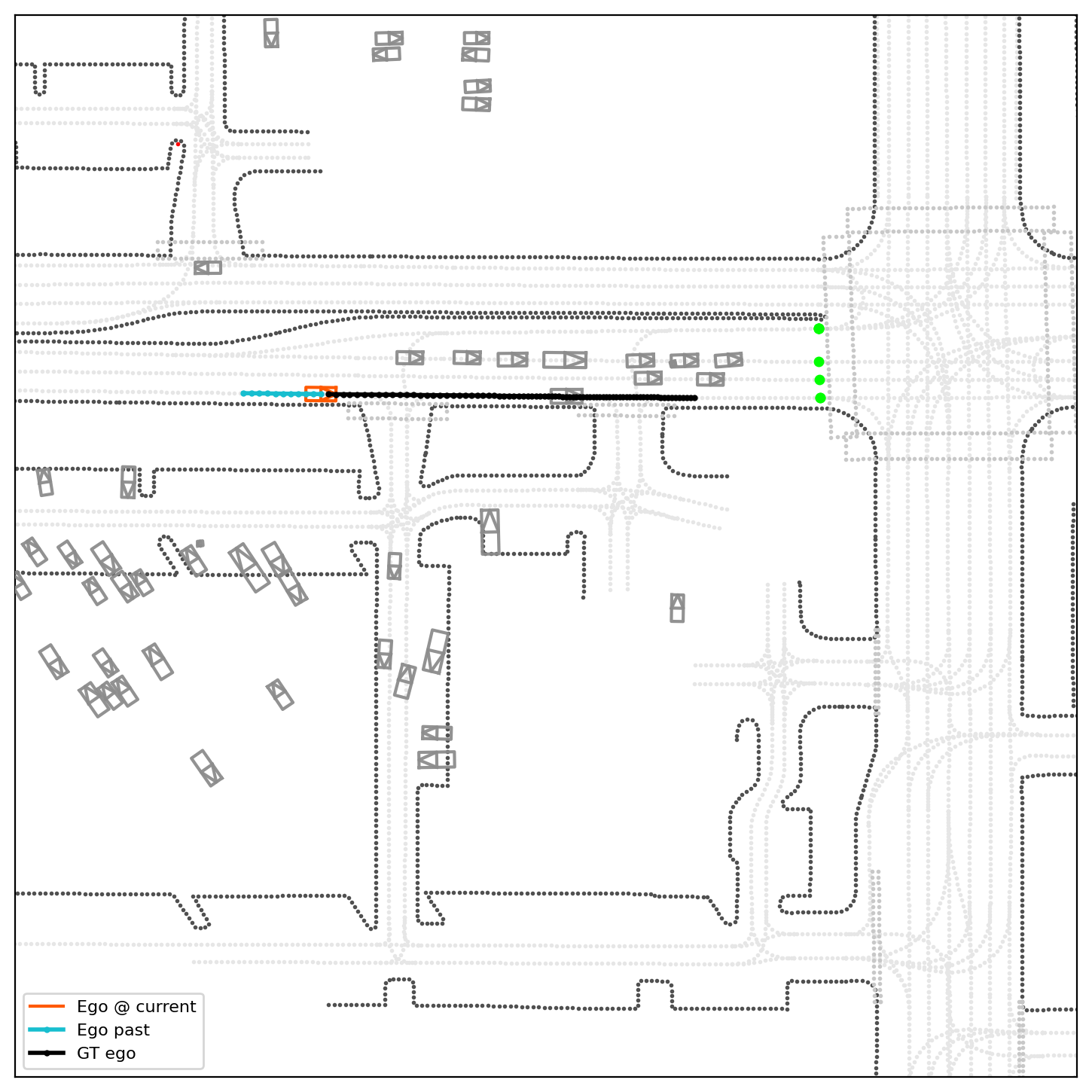} &
\includegraphics[width=0.30\textwidth]{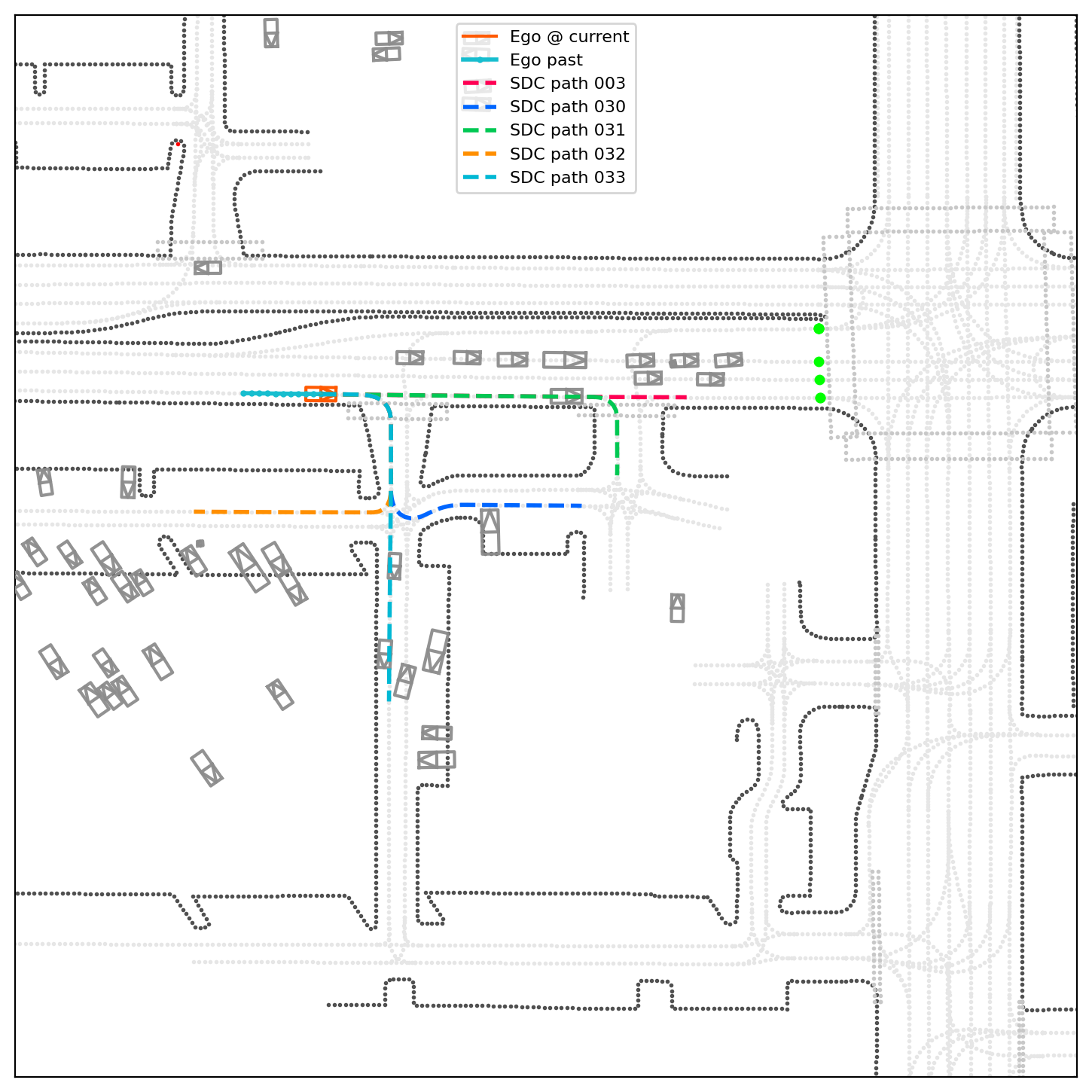} &
\includegraphics[width=0.30\textwidth]{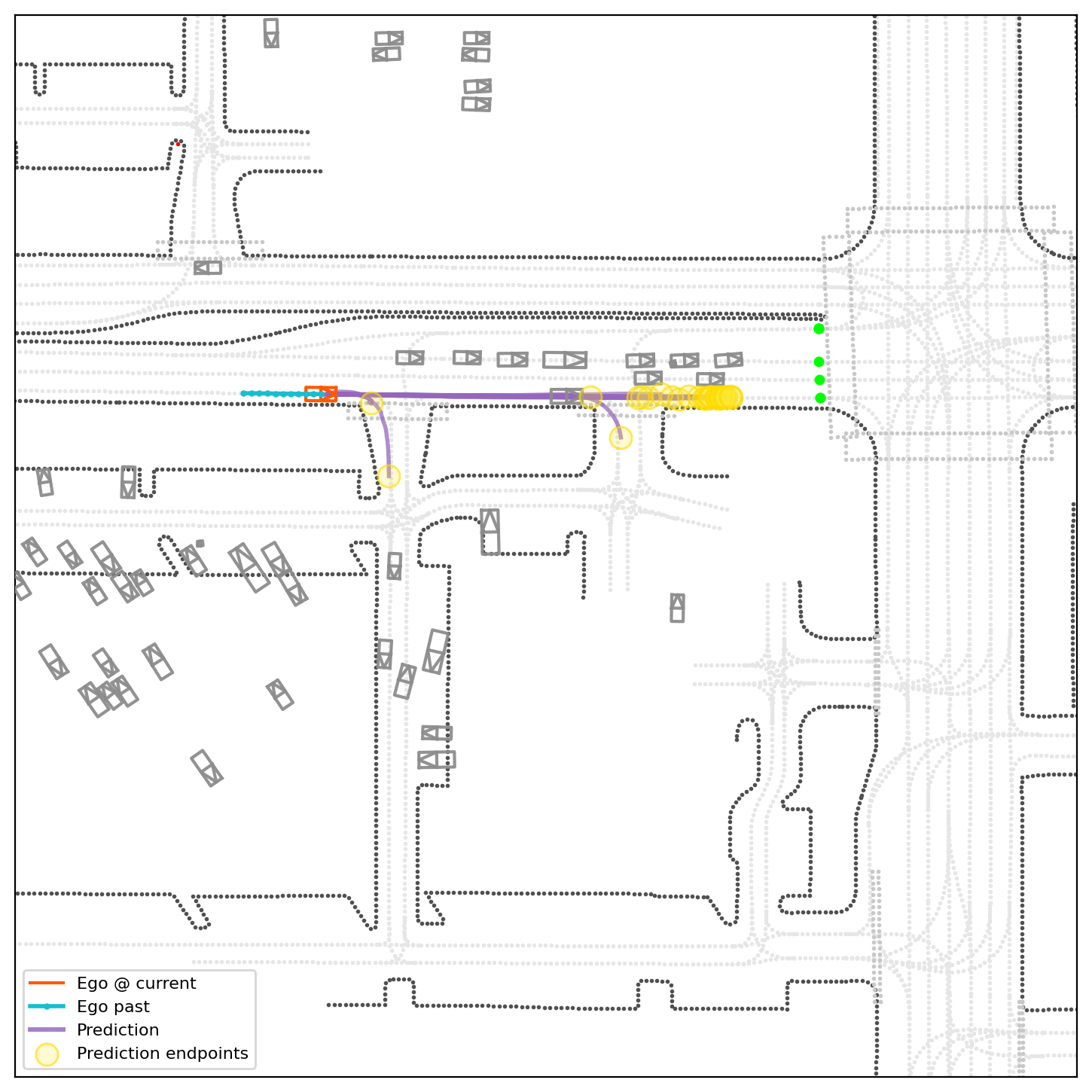} \\
\includegraphics[width=0.30\textwidth]{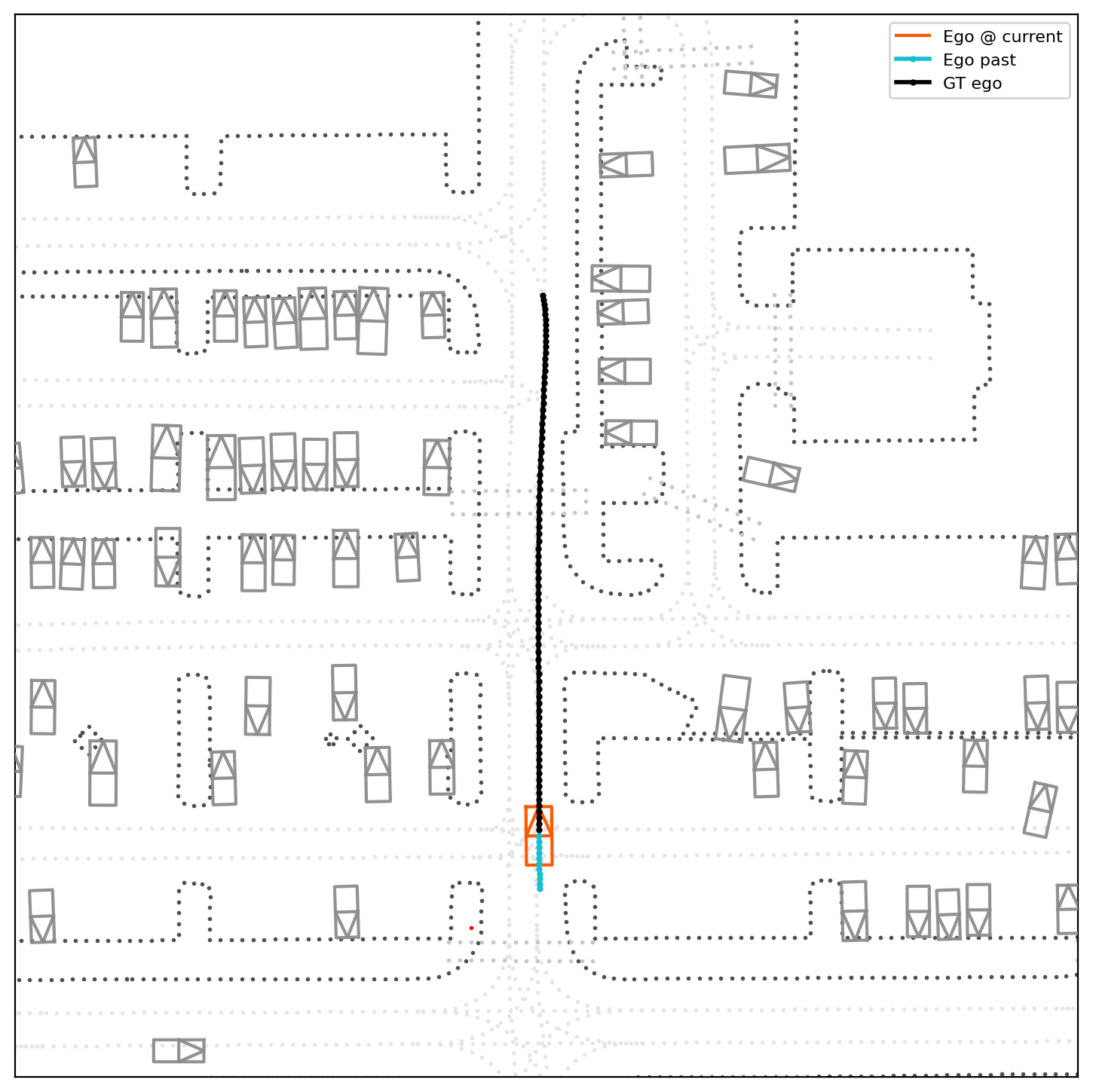} &
\includegraphics[width=0.30\textwidth]{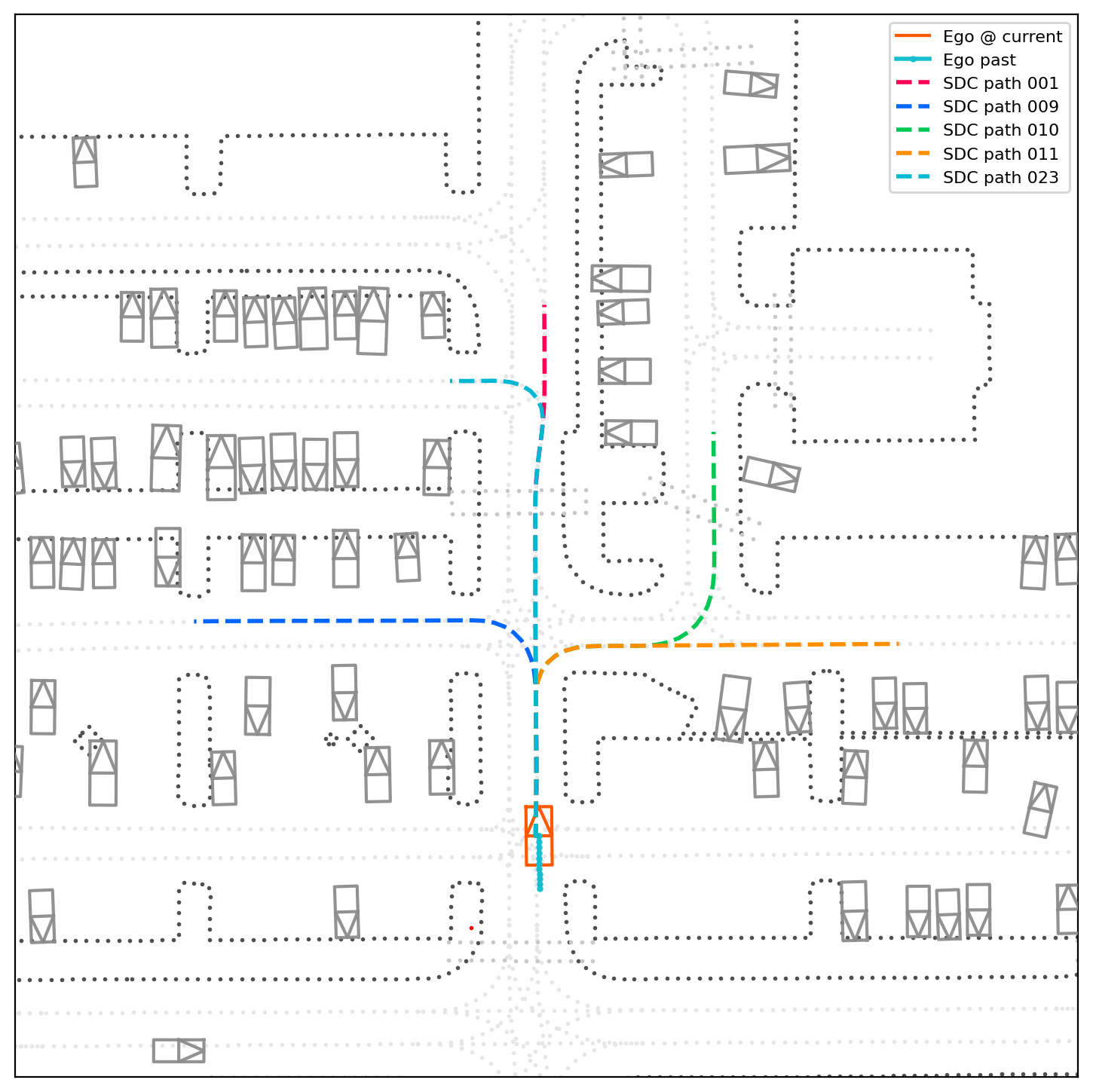} &
\includegraphics[width=0.30\textwidth]{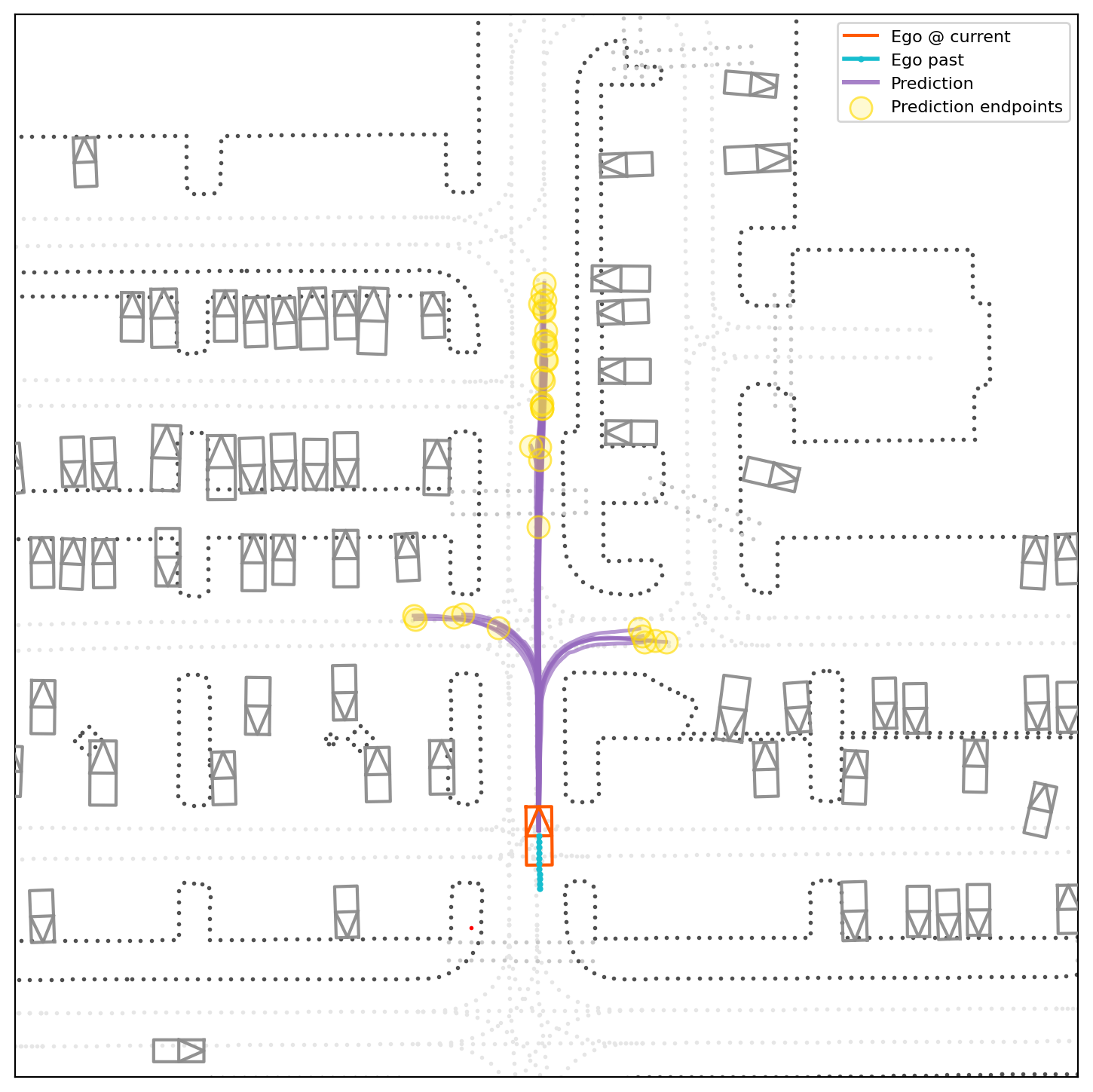} \\
\includegraphics[width=0.30\textwidth]{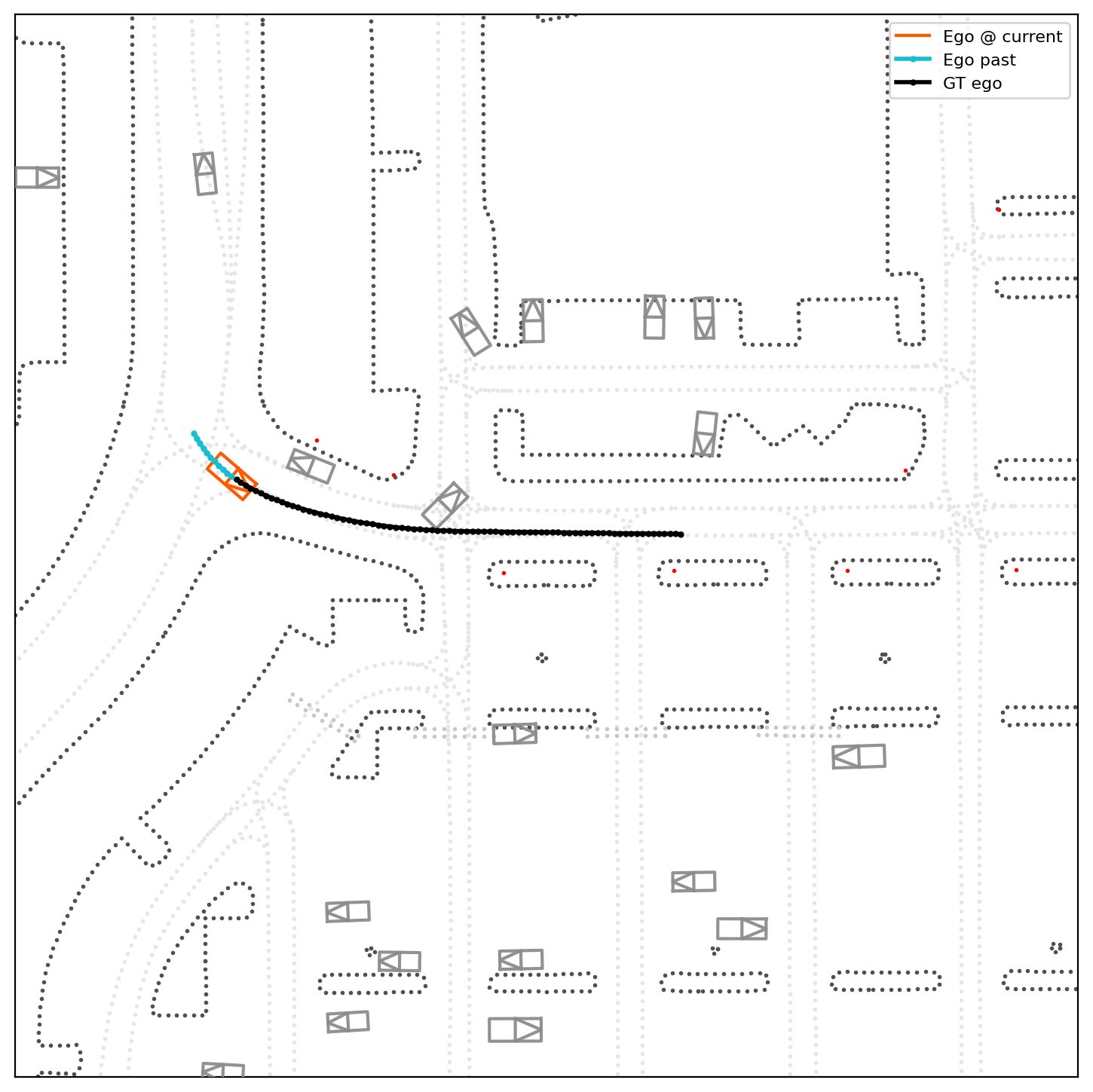} &
\includegraphics[width=0.30\textwidth]{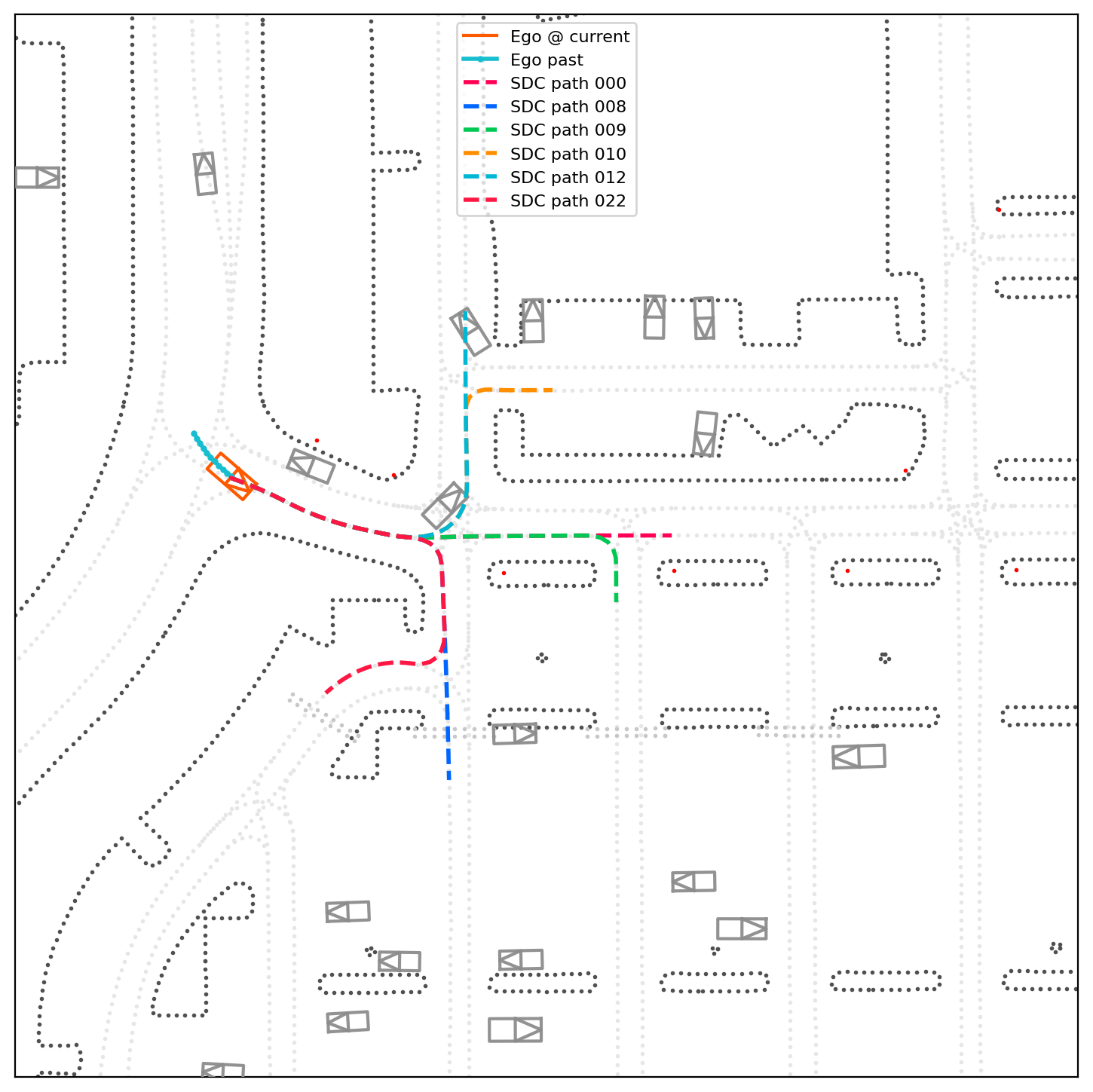} &
\includegraphics[width=0.30\textwidth]{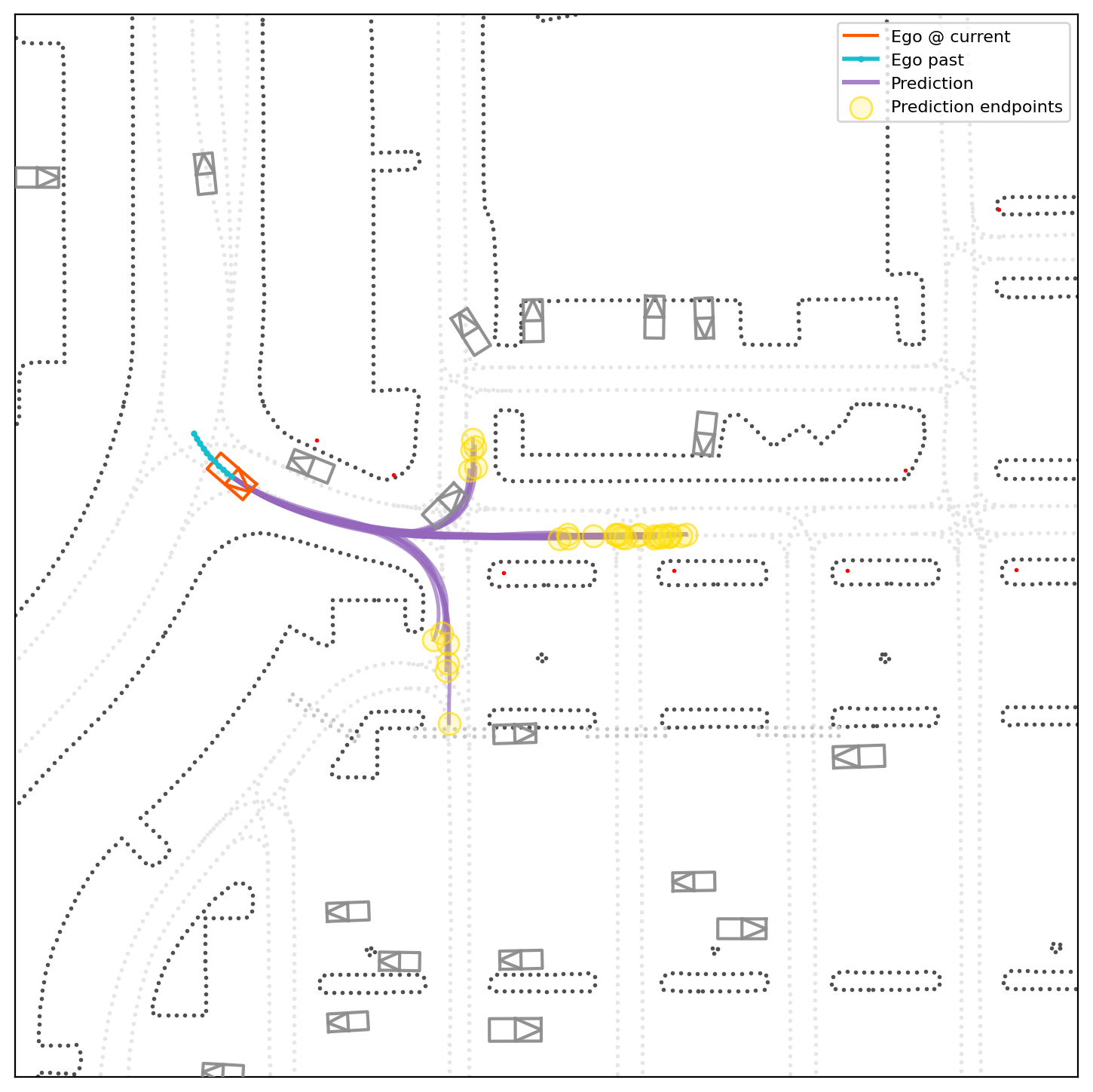} \\
\includegraphics[width=0.30\textwidth]{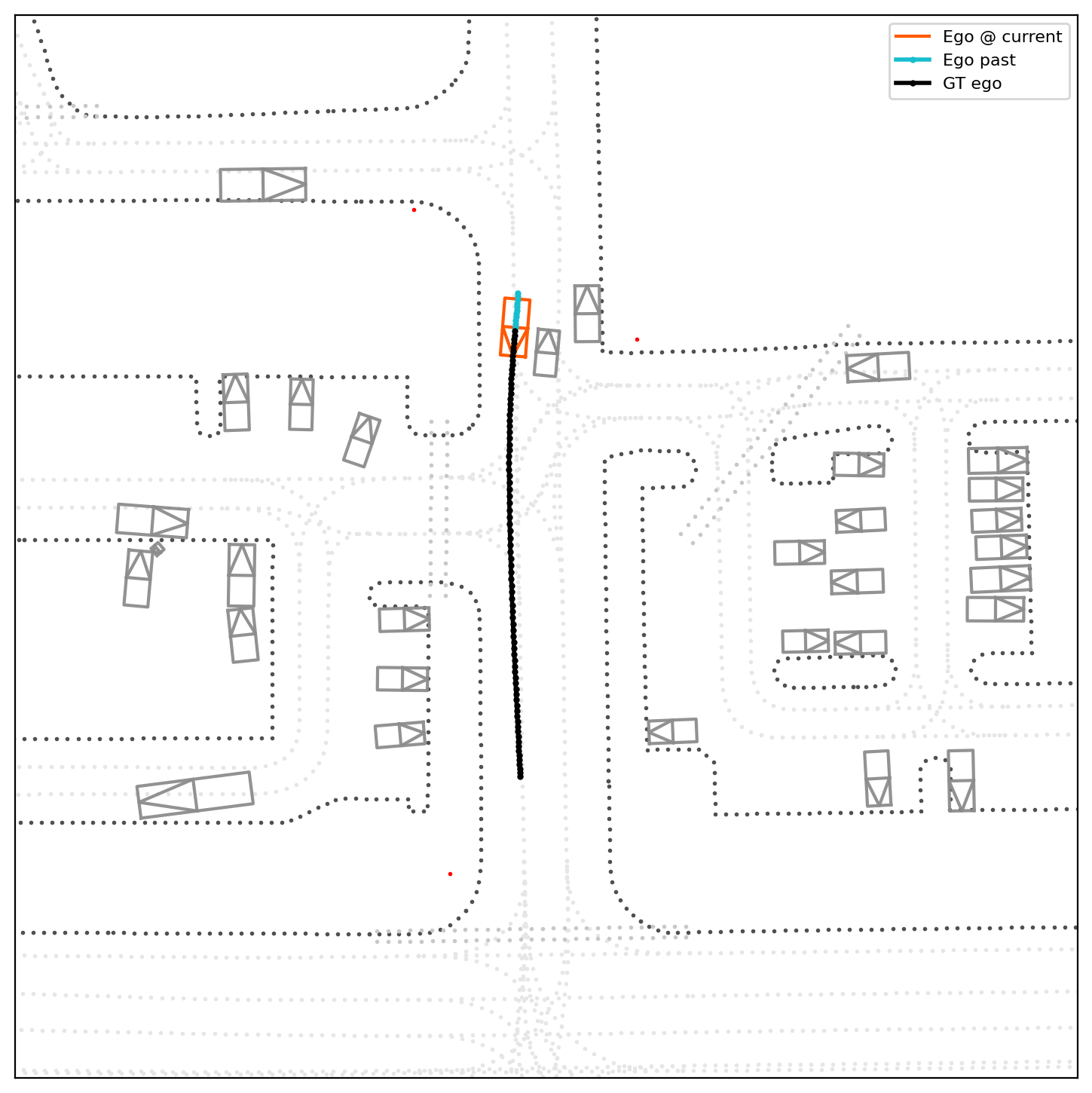} &
\includegraphics[width=0.30\textwidth]{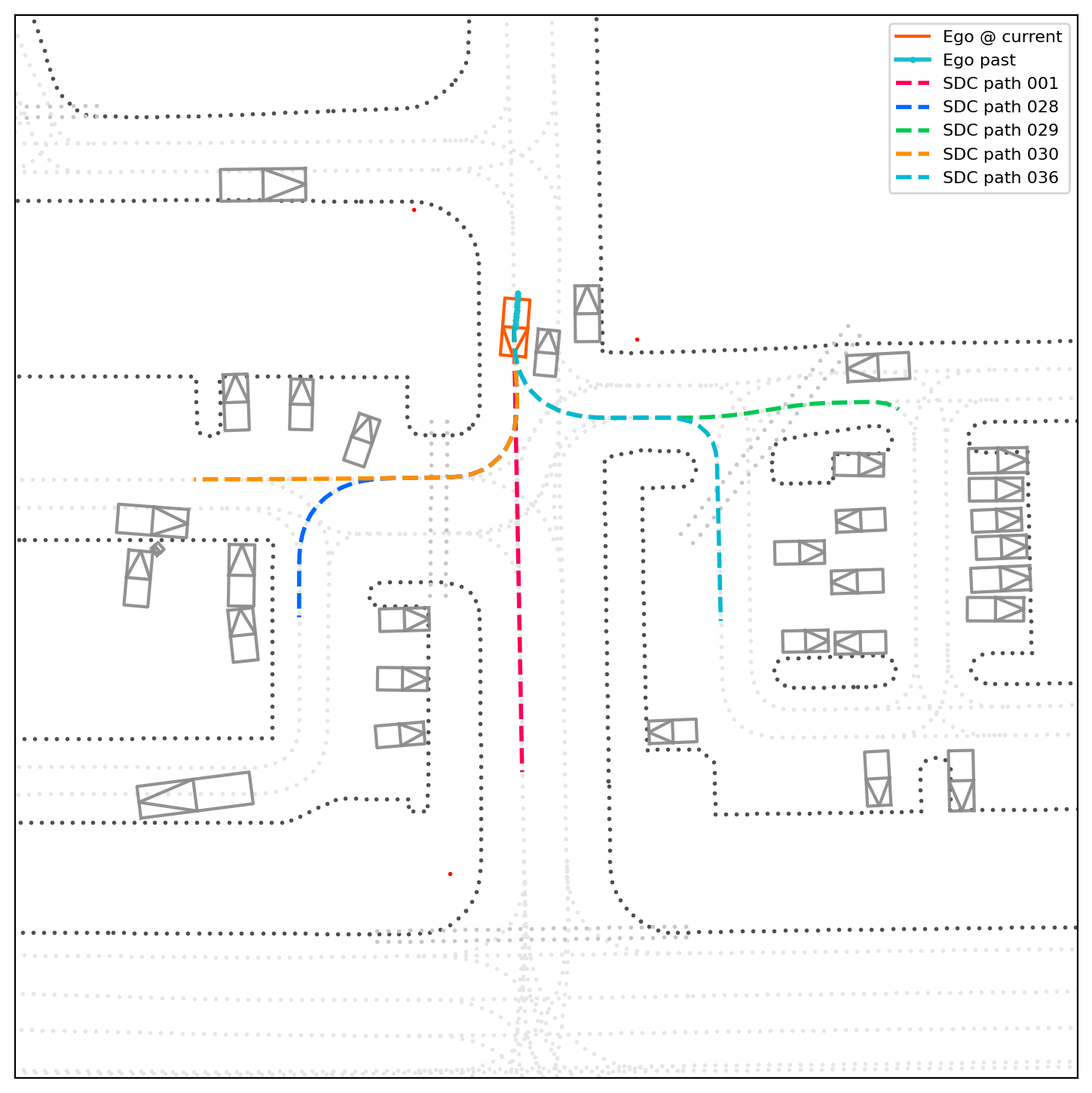} &
\includegraphics[width=0.30\textwidth]{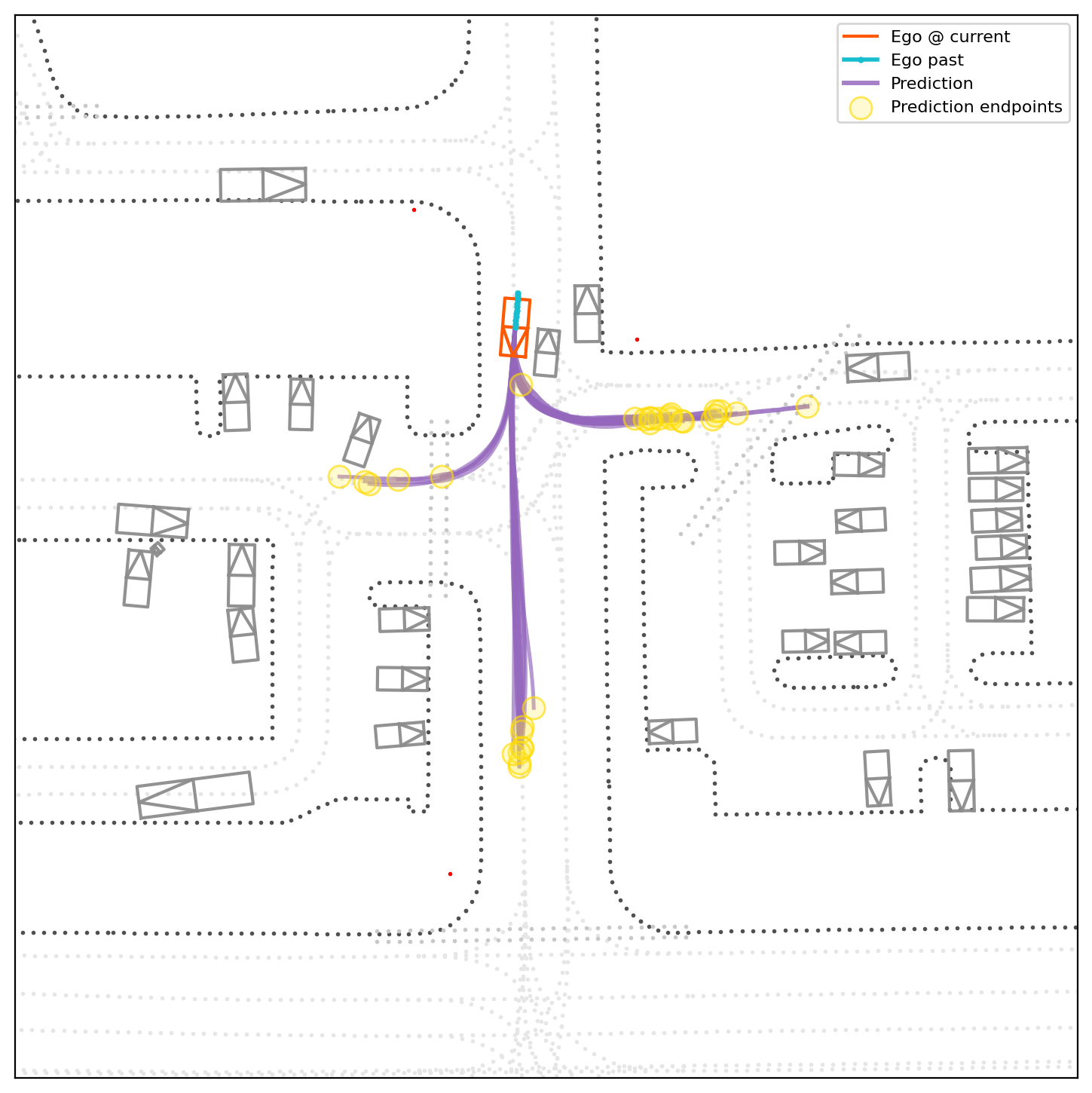} \\
\includegraphics[width=0.30\textwidth]{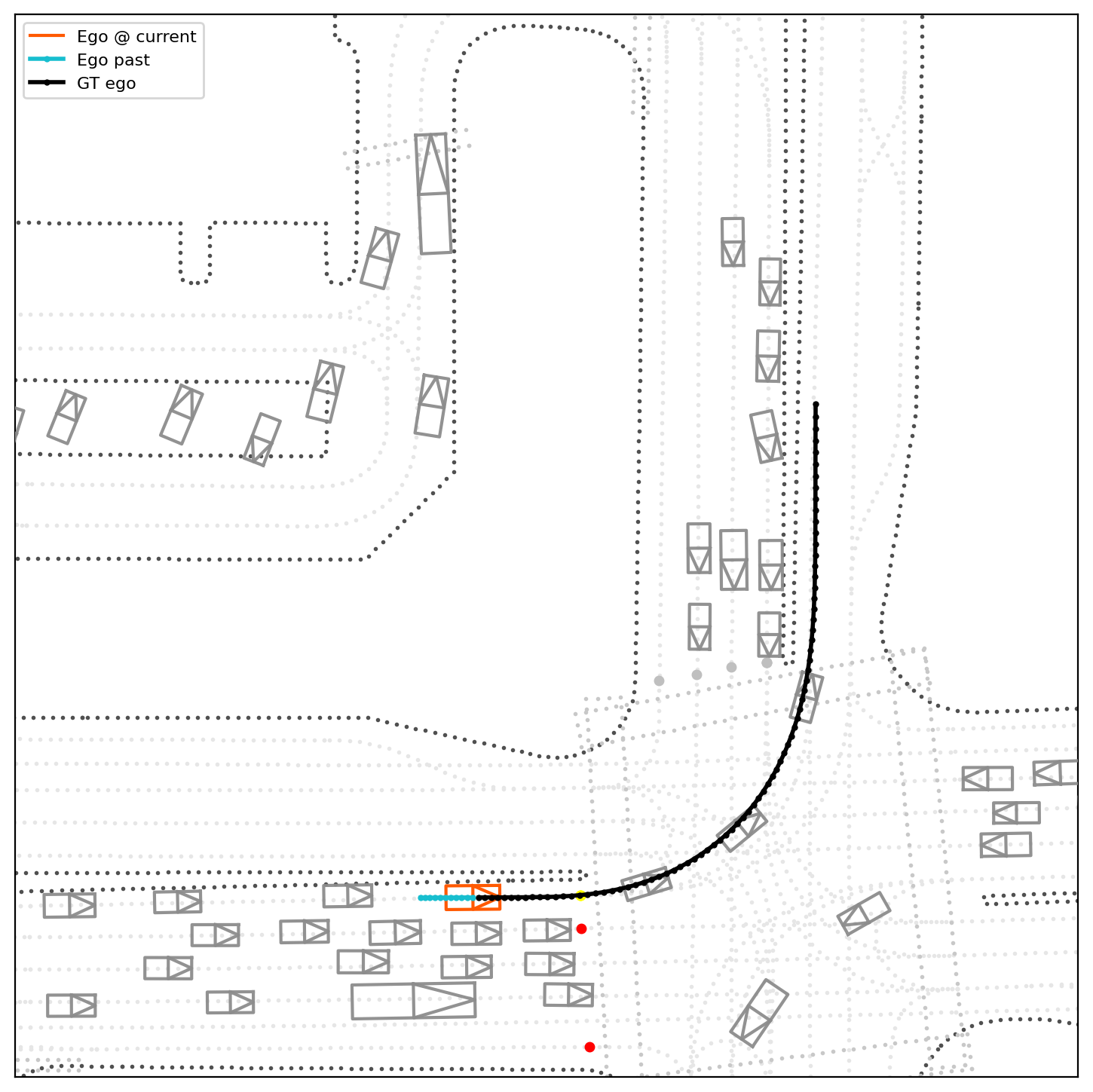} &
\includegraphics[width=0.30\textwidth]{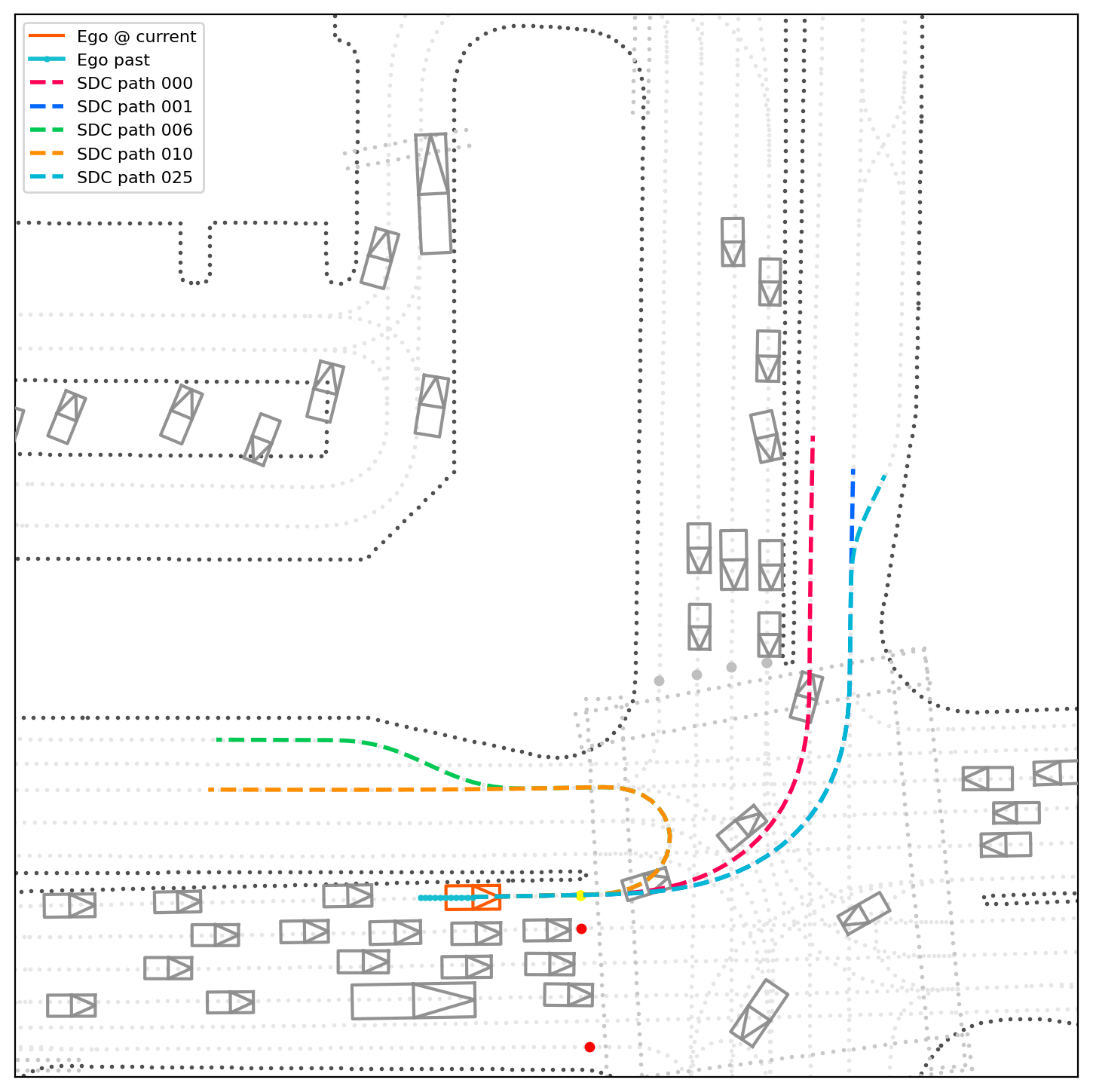} &
\includegraphics[width=0.30\textwidth]{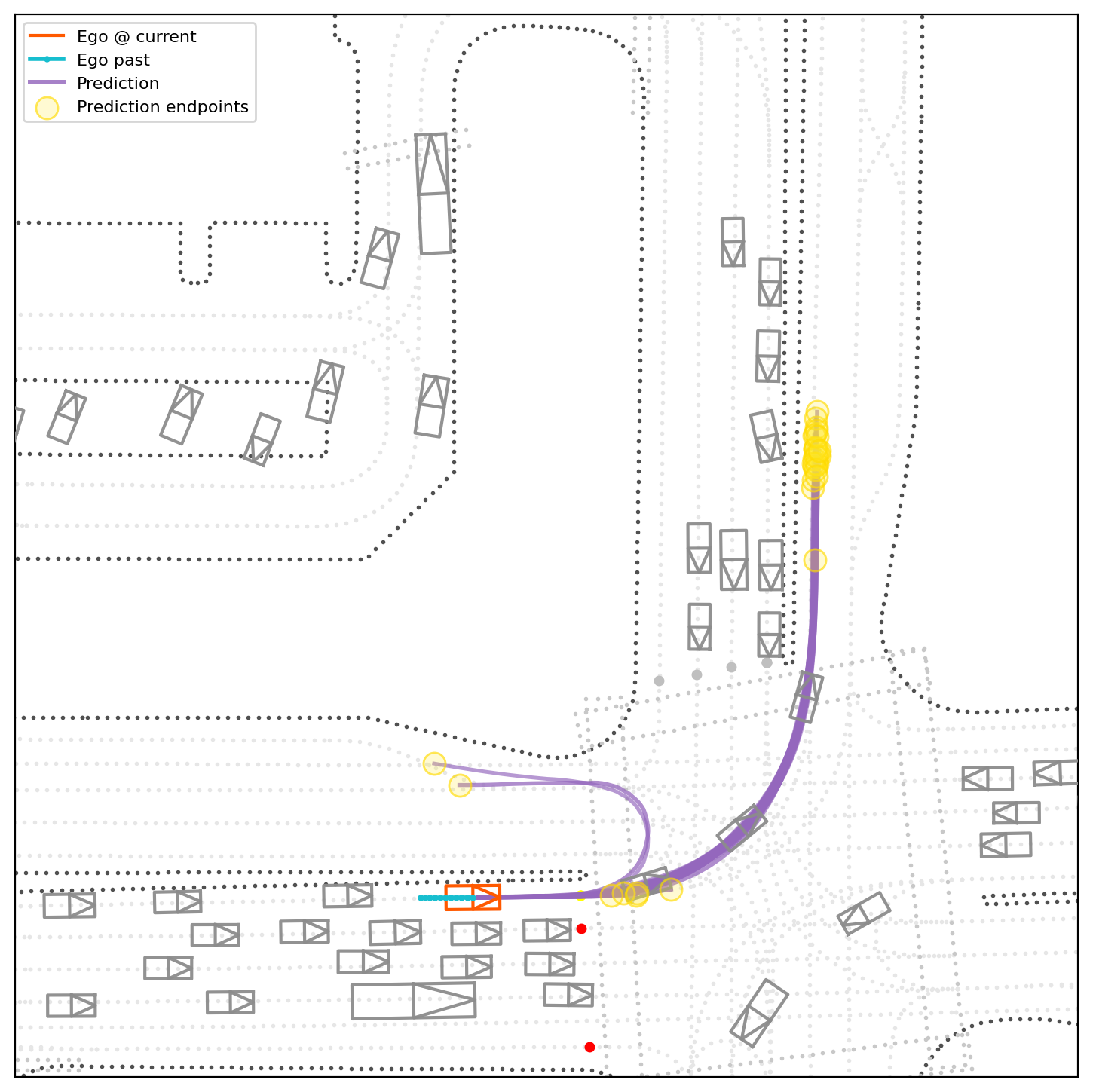} \\
\includegraphics[width=0.30\textwidth]{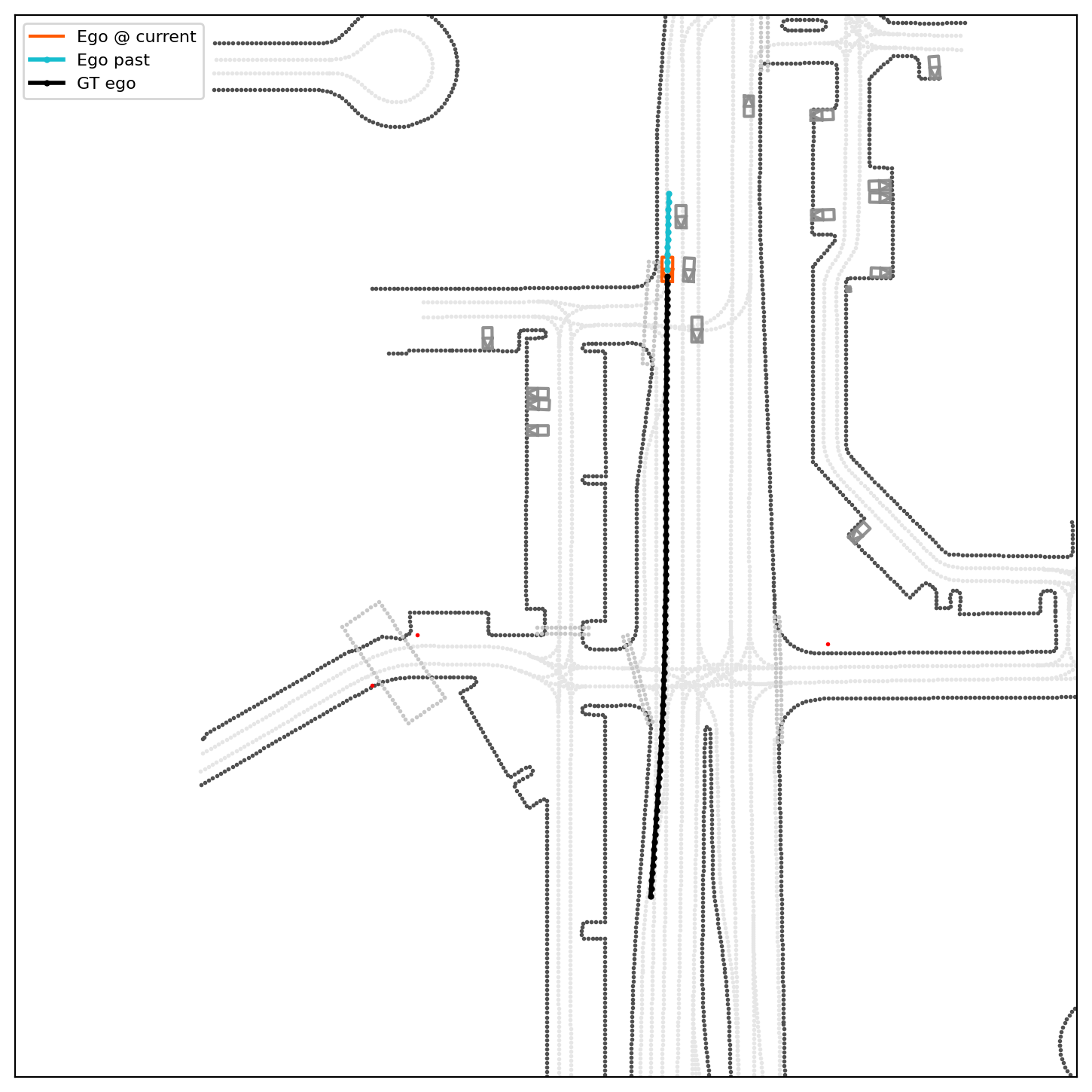} &
\includegraphics[width=0.30\textwidth]{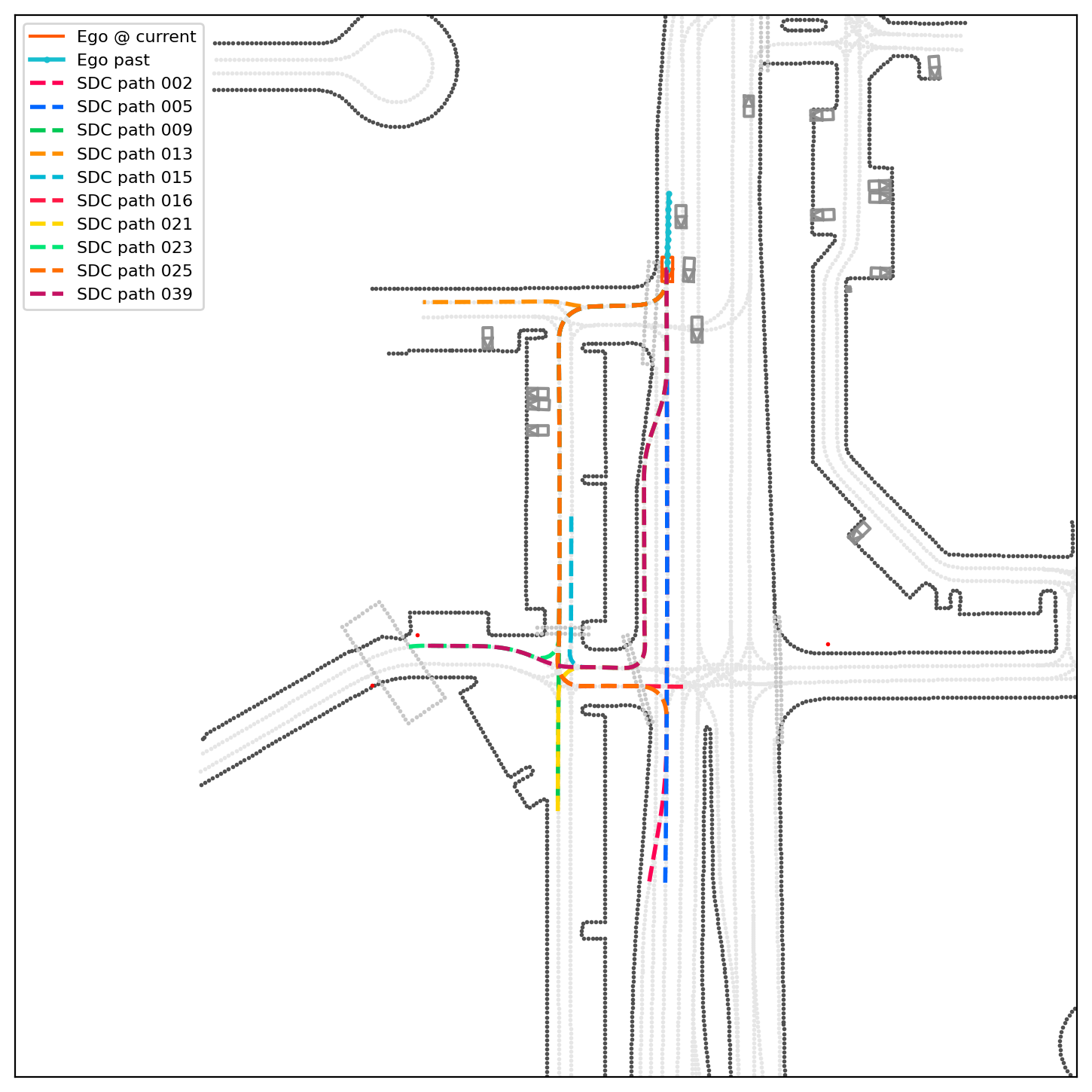} &
\includegraphics[width=0.30\textwidth]{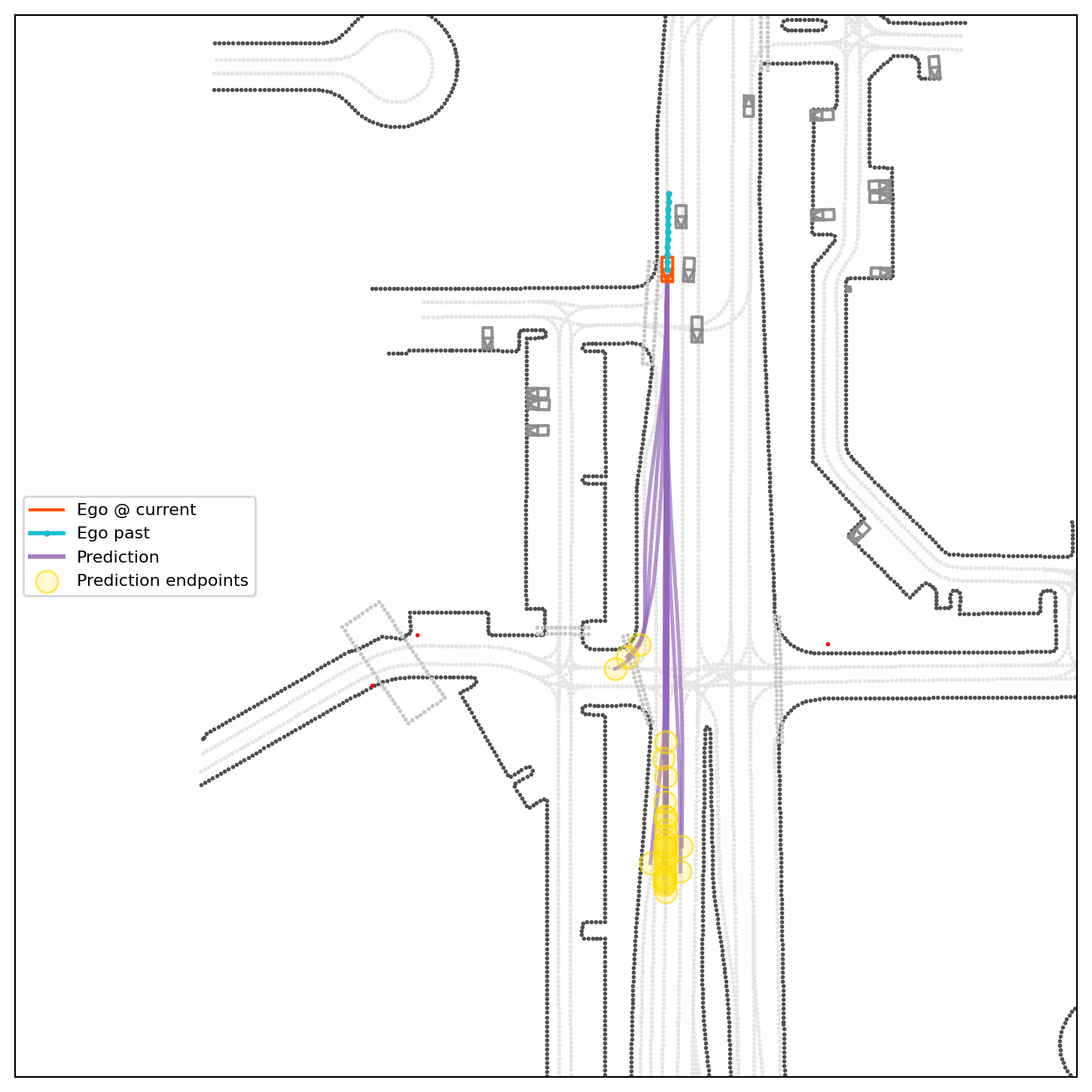} \\
\includegraphics[width=0.30\textwidth]{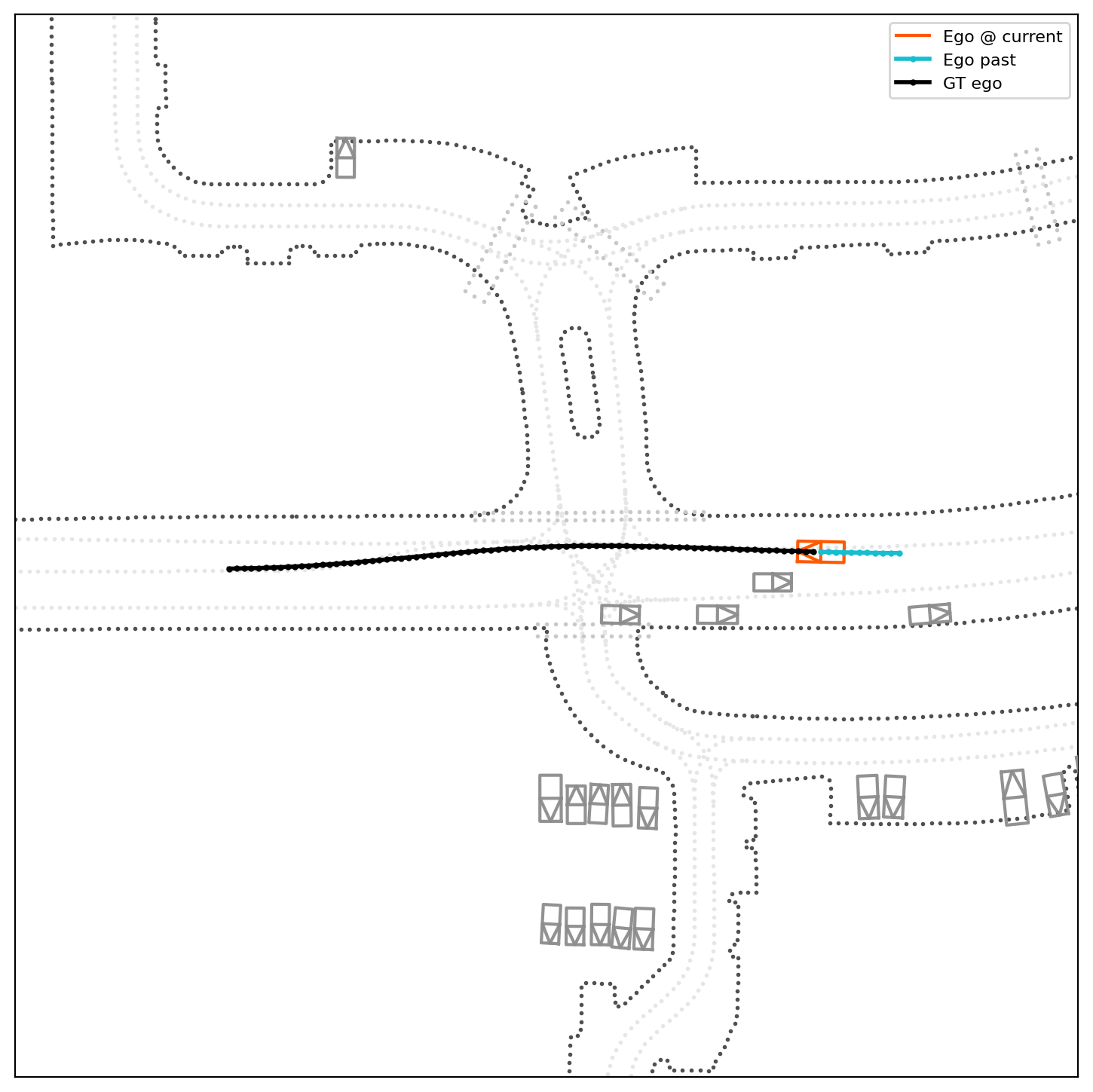} &
\includegraphics[width=0.30\textwidth]{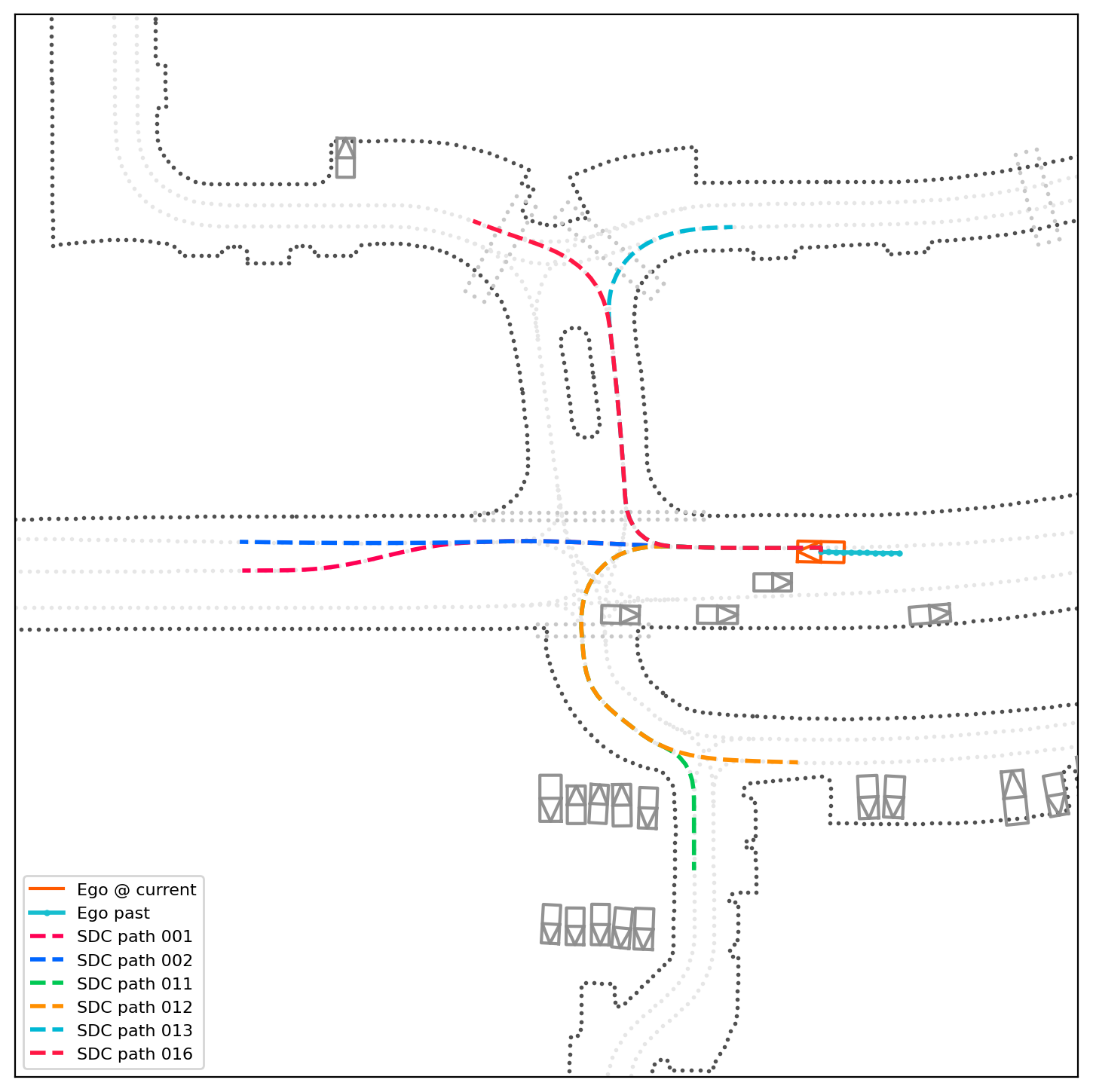} &
\includegraphics[width=0.30\textwidth]{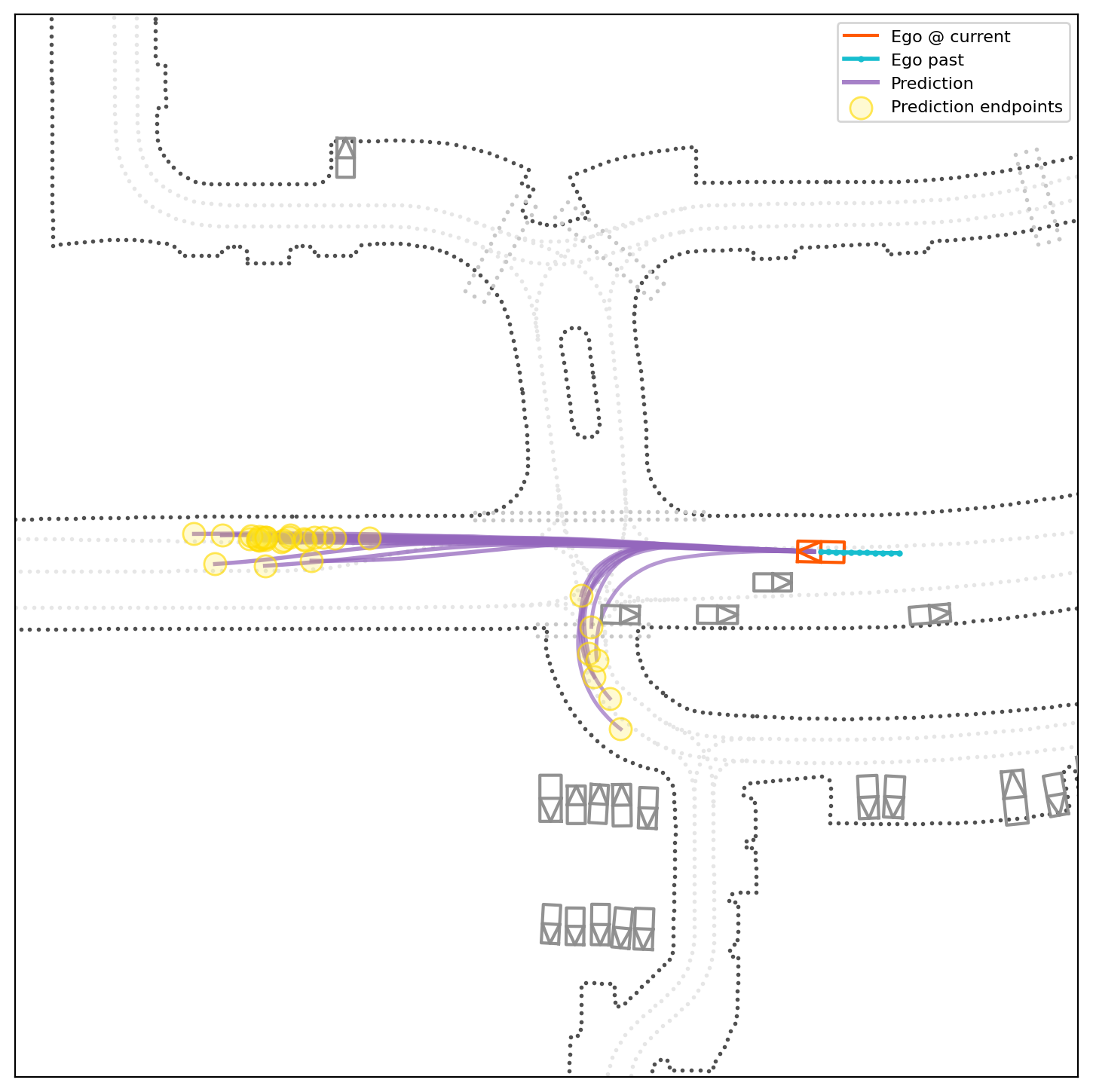} \\
\includegraphics[width=0.30\textwidth]{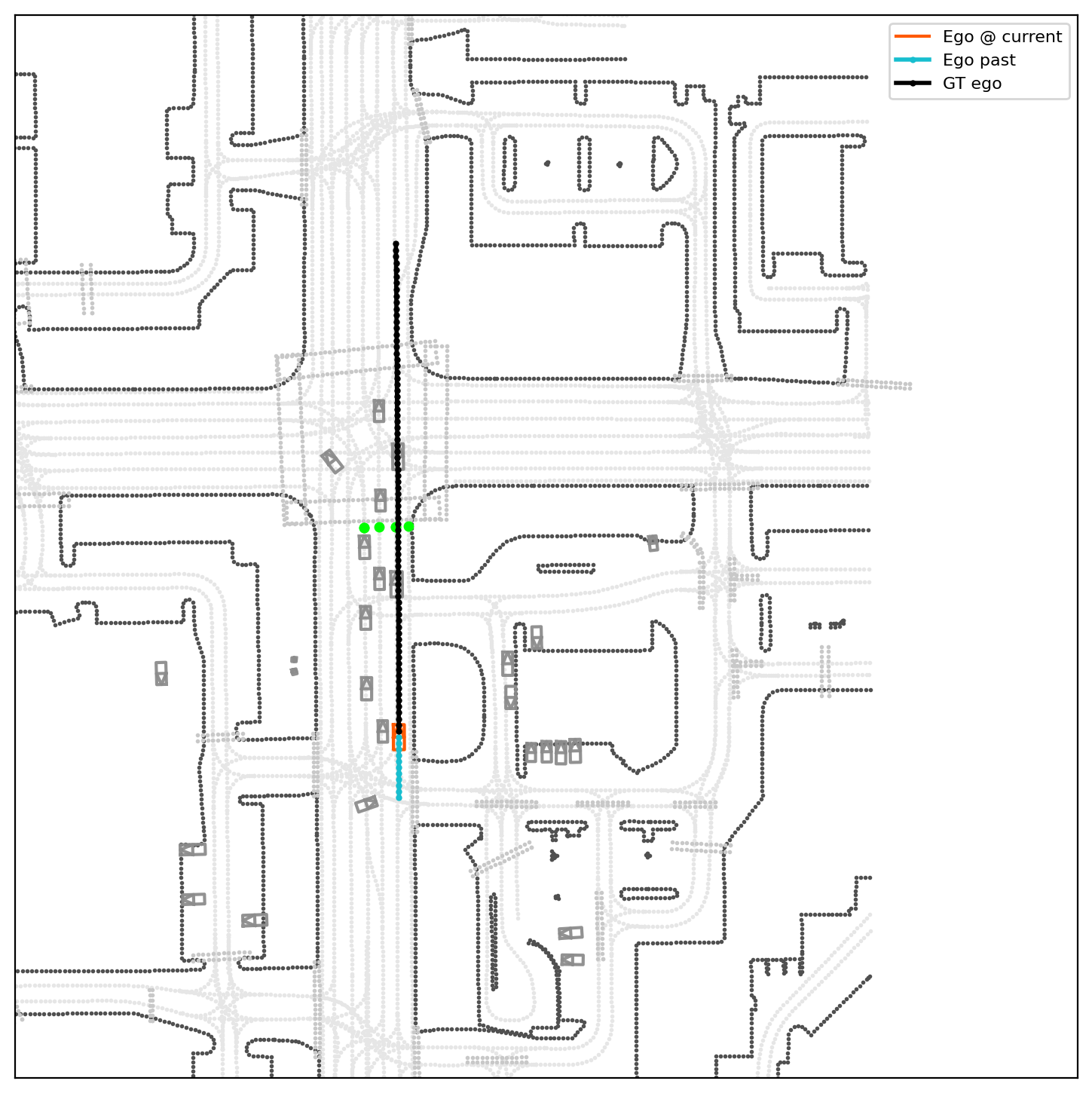} &
\includegraphics[width=0.30\textwidth]{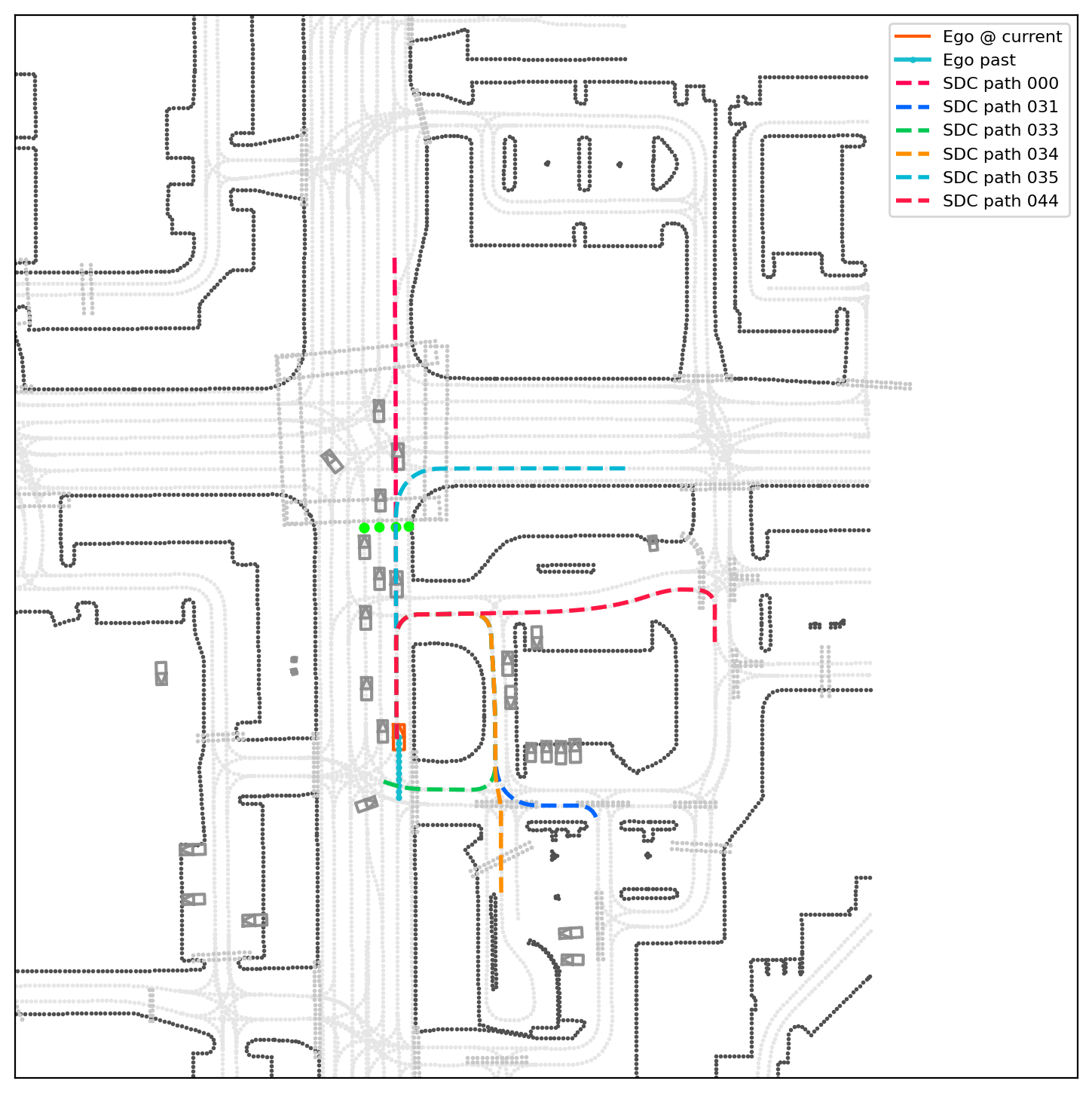} &
\includegraphics[width=0.30\textwidth]{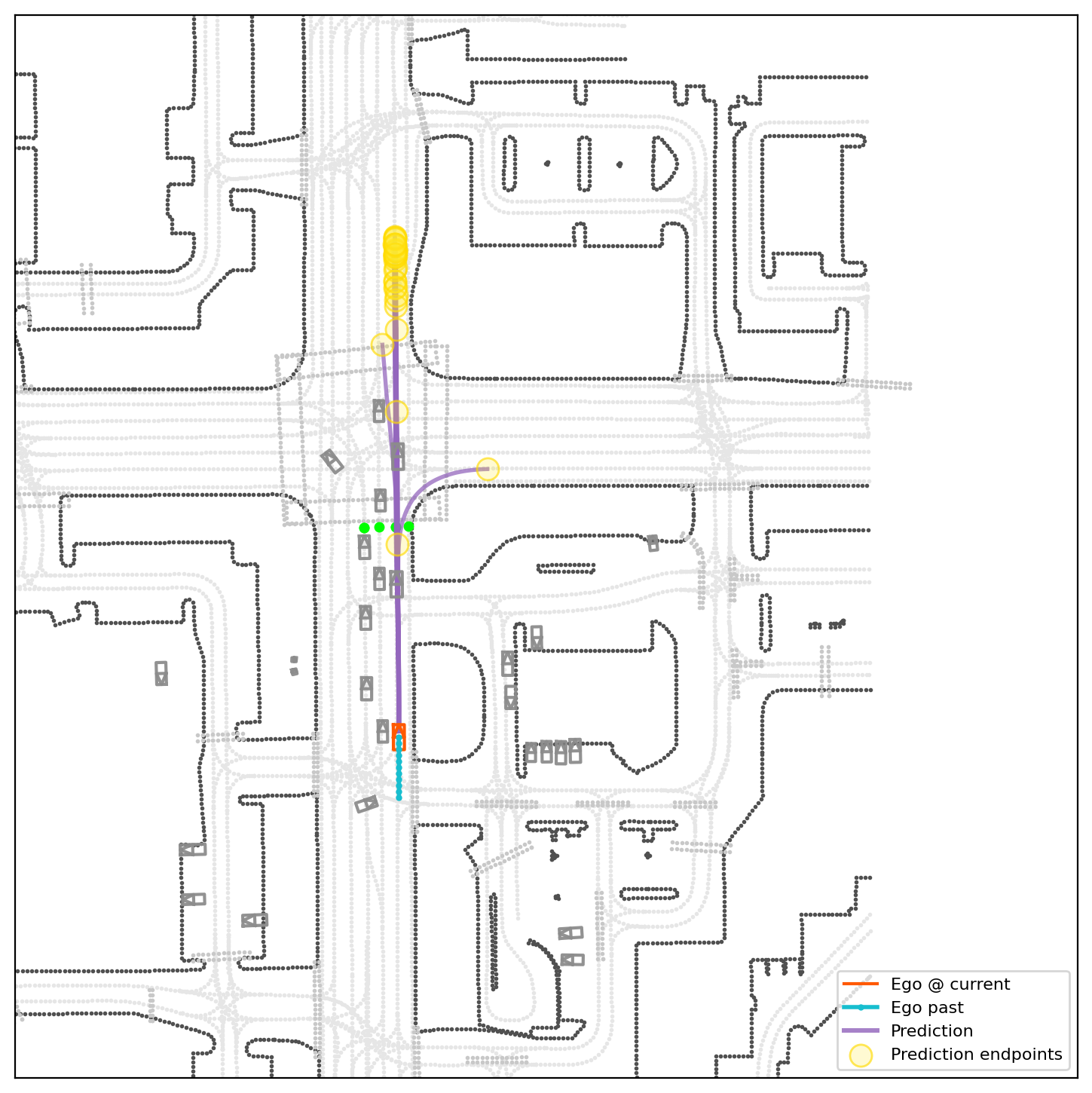} \\
\end{longtable}
\vspace{-8mm}
\end{center}

\section{Implementation Details} 
\label{app_implementation}
In this section, we describe the implementation details of the main components of AutoWorld. 

\subsection{Scene Encoder} The scene encoder processes three inputs derived from the problem formulation: the agent history $S_{t-T_h:t} \in \mathbb{R}^{A \times (T_h+1) \times d_a}$, the HD map $V \in \mathbb{R}^{v \times d_v}$, and the traffic light states $R \in \mathbb{R}^{r \times d_r}$. In our implementation, we use $A=32$ agents, $v = v_l v_p$ map vector points corresponding to $v_l=256$ polylines each containing $v_p=30$ sampled waypoints, and $r=16$ traffic lights. The agent history is encoded using a shared GRU, producing agent embeddings of shape $[A, D]$, which are combined with agent-type embeddings. The map representation $V$ is organized into polylines and encoded using a PointNet-like polyline encoder: each waypoint is processed by a shared MLP to produce features of dimension $D$, followed by max pooling along the waypoint dimension to obtain polyline-level features of shape $[v_l, D]$. The resulting agent, map, and traffic-light embeddings are concatenated to form the initial scene representation of shape $[A + v_l + r, D]$. To model interactions among scene elements, we apply six query-centric Transformer layers following \cite{shi2024mtr++}. This architecture encodes relationships between agents, map polylines, and traffic lights using relative positional features while preserving the element-centric representation of each token. The encoder outputs a final scene representation of shape $[A + v_l + r, D]$, with $D=256$.


\subsection{World Model} 
We build the latent occupancy world model on top of the LiDAR-based framework of~\cite{liu2025foundationallidarworldmodels}. Raw LiDAR point clouds are first voxelized into non-semantic occupancy grids with voxel size $v=[0.4,0.4,0.4]$ meters within spatial range $[-40,-40,-1,$ $40,40,5.4]$. This results in occupancy volumes of size $( H_o, W_o, D_o )=(200,200,16)$, producing binary occupancy tensors $Y_t \in \{0,1\}^{H_o \times W_o \times D_o}$ with corresponding validity masks $M_t$, following~\cite{tian2023occ3d}. We employ a Swin Transformer-based VAE~\cite{liu2025foundationallidarworldmodels} to encode voxel occupancies into spatial latent representations $Z_t \in \mathbb{R}^{H \times W \times C}$. The VAE is trained for 100 epochs with batch size 16 on the voxelized occupancy data. The latent representation has spatial size $H \times W \times C = 16 \times 16 \times 50$.

For latent occupancy forecasting, we follow the rectified flow matching architecture of~\cite{liu2025foundationallidarworldmodels}. The world model predicts future latent occupancies $\hat Z_{t+1:t+T_f} \in \mathbb{R}^{T_f \times H \times W \times C}$ conditioned on past latent context $c^\mu = \{Z_{t-T_h:t}, \tau_{t-T_h:t}^{\text{ego}}\}$. Here, $Z_{t-T_h:t}$ denotes the latent occupancy history and $\tau_{t-T_h:t}^{\text{ego}}$ represents the ego trajectory over the same horizon. The ego trajectory is encoded using the same trajectory encoder employed for agent history in the scene encoder. The encoded trajectory is concatenated with the latent history features and projected through an MLP to produce the final conditioning representation. The temporal horizons $T_h$ and $T_f$ are shared with the motion generation model.

The rectified flow model operates on latent sequences $z_t, z_0, \tilde{z} \in \mathbb{R}^{T_f \times H \times W \times C}$, where $z_0 \sim \mathcal{N}(0,I)$ and $\tilde{z}$ denotes the ground-truth future latent occupancy sequence. The flow model is first trained for 100 epochs with batch size 8 using the standard rectified flow objective. Training then continues for an additional 50 epochs using the proposed motion-aware latent supervision objective, where the motion-aware loss weight is empirically set to $\lambda=0.2$. Following~\cite{liu2025foundationallidarworldmodels}, we maintain an exponential moving average (EMA) of model parameters with decay rate $0.9999$ to stabilize training. We also employ classifier-free guidance by randomly dropping the conditional input $c^\mu$ with probability 0.25.

\subsection{Motion Generation} Given a sampled world-model rollout $\hat Z_{t+1:t+T_f}$, we flatten each latent into $HW$ spatial tokens and project them into the denoiser embedding space of dimension $D$. We then build the predictive scene context $g \in \mathbb{R}^{1 \times D}$ using temporal subsampling at 1~Hz (i.e., stride $\delta=10$ frames). For each future step, we concatenate the corresponding latent occupancy and the predictive scene context along the token dimension. These tokens are fused with the scene encoding through cross-attention to produce a refined step-wise conditioning representation $c_{t+1:T_f}^f$, matching the shape of the scene encoding and used to condition the diffusion denoiser. The denoising network consists of two decoding blocks, each composed of two Transformer decoder layers. Within each block, self-attention models the joint distribution of multi-agent future trajectories while enforcing temporal causality through a causal mask \cite{huang2024dtpp} that prevents information leakage from future timesteps. Cross-attention layers condition the trajectory tokens on the world-model-guided context $c_{t+1:T_f}^f$, allowing the predicted actions to remain consistent with both the observed scene history and the predicted scene evolution. The decoder outputs clean actions $\hat{\tau}_0 \in \mathbb{R}^{A \times T_f \times 2}$. These actions are then integrated using a unicycle dynamics model to obtain the corresponding future agent states $(x, y, v, \theta)$ over the horizon. We set the number of diffusion timesteps to $K_f = 50$.

\subsection{Cascaded Latent Diversity} We use time-dependent scaling coefficients $\gamma_\mu(k_\mu)$ and $\gamma_f(k_f)$ to control the strength of the diversity gradients during sampling in the world model and motion generation stages, respectively. These coefficients follow the diffusion and flow-matching timestep schedules and are normalized by the norm of the corresponding DPP gradients to ensure stable updates. When $\gamma(\cdot)=0$, the procedure reduces to standard IID sampling. When $\gamma(\cdot)>0$, the DPP gradient encourages diversity by pushing similar samples apart. If the generated samples are already sufficiently diverse, the DPP gradient becomes negligible, and the process naturally behaves like IID sampling.

As stated in the main paper, we evaluate the plausibility of generated trajectories using the violation metrics provided by the Waymax simulation platform \cite{gulino2023waymax}. Specifically, we consider the kinematic infeasibility (kin.), collision (col), off-road (off.), and wrong-way metrics (wro.). To obtain a single scalar score for use in DPP-based sampling, we combine these metrics using weights inspired by the Realism Meta-Metric (RMM), which aggregates kinematic, interactive, and map-based realism as $\mathrm{RMM} = 0.20\,\text{(kinematic)} + 0.45\,\text{(interactive)} + 0.35\,\text{(map-based)}$. Following this structure, we compute the trajectory quality score as $Q = 1 - (0.20\,\text{(kin.)} + 0.45\,\text{(col.)} + 0.35\,\frac{\text{off.} + \text{wro.}}{2})$. The weights reflect the relative importance of kinematic feasibility, interaction safety, and adherence to map constraints. This scalar score is used as the quality term in the DPP kernel, encouraging the selection of trajectories that are physically feasible, collision-free, and consistent with the road geometry.

\subsection{AutoWorld's Notations}

For clarity, the key notations used throughout the method are summarized in Table~\ref{tab:notations}.

{\scriptsize
\begin{longtable}{lll}
\caption{Main notations used in AutoWorld. For functions, ``Domain'' refers to the domain of the function's output.}
\label{tab:notations} \\

\toprule
\textbf{Notation} & \textbf{Name / Description} & \textbf{Domain} \\
\midrule
\endfirsthead

\toprule
\textbf{Notation} & \textbf{Name / Description} & \textbf{Domain} \\
\midrule
\endhead

\midrule
\multicolumn{3}{r}{Continued on next page}
\endfoot

\bottomrule
\endlastfoot

\multicolumn{3}{l}{\texttt{Scene encoder}} \\

$A$ & Number of agents & $\mathbb N$ \\

$v_l$ & Number of map polylines & $\mathbb N$ \\

$v_p$ & Number of sampled waypoints per polyline & $\mathbb N$ \\

$v$ & Total number of map vector points ($v=v_l v_p$) & $\mathbb N$ \\

$r$ & Number of traffic lights & $\mathbb N$ \\

$d_a$ & Agent state dimension & $\mathbb N$ \\

$d_v$ & Map feature dimension & $\mathbb N$ \\

$d_r$ & Traffic-light feature dimension & $\mathbb N$ \\

$V$ & HD map vector representation & $\mathbb{R}^{v \times d_v}$ \\

$R$ & Traffic light states & $\mathbb{R}^{r \times d_r}$ \\

$s_t^i$ & State of agent $i$ at time $t$ & $\mathbb{R}^{4}$ \\

$s_t$ & Joint agent states & $\mathbb{R}^{A \times 4}$ \\

$S_{t-T_h:t}$ & Agent state history & $\mathbb{R}^{A \times (T_h+1) \times 4}$ \\

$h$ & Scene encoding output & $\mathbb{R}^{(A + v_l + r) \times D}$ \\

$D$ & Scene embedding dimension & $\mathbb N$ \\

\midrule

\multicolumn{3}{l}{\texttt{World model}} \\

$t$ & Discrete scene timestep & $\mathbb N$ \\
$T_h$ & History horizon & $\mathbb N$ \\
$T_f$ & Future prediction horizon & $\mathbb N$ \\


$Y_t$ & Occupancy grid at time $t$ & $\{0,1\}^{H_o \times W_o \times D_o}$ \\
$M_t$ & Validity mask & $\{0,1\}^{H_o \times W_o \times D_o}$ \\

$\tilde{Y}_{t\rightarrow t+\Delta}$ & Ego-motion compensated occupancy grid & $\{0,1\}^{H_o \times W_o \times D_o}$ \\

$C_{t,\Delta}(v)$ & Voxel occupancy transition indicator & \{0,1\} \\

$\bar{C}_{t,\Delta}$ & Downsampled motion map & $\mathbb{R}^{H \times W}$ \\

$W_t$ & Motion-aware latent weight map & $\mathbb{R}^{H \times W}$ \\

$\lambda$ & Motion emphasis coefficient & $\mathbb{R}_{>0}$ \\

$Z_t$ & Latent occupancy representation at time $t$ & $\mathbb{R}^{H \times W \times C}$ \\

$Z_{t-T_h:t}$ & Latent occupancy history & $\mathbb{R}^{(T_h+1)\times H \times W \times C}$ \\

$Z_{t+1:t+T_f}$ & Ground-truth future latent occupancies & $\mathbb{R}^{T_f \times H \times W \times C}$ \\

$\hat{Z}_{t+1:t+T_f}$ & Predicted future latent occupancies & $\mathbb{R}^{T_f \times H \times W \times C}$ \\

$z_0$ & Gaussian prior latent sequence sample & $\mathbb{R}^{T_f \times H \times W \times C}$ \\

$z_{k_\mu}$ & Latent variable at rectified-flow step $k_\mu$ & $\mathbb{R}^{T_f \times H \times W \times C}$ \\


$\mu_\theta(\cdot)$ & World-model velocity field network & $\mathbb{R}^{T_f \times H \times W \times C}$ \\

$k_\mu$ & Rectified-flow timestep & $\mathbb N$ \\

\midrule

\multicolumn{3}{l}{\texttt{Motion generation}} \\

$a_t^i$ & Agent action (acceleration, yaw rate) & $\mathbb{R}^{2}$ \\

$a_t$ & Joint agent actions & $\mathbb{R}^{A \times 2}$ \\

$\tau_0$ & Clean future trajectory (action sequence) & $\mathbb{R}^{A \times T_f \times 2}$ \\

$\tau_{k_f}$ & Noisy trajectory during diffusion step $k_f$ & $\mathbb{R}^{A \times T_f \times 2}$ \\

$g$ & Predictive scene context & $\mathbb{R}^{1 \times D}$ \\

$\delta$ & Temporal subsampling stride & $\mathbb N$ \\

$\xi$ & Learned pooling query token & $\mathbb{R}^{1 \times D}$ \\

$\phi_1(\cdot)$ & Latent projection for context pooling & $\mathbb R^{D_{\phi_1} \times 1}$ \\

$\phi_2(\cdot)$ & Conditioning projection & $\mathbb R^{D_{\phi_2} \times 1}$\\

$c_t^f$ & Diffusion conditioning representation at timestep $t$ & $\mathbb{R}^{(A+v_l+r) \times D}$ \\

$c^f$ & Full diffusion conditioning sequence & $\mathbb{R}^{T_f \times (A+v_l+r) \times D}$ \\

$f_\psi(\cdot)$ & Diffusion denoiser network & $\mathbb{R}^{A \times T_f \times 2}$ \\

$\Sigma_{k_f}$ & Diffusion variance schedule & $M_{\succeq 0}$ \\

$k_f$ & Diffusion timestep & $\mathbb N$ \\

\midrule
\multicolumn{3}{l}{\texttt{Cascaded latent sampling}} \\

$\mathcal{X}$ & Candidate sample set & $\{\hat{x}^{(i)}\}_{i=1}^{K_s}$ \\

$\hat{x}^{(i)}$ & Candidate sample representation & $\mathbb{R}^{d}$ \\

$K_s$ & Number of samples in DPP set & $\mathbb N$\\

$\kappa(\cdot,\cdot)$ & Similarity kernel & $\mathbb{R}$ \\

$\Lambda$ & Kernel matrix & $\mathbb{R}^{K_s \times K_s}$ \\

$\mathcal{P}^q_\kappa(\mathcal{X})$ &  Quality-aware DPP diversity probability & [0,1] \\

$q^{(i)}$ & Quality weight for sample $i$ & $\mathbb {R}$ \\

$\Lambda^q$ & Quality-aware kernel matrix & $\mathbb{R}^{K_s \times K_s}$ \\

$\mathcal{X}^{\text{world}}\!\doteq\!\{\hat{Z}^{(i)}_{t+1:t+T_f}\}_{i=1}^N$ & World-model rollout set &  $\mathbb{R}^{N\times T_f \times H \times W \times C}$\\

$\gamma_\mu(k_\mu)$ & World-model diversity guidance strength & $\mathbb R_{>0}$ \\

$N$ & Number of sampled world-model rollouts & $\mathbb N$ \\

$\mathcal{X}^{\text{motion}}\doteq \{\tau^{(i,j)}\}_{j=1}^M$ & Motion trajectory sample set &  $\mathbb{R}^{M\times A \times T_f \times 2}$\\

$\gamma_f(k_f)$ & Motion diversity guidance strength & $\mathbb R_{>0}$ \\

$M$ & Number of motion samples per rollout & $\mathbb N$ \\
\end{longtable}
}

\end{document}